\documentclass{article}

\usepackage{microtype}
\usepackage{graphicx}
\usepackage{subcaption}
\usepackage{booktabs} 
\usepackage{fvextra}

\newcommand{\OG}{OG}
\usepackage{hyperref}

\usepackage[accepted]{icml2026}

\usepackage{amsmath}
\usepackage{amssymb}
\usepackage{mathtools}
\usepackage{amsthm}
\usepackage{times}
\usepackage{latexsym}
\usepackage{microtype}
\usepackage{inconsolata}
\usepackage{graphicx}
\usepackage{amssymb}      
\usepackage{amsfonts}     
\usepackage{enumitem}     
\usepackage[capitalize,noabbrev]{cleveref}
\usepackage{booktabs} 
\usepackage{multirow} 
\usepackage{makecell}
\usepackage[table]{xcolor}%
\usepackage{longtable}
\usepackage[section]{placeins}
\usepackage{caption}
\usepackage{enumitem}

\theoremstyle{plain}

\theoremstyle{definition}

\theoremstyle{remark}

\usepackage[textsize=tiny]{todonotes}

\icmltitlerunning{Toward Culturally Aligned LLMs through Ontology-Guided Multi-Agent Reasoning}

\begin{document}

\twocolumn[
  \icmltitle{Toward Culturally Aligned LLMs through Ontology-Guided Multi-Agent Reasoning}

  \icmlsetsymbol{equal}{*}

  \begin{icmlauthorlist}
    \icmlauthor{Wonduk Seo}{enhans}
    \icmlauthor{Wonseok Choi}{pku,equal}
    \icmlauthor{Junseo Koh}{pku,equal}
    \icmlauthor{Juhyeon Lee}{pku}
    \icmlauthor{Hyunjin An}{enhans}
    \icmlauthor{Minhyeong Yu}{enhans}
    \icmlauthor{Jian Park}{fudan}
    \icmlauthor{Qingshan Zhou}{pku}
    \icmlauthor{Seunghyun Lee}{enhans}
    \icmlauthor{Yi Bu}{pku}
  \end{icmlauthorlist}

  \icmlaffiliation{enhans}{AI Research, Enhans, Seoul, South Korea}
  \icmlaffiliation{pku}{Department of Information Management, Peking University, Beijing, China}
  \icmlaffiliation{fudan}{Department of Data Science, Fudan University, Shanghai, China}

  \icmlcorrespondingauthor{Yi Bu}{buyi@pku.edu.cn}
  \icmlkeywords{Large Language Model, Cultural Alignment, Information Retrieval, Ontology Retrieval, Multi-Agent Reasoning}

  \vskip 0.3in
]

\printAffiliationsAndNotice{\icmlEqualContribution}

\begin{abstract}
Large Language Models (LLMs) increasingly support culturally sensitive decision making, yet often exhibit misalignment due to skewed pretraining data and the absence of structured value representations. Existing methods can steer outputs, but often lack demographic grounding and treat values as independent, unstructured signals, reducing consistency and interpretability. We propose \textbf{OG-MAR}, an \textbf{\underline{O}}ntology-\textbf{\underline{G}}uided \textbf{\underline{M}}ulti-\textbf{\underline{A}}gent \textbf{\underline{R}}easoning framework. \textbf{OG-MAR} summarizes respondent-specific values from the World Values Survey (WVS) and constructs a global cultural ontology by eliciting relations over a fixed taxonomy via competency questions. At inference time, it retrieves ontology-consistent relations and demographically similar profiles to instantiate multiple value-persona agents, whose outputs are synthesized by a judgment agent that enforces ontology consistency and demographic proximity. Experiments on regional social-survey benchmarks across four LLM backbones show that \textbf{OG-MAR} improves cultural alignment and robustness over competitive baselines, while producing more transparent reasoning traces.
\end{abstract}

\section{Introduction}
Large Language Models (LLMs) are predominantly trained on web-scale corpora that are unevenly distributed across regions and sociocultural contexts \cite{bender2021dangers, dodge2021documenting, achiam2023gpt, jiang2023mistral7b, touvron2023llama}. As a result, they often inherit culture-default biases, prioritizing high-resource Western-centric viewpoints while underrepresenting diverse cultural value systems \cite{durmus2023globalopinionqa, gallegos2024bias, xie2024can}. These biases lead to systematic misalignment in culturally sensitive tasks, particularly ones involving social norms and value-based decisions \cite{karinshak2024llm, pistilli2024civics, tao2024cultural}. In response, several countries and organizations have developed localized LLMs to better reflect region-specific values \cite{zeng2022glm, sengupta2023jais, avramidis2024occiglot, nguyen2024seallms, yoo2024hyperclova}. Nevertheless, cultural bias and value misalignment continue to pose challenges for LLM deployments \cite{hershcovich2022challenges, kreutzer2022quality}.

Prior work has proposed several strategies to reduce cultural bias in LLMs. Role-assignment methods \cite{tao2024cultural} steer behavior by specifying culturally grounded personas, while few-shot prompting uses curated cultural exemplars to guide generation \cite{choenni2024self}. Retrieval-based approaches such as ValuesRAG \cite{seo2025valuesrag} further ground outputs in external survey evidence to better match cultural preferences. More recently, multi-agent frameworks \cite{baltaji2024persona, ki2025multiple, wan2025cultural} simulate diverse viewpoints through agent interaction and deliberation. In particular, the debate-only framework \citep{ki2025multiple} relies on iterative critique and refinement to improve cultural adaptability.

Despite their promise, existing approaches share several fundamental limitations: (1) they often depend on implicit cultural assumptions that are weakly grounded in empirical value distributions, making outputs brittle and sensitive to prompting choices; (2) even with external evidence, cultural values are frequently treated as independent signals, missing structural relationships and cross-topic dependencies; (3) aggregation and multi-agent methods can boost robustness and diversity but using multiple agents without concrete value structure or grounding often reduces interpretability, offering limited visibility into why specific viewpoints emerge.

To address these issues, we propose \textbf{OG-MAR}, an ontology-driven cultural reasoning framework that integrates structured value knowledge, demographic grounding, and multi-agent simulation. We specifically use the World Values Survey (WVS) \cite{zhao2024worldvaluesbench} as an empirically grounded retrieval corpus capturing diverse value distributions across regions. Raw survey responses are converted into topic-aware value summaries, and a global cultural ontology is built using expert-designed Competency Questions (CQs) \cite{gruninger1995methodology, gruninger1995role}, region-stratified LLM reasoning, and human-guided consolidation. At inference time, we retrieve ontology-consistent value structures and demographically similar individuals, instantiate value-persona agents for culturally grounded reasoning, and synthesize their outputs via a principled judgment mechanism.

We evaluate our framework on six regional benchmarks from major social surveys spanning East Asia, South Asia, Europe, North America, Latin America, and Africa. Results show consistent gains in cultural alignment, robustness across question types, and interpretability of reasoning traces. Quantitative analyses and qualitative case studies further indicate that ontology-guided multi-agent simulation offers a scalable and reliable path to culturally aligned LLM inference.

\section{Related Work}
\subsection{Evaluation of LLM's Cultural Alignment}
Large Language Models (LLMs) exhibit strong linguistic ability, yet geographic and linguistic skew in pretraining data can embed dominant-region norms as implicit defaults, causing cultural misalignment and potential inequities across contexts \citep{alkhamissi2024investigating}. Prior work evaluated these biases using stereotype-focused benchmarks \citep{nadeem2021stereoset}, open-ended generation measures \citep{dhamala2021bold}, and task-level tests such as ambiguous question answering \citep{parrish2022bbq}. Later studies measured cultural alignment by comparing model outputs with representative value surveys like the World Values Survey \citep{haerpfer2022world}. Recent benchmarks further assess culture-specific everyday knowledge \citep{chiu2024culturalbench}, cross-national norm adaptation under different cultural frames \citep{rao2025normad}, and value structure in open-ended generations through cultural psychology lenses \citep{karinshak2024llm}. Building on this line of work, we use six regionally diverse survey datasets to evaluate value alignment across a broad set of cultural contexts.

\subsection{Mitigating Cultural Bias in LLMs}
Cultural bias mitigation for Large Language Models has evolved from in-context prompting to structured, evidence-grounded methods. Cultural prompting steers outputs by specifying a cultural frame \citep{tao2024cultural}, while Anthropological Prompting adds richer context and reasoning for underrepresented personas \citep{alkhamissi2024investigating}. To reduce sensitivity to examples and language, self-alignment selects culturally aligned demonstrations for in-context learning \citep{choenni2024self}. When demonstrations are limited, ValuesRAG retrieves cultural and demographic cues as external evidence \citep{seo2025valuesrag}. Agentic approaches further improve reliability and parity through multi-agent debate and planning--critique--refinement pipelines \citep{ki2025multiple,wan2025cultural}. Despite this shift toward grounding and multi-agent reasoning, most methods still model cultural knowledge as unstructured, motivating ontology engineering to capture explicit value relationships.

\subsection{Ontology Engineering with LLMs}
An ontology is a formal specification of domain concepts and their relations \citep{Gruber93}, supporting consistent and interpretable retrieval, integration, and reasoning. METHONTOLOGY \citep{Fernandez1997}, On-To-Knowledge \citep{SureSS04}, and NeOn \citep{Suarez2012} provide structured lifecycles and emphasize reuse. Recent work uses LLMs for ontology extraction via zero-shot prompting \citep{GiglouDA23} or fine-tuning \citep{MateiuG23}, and increasingly automates end-to-end development: CQbyCQ matches novice-level performance \citep{SaeedizadeB24}, while Memoryless CQbyCQ and Ontogenia improve context efficiency and reasoning quality \citep{LippolisSKZCGBN25}. Despite the benefits of ontology-based structuring for stereotype reduction, it remains underused in cultural-bias mitigation. We therefore propose a multi-agent framework for ontology-aware reasoning and pseudo-answer simulation.

\section{Proposed Framework}
\label{sec:framework}

\begin{figure*}[h]
    \centering
    \includegraphics[width=0.8\textwidth]{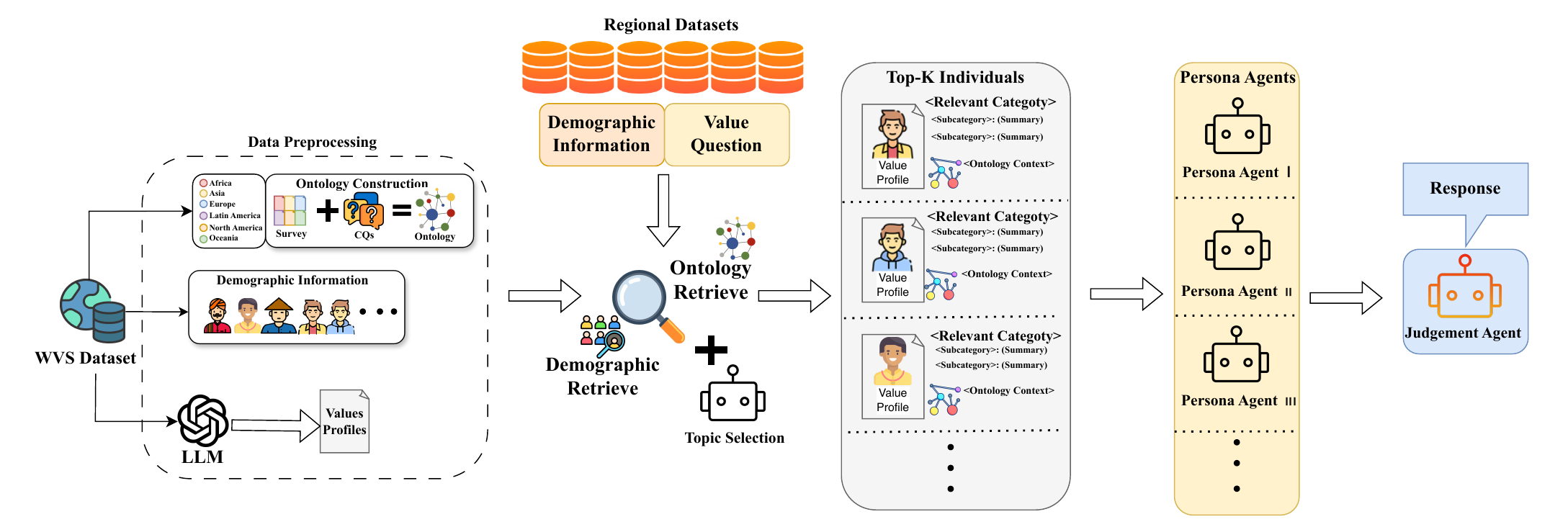}
    \caption{\textbf{Overall architecture of the \textbf{OG-MAR} framework.} The pipeline illustrates the overall architecture of  \textbf{OG-MAR}. It begins with Data Preprocessing \& Ontology Construction (left). During inference, for a given query and target demographics, it performs Ontology \& Demographic Retrieval (center) to gather relevant context. This context is used to instantiate multiple Persona Agents (top right) whose outputs are synthesized by a Judgment Agent (bottom right) to produce the final, culturally aligned prediction.}
    \label{fig:main_architecture}
\end{figure*}

We propose \textbf{OG-MAR}, \textbf{\underline{O}}ntology-\textbf{\underline{G}}uided \textbf{\underline{M}}ulti-\textbf{\underline{A}}gent \textbf{\underline{R}}easoning framework that (i) summarizes respondent values under a fixed taxonomy, (ii) constructs CQ-derived cross-category relations as an ontology, (iii) retrieves ontology triples and demographically similar profiles for a query, and (iv) performs multi-persona simulation with ontology-constrained final adjudication.

\subsection{Data Preprocessing \& Structuring}
\label{sec:data_preprocess_structuring}

\subsubsection{Topic-Aware Value Summary Generation}
\label{sec:topic_aware_summary}
Large-scale surveys capture cultural values through diverse question types, varying response scales, and heterogeneous answer formats.
Directly operating on raw survey answers risks conflating unrelated signals and amplifying noise.
To address this issue, we generate structured value summaries aligned with a predefined ontology taxonomy\footnote{Taxonomy process and results are provided in Appendix~\ref{sec:ontology_details}.}.
Each World Values Survey (WVS) respondent record is decomposed into (1) demographic attributes and (2) values-related responses.
Given the fixed ontology class set $C$ ($|C|=76$), which comprises 12 top-classes $c_1,\dots,c_{12}$ and their 64 sub-classes $c_{i,j}$\footnote{We write $c_i$ for the $i$-th top-class and $c_{i,j}$ for its $j$-th sub-class. The full taxonomy is detailed in Table~\ref{tab:value_taxonomy}.}, a Summarization Agent $G_{\text{sum}}$ generates a concise, category-specific synopsis of the respondent's stance within the semantic scope of each class in the taxonomy.
The agent is instructed to summarize only information relevant to the given class.
Formally, let $\mathcal{R}_i$ denote the raw response set for individual $i$. For each class $c\in C$, we obtain a category-conditioned synopsis:
\begin{equation}
s_i(c) = G_{\text{sum}}(\mathcal{R}_i \mid c).
\end{equation}
Aggregating over all classes yields a structured value profile:
\begin{equation}
V_i = \{ s_i(c) \mid c \in C \}.
\end{equation}

Consequently, each individual is represented by a structured value profile that supports subsequent demographic grounding and persona simulation.

\subsubsection{CQ-Guided Ontology Relation Construction}
\label{sec:cq_ontology}
To model relationships between value categories, we adopt a human-guided ontology construction process based on Competency Questions (CQs)\footnote{Details of the ontology construction process and CQ examples are provided in Appendix~\ref{sec:ontology_details}, Table~\ref{tab:cq_examples}.}. Domain experts curate CQs, each designed to probe meaningful interactions between sub-classes of two given top-classes in the fixed taxonomy.
For each CQ, we prompt a Large Language Model to describe sub-class-level relationships between the given top-classes.
The model is constrained to:
(i) use only the predefined taxonomy classes,
(ii) avoid introducing any new classes, and
(iii) focus solely on articulating relationships between sub-classes of the given top-classes.

\paragraph{Ontology Triple generation under cultural conditioning.}
To incorporate diverse cultural perspectives during ontology construction, we condition the LLM on value profiles sampled from 120 individuals (20 per region) spanning six major world regions. Each CQ yields candidate relational statements, represented as ordered sentence triples:\footnote{Construction prompts are provided in Appendix~\ref{app:D}, Table~\ref{tab:object_property_prompt}.}
\begin{equation}
t_{a,b} = (c_a, p_{a,b}, c_b),
\end{equation}
where $c_a$ and $c_b$ are short noun phrases describing sub-classes from the two queried top-classes, and $p_{a,b}$ is a natural-language relation verb phrase (distinct from $p$ in top-$p$ selection). Although these correspond to ontology classes and object properties, we express them as natural language sentences to maintain human interpretability and ensure alignment with the phrasing of the Competency Questions. We use parentheses to emphasize an ordered ontology triple of text rather than symbolic identifiers.

\paragraph{Consolidation and human review.}
\begin{figure}[H]
    \centering
    \includegraphics[width=0.45\linewidth]{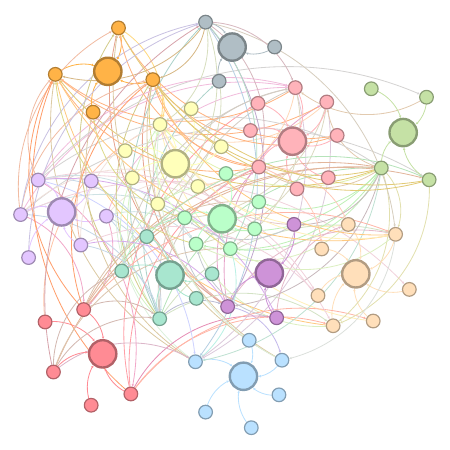}
    \hfill
    \includegraphics[width=0.45\linewidth]{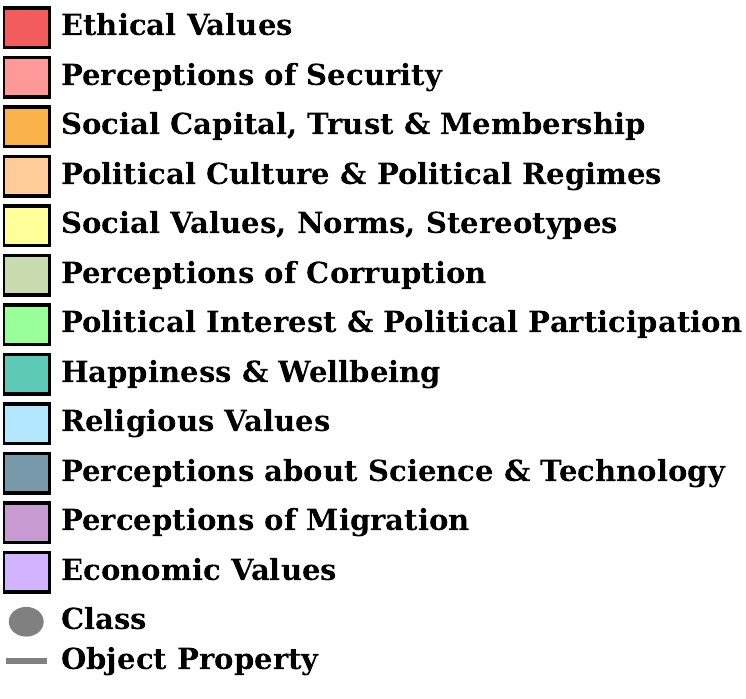}
    \caption{\textbf{Visualization of the final ontology structure.} The ontology consists of 12 top-classes and 64 sub-classes, connected by 150 object-property relations to form a comprehensive semantic network.}
    \label{fig:ontology_final_with_legend}
\end{figure}
Compared to traditional ontology engineering, the taxonomy remains fixed: no classes are merged, split, or added.
Human experts review candidate object properties by
(1) validating cultural plausibility,
(2) editing relation descriptions for clarity and consistency, and
(3) removing spurious or inconsistent relations\footnote{The collaborative construction process is detailed in Appendix~\ref{subsec:expert_validation}, and inter-group reliability in Appendix~\ref{subsubsec:ontology_analysis}.}.

The resulting ontology consists of a two-level class hierarchy. The 12 top-classes are $c_1, \dots, c_{12}$, and we write $c_{i,j}$ for the $j$-th sub-class under top-class $c_i$ (with $n_i$ sub-classes each). The complete taxonomy is
\begin{equation}
\begin{aligned}
C &= \{c_i\}_{i=1}^{12} \;\cup\; \{c_{i,j} \mid 1 \le i \le 12,\; 1 \le j \le n_i\}, \\
&\quad \textstyle\sum_{i=1}^{12} n_i = 64, \; |C| = 76,
\end{aligned}
\end{equation}
and a curated ontology triple set
\begin{equation}
T = \{ t_h \}_{h=1}^{|T|},\quad t_h = (c_a, p_{a,b}, c_b).
\end{equation}
Here $P$ denotes the set of relation texts, and we index the curated ontology triples as $\{t_h\}$, where each $t_h$ corresponds to an ontology triple of the form $t_{a,b}$ above.
Thus $T \subseteq \{c_{i,j}\} \times P \times \{c_{i,j}\}$ is the curated subset of CQ-derived cross-class relations, and $|T|$ is determined by expert review rather than by the combinatorics of $|\{c_{i,j}\}|$.

\subsection{\textbf{OG-MAR} Inference Pipeline}
\label{sec:OG-MAR_pipeline}

\subsubsection{Query Analysis \& Context Retrieval}
\label{sec:retrieval}

Given an input query $q$ and target respondent demographics $d_q$, we retrieve (i) ontology-consistent triples and (ii) demographically grounded respondent profiles, which jointly define the context for downstream multi-agent simulation. We use $D_q$ for the top-$k$ top-classes, $O_q$ for retrieved ontology triples, $R_q$ for retrieved respondents, and $\mathcal{V}_q$ for their value profiles. Here, $k$, $M$, and $K$ are the corresponding selection sizes.

\paragraph{(a) Topic/category identification.}
A Topic-Selection Model $G_{\text{topic}}$ identifies relevant top-classes.
We implement $G_{\text{topic}}$ as a pretrained text encoder fine-tuned on WVS data labeled with the 12 top-classes.
These selected top-classes determine the scope of sub-classes in subsequent ontology retrieval.

Let $D=\{c_1,\dots,c_{12}\}$ denote the set of top-classes. The encoder outputs a logit score $\ell_u$ for each $c_u$, and the top-$k$ top-classes are selected:
\begin{equation}
D_q = \{ c_u \mid \ell_u \text{ is among the top-}k \text{ scores of } \{\ell_1,\dots,\ell_{12}\} \}.
\end{equation}

\paragraph{(b) Ontology triple retrieval.}
We retrieve ontology knowledge in the form of triples, where each triple
$t_h = (c_a, p_{a,b}, c_b)$ is treated as a single semantic unit. In this step, we
\emph{only} use node-level similarity within the sub-classes of the selected domains $D_q$.
Specifically, for each sub-class $c \in \{c_{k,j} \mid c_k \in D_q\}$, we compute a node relevance:
\begin{equation}
\alpha(c) = \mathrm{sim}(\mathbf{e}_q, \mathbf{e}_c),
\end{equation}
where $\mathbf{e}_c$ is the embedding of the category text $c$.

We then score each ontology triple by the relevance of its endpoint nodes:
\begin{equation}
\alpha_{\text{triple}}(t_h) = \max\big(\alpha(c_a), \alpha(c_b)\big).
\end{equation}
Finally, we restrict retrieval to ontology triples whose endpoints are both sub-classes belonging to top-classes in $D_q$,
and select the top-$M$ ontology triples by $\alpha_{\text{triple}}$ to form the ontology context $O_q$.

\paragraph{(c) Similar individual retrieval with dense embeddings.}
To ground reasoning in real-world perspectives, we retrieve individuals demographically similar to $d_q$ using dense embedding retrieval. We encode demographic descriptions with an embedding model and rank respondents in $I$ by embedding similarity to $d_q$. The top-$K$ individuals form the demographic set $R_q$,
\begin{equation}
R_q = \{ i_1, i_2, \dots, i_K \},
\end{equation}
with corresponding value profiles
\begin{equation}
\mathcal{V}_q = \{ V_i \mid i \in R_q \}.
\end{equation}

\subsubsection{Multi-Value Persona Agent Simulation}
\label{sec:persona_simulation}
Given $\mathcal{V}_q$, we instantiate Value-Persona Agents that simulate culturally grounded reasoning under ontology constraints. We denote the agent as $G_{\text{persona}}$ and instantiate it per retrieved individual $i \in R_q$.
For convenience, index the retrieved ontology triples as $O_q = \{ t_h \}_{h=1}^{|O_q|}$ and let $C_q = \{ c_a, c_b \mid t_h \in O_q \}$ denote the set of sub-classes referenced by the retrieved ontology triples.
Concretely, each agent is conditioned on the ontology context $O_q$ (typically 3--9 triples), the individual's value summaries restricted to $C_q$, and demographic attributes $d_i$. We denote the filtered profile as $V_{i,q} = \{ s_i(c) \mid c \in C_q \}$.
The agent-specific conditioning context is:
\begin{equation}
z_i = \mathrm{Concat}(O_q, V_{i,q}, d_i).
\end{equation}
Here $z_i$ denotes the conditioning context for the persona associated with individual $i$.

\paragraph{Simulated reasoning trace.}
Given query $q$, each agent generates (i) an answer and (ii) an explicit simulated reasoning trace that forms a coherent chain of thought under the provided evidence and ontology constraint:
\begin{equation}
G_{\text{persona}}(q, z_{i}) = (\hat{y}_{i}, \rho_{i}).
\end{equation}
Here $\hat{y}_{i}$ denotes the agent's answer and $\rho_{i}$ denotes its natural-language reasoning trace.
We collect outputs across all retrieved individuals and ontology triples and denote the set by $A$:
\begin{equation}
A = \{ G_{\text{persona}}(q, z_{i}) \mid i\in R_q\ \}.
\end{equation}

\subsubsection{Ontology-Guided Final Judgment via Constrained Meta-Adjudication}
\label{sec:final_judgment}
A Final Judgment Agent $G_{\text{judge}}$ synthesizes the final prediction by performing
\emph{Constrained Meta-Adjudication} over the candidate outputs from Value-Persona Agents.
Given the multi-agent set
 $A$ and the query $q$ (question and response options),
the judge outputs:
\begin{equation}
\hat{y}_q = G_{\text{judge}}(A, q).
\end{equation}
Here $\hat{y}_q$ denotes the final prediction for query $q$. The judge does not receive $O_q$ or $\mathcal{V}_q$ directly; ontology and profile grounding are carried through the persona outputs.

Compared to majority voting, $G_{\text{judge}}$ uses a constrained, evidence-first protocol. The Judgment Agent follows an evidence-first adjudication protocol that prioritizes grounded support and ontology consistency before consulting vote signals:\footnote{Mechanism-level cases distinguishing OG-MAR from ensemble-style aggregation are provided in Appendix~\ref{app:case_not_ensemble}.}
\begin{enumerate}[leftmargin=*, itemsep=2pt, topsep=2pt]
    \item \textbf{Evidence \& consistency.} For each $(\hat{y}_i,\rho_i)$, score grounding of $\rho_i$ and ontology compliance, then aggregate scores per option $y$.
    \item \textbf{Vote as secondary cue.} If the leading options have comparable evidence strength, consult a Vote summary as a secondary signal (otherwise ignore it).
    \item \textbf{Relevance tie-break.} If still tied, choose the option supported by personas more relevant to $d_q$.
\end{enumerate}
This adjudication is conducted within a single LLM call, with the above criteria guiding the judge's internal reasoning rather than operating as separate rule-based modules.

\subsection{Implementation Details}
\label{sec:impl_details}
Our implementation consists of: (i) a fixed ontology \textbf{taxonomy} with 12 top-classes and 64 sub-classes; (ii) a \textbf{Summarization Agent $G_{\text{sum}}$}, an LLM-based summarizer that generates category-specific summaries $s_i(c)$ from raw respondent responses $\mathcal{R}_i$ under a ``no-new-concepts'' constraint; (iii) a \textbf{Topic-Selection Agent $G_{\text{topic}}$}, a pretrained text encoder fine-tuned on WVS with top-class supervision, which selects the top-$k$ top-classes; (iv) a \textbf{category selection} policy that retains the top-3 fine-grained value categories per query; (v) an \textbf{ontology triple retrieval} step that performs dense retrieval over ontology triples for each selected category and returns up to the top-3 triples per category; and (vi) a \textbf{persona retrieval setting} where the default number of retrieved individuals (personas) is set to $K{=}5$.



\section{Experimental Design and Setup}
\subsection{Setup}
\paragraph{Models Used.} We use GPT-4o-mini~\citep{achiam2023gpt} and Gemini 2.5~\citep{google2025gemini25flashlite} via APIs, and Qwen 2.5~\citep{qwen2.5} and EXAONE 3.5~\citep{an2024exaone} as open-source models for the generation task in both our Persona Agent and Final Judgment Agent. We also used GPT-o4-mini~\citep{openai2025o4} for object properties construction and values profile generation. To ensure stable behavior, we set the temperature to 0 across all models.

We use dense embedding retrieval with E5-base embeddings~\cite{wang2022text}. For demographic retrieval, we encode the target demographic description and each respondent’s demographic profile, then rank respondents by embedding similarity to obtain the top-$K$ demographically similar individuals. For ontology retrieval, we embed ontology triples and retrieve the top-$M$ triples by similarity to the query.

Additionally, for topic classification, we fine-tune DeBERTa-v2-xxlarge~\citep{he2020deberta} on WVS data for 3 epochs (batch size=4, learning rate=$5 \times 10^{-6}$), following optimized configurations for value identification~\citep{kiesel2023semeval, balikas2023john}\footnote{Detailed training analyses are provided in Appendix~\ref{app:deberta_training}.}.

\subsection{Datasets}

\paragraph{Retrieval Corpus}
We use the \textit{World Values Survey} (WVS)~\citep{haerpfer2022world} as the retrieval corpus. WVS is a large-scale cross-national survey of human values and socio-cultural attitudes with structured demographic attributes (e.g., country, age, gender, education). In our setting, we use predefined 12 topics in the WVS\footnote{The 12 topics are provided in Appendix~\ref{app:dataset_details}, Table~\ref{tab:wvs_values_topics}.}, which provide a globally diverse and publicly available source of value-related responses and enable consistent retrieval of relevant demographic evidence for downstream inference.

\paragraph{Test Datasets}
To evaluate generalization beyond the retrieval corpus, we use six regional social-survey datasets with value-related questions and WVS-comparable demographic metadata: \textbf{EVS} (Europe), \textbf{GSS} (U.S.), \textbf{CGSS} (China), \textbf{ISD} (India), \textbf{LAPOP} (Latin America and the Caribbean), and \textbf{Afrobarometer} (Africa). We use clustering-based sampling to select 2{,}000 representative instances as test data, for efficient yet balanced evaluation across diverse datasets\footnote{Detailed dataset descriptions, sampling process, and an additional post-cutoff experiment are provided in Appendix~\ref{app:dataset_details}.}.

\subsection{Evaluation Metrics}
We evaluate model performance using both item-level and distribution-level metrics. For item-level evaluation, we report binary accuracy, where each prediction is scored as correct or incorrect after applying the corresponding response-type rule. This yields a unified 0/1 correctness criterion across datasets while retaining item-specific answer formats. We use four response types:
\begin{itemize}
    \item \textbf{Two-choice binary items}: two-option questions, evaluated by exact matching between the predicted label and the gold label.
    
    \item \textbf{Derived binary items}: multiple-choice questions mapped into two fixed semantic buckets; a prediction is correct if the predicted and gold labels fall into the same bucket.
    
    \item \textbf{Ordinal items}: ordered response items. For accuracy, raw responses are binarized using the midpoint of the original scale, and compared with the gold label.
    
    \item \textbf{Nominal exact-match items}: unordered multi-choice questions, evaluated by exact matching between the predicted label and the gold label.
\end{itemize}

For ordinal items, we additionally report Mean Absolute Error (MAE) on the valid ordinal scale:
\begin{equation}
\mathrm{MAE}=\frac{1}{N}\sum_{i=1}^{N}\left|y_i-\hat{y}_i\right|.
\label{eq:mae}
\end{equation}
Here, $y_i$ and $\hat{y}_i$ denote the gold and predicted ordinal values after applying the item-specific valid response range.\footnote{Average MAE scores across regional datasets are shown in Figure~\ref{fig:eval_token}.}

For distribution-level evaluation, we compare the predicted and gold survey response distributions using EMD, Proportion Correlation, and TVD.\footnote{Detailed evaluation results are provided in Appendix~\ref{app:evaluation_details}.}

\subsection{Baselines}
We compare against two single-pass prompting baselines, one single-agent aggregation baseline, and two multi-step baselines spanning multi-agent deliberation and retrieval augmentation:
(1) \textbf{\textit{Zero-shot}}, a single prompt without scaffolding;
(2) \textbf{\textit{Role Assignment}}~\citep{tao2024cultural}, which conditions generation on an explicit culturally grounded role;
(3) \textbf{\textit{Self-consistency}}~\citep{wang2022self}, which samples multiple reasoned outputs from the same model and takes a majority vote over them;
(4) \textbf{\textit{Debate}}~\citep{ki2025multiple}, a multi-agent framework with iterative critique and refinement;
and (5) \textbf{\textit{ValuesRAG}}~\citep{seo2025valuesrag}, which grounds generation in retrieved survey evidence.
For a fair comparison, we use the same retrieval setting for \textit{ValuesRAG} in the main experiments\footnote{Detailed baseline explanations and an additional post-training comparison are provided in Appendix~\ref{app:baseline_details}.}.

\section{Experimental Results}

Table~\ref{tab:main_results} reports results on six regional benchmarks and four Large Language Model backbones, showing that baselines offer only incremental gains and fail under cultural distribution shift. \textbf{Zero-shot} Inference is efficient but often reverts to culture-default priors, yielding inconsistent value-sensitive judgments across regions. \textbf{Role Assignment} adds culturally framed prompting, yet remains weakly grounded in empirically observed value distributions and is sensitive to prompt formulation. \textbf{Self-Consistency} improves robustness via sample aggregation, but it does not enforce demographic grounding or conceptual coherence and may amplify majority bias. \textbf{Debate} introduces critique-and-refinement, but without explicit evidence constraints it is prone to drift and still fails to capture cross-topic value dependencies. \textbf{ValuesRAG} is competitive by grounding generation in survey evidence but treats values as unstructured snippets, which restricts control over structured relationships.

In contrast, \textbf{OG-MAR} integrates ontology-guided triple retrieval and multi-persona simulation with a judgment agent that explicitly weighs ontology consistency and demographic proximity. This design yields strong and consistent performance across regions, achieving average accuracies of 0.6308 on Gemini 2.5 Flash Lite, 0.5705 on Qwen 2.5, and 0.6317 on EXAONE, while remaining competitive on GPT-4o-mini with an average accuracy of 0.6007. Notably, \textbf{OG-MAR} delivers particularly large gains on culturally challenging settings such as CGSS and ISD, suggesting that structured cultural relations and demographically grounded personas are most beneficial when the target distribution deviates from dominant pretraining priors.

\begin{table*}[t]
\caption{\textbf{Accuracy of baseline methods across regional datasets.} \textbf{Bold} text indicates the best performance, \underline{underlined} text the second-best performance. $\ast$ denotes significant improvements (paired $t$-test with Holm--Bonferroni correction, $p<0.05$) over all baselines. $\dagger$ denotes our proposed method.}
\centering
\scriptsize
\setlength{\tabcolsep}{6.6pt}
\renewcommand{\arraystretch}{1.08}
\begin{tabular}{lccccccc}
\toprule
\textbf{Method} & \textbf{EVS \textit{(Europe)}} & \textbf{GSS \textit{(United States)}} & \textbf{CGSS \textit{(China)}} & \textbf{ISD \textit{(India)}} & \textbf{AFRO \textit{(Africa)}} & \textbf{LAPOP \textit{(Latin America)}} & \textbf{Avg.} \\
\hline

\addlinespace[2pt]
\multicolumn{8}{>{\columncolor{gray!15}}l}{\textbf{GPT-4o mini}} \\
Zero-shot  & 0.5606 & 0.5164 & 0.5847 & 0.6139 & 0.5324 & 0.5760 & 0.5640 \\
Role~(\citeyear{tao2024cultural})       & 0.5892 & 0.5184 & \underline{0.6014} & 0.6060 & \underline{0.5505} & 0.5674 & 0.5722 \\
Self-consistency~(\citeyear{wang2022self})      & 0.5558 & 0.4920 & 0.5631 & 0.5976 & 0.5224 & 0.5551 & 0.5477 \\
Debate~(\citeyear{ki2025multiple})     & 0.5985 & \underline{0.5509} & 0.5993 & \textbf{0.6568} & 0.5343 & 0.5306 & 0.5784 \\
ValuesRAG~(\citeyear{seo2025valuesrag})  & \underline{0.6127} & \textbf{0.5589} & 0.5889 & \underline{0.6420} & \textbf{0.5654} & \underline{0.6085} & \underline{0.5961} \\
\textbf{OG-MAR} (Ours)$^\dagger$ & \textbf{0.6206}* & 0.5480 & \textbf{0.6509}* & 0.6192 & 0.5389 & \textbf{0.6268} & \textbf{0.6007*} \\
\hline
\addlinespace[2pt]
\multicolumn{8}{>{\columncolor{gray!15}}l}{\textbf{Gemini 2.5 Flash Lite}} \\
Zero-shot  & 0.5681 & 0.4957 & 0.6467 & 0.5000 & 0.5282 & 0.6225 & 0.5602 \\
Role~(\citeyear{tao2024cultural})         & 0.5786 & 0.4992 & \underline{0.6669} & 0.5521 & 0.5313 & 0.5852 & 0.5689 \\
Self-consistency~(\citeyear{wang2022self})      & 0.5489 & 0.4728 & 0.6063 & 0.4705 & 0.5182 & \underline{0.6268} & 0.5406 \\
Debate~(\citeyear{ki2025multiple})    & 0.5977 & 0.5138 & 0.6348 & \underline{0.6335} & 0.5046 & 0.5331 & 0.5696 \\
ValuesRAG~(\citeyear{seo2025valuesrag})  & \underline{0.6075} & \underline{0.5376} & 0.6084 & 0.6041 & \underline{0.5472} & 0.5339 & \underline{0.5731} \\
\textbf{OG-MAR} (Ours)$^\dagger$       & \textbf{0.6249}* & \textbf{0.5489}* & \textbf{0.7017}* & \textbf{0.7007}* & \textbf{0.5701}* & \textbf{0.6385}* & \textbf{0.6308*} \\
\hline

\addlinespace[2pt]
\multicolumn{8}{>{\columncolor{gray!15}}l}{\textbf{QWEN 2.5}} \\
Zero-shot  & 0.5199 & 0.5069 & 0.2704 & \underline{0.7222} & 0.4814 & 0.4908 & 0.4986 \\
Role~(\citeyear{tao2024cultural})        & 0.5357 & 0.5037 & 0.3463 & \textbf{0.7452} & \underline{0.5014} & 0.4712 & 0.5172 \\
Self-consistency~(\citeyear{wang2022self})      & 0.5096 & 0.4975 & 0.3289 & 0.6278 & 0.4080 & 0.4975 & 0.4782 \\
Debate~(\citeyear{ki2025multiple})     & 0.5511 & 0.5174 & 0.4578 & 0.6320 & 0.4875 & 0.4332 & 0.5132 \\
ValuesRAG~(\citeyear{seo2025valuesrag})  & \underline{0.5538} & \underline{0.5215} & \underline{0.4697} & 0.6591 & 0.4724 & \underline{0.5268} & \underline{0.5339} \\
\textbf{OG-MAR} (Ours)$^\dagger$       & \textbf{0.5898}* & \textbf{0.5325}* & \textbf{0.5220}* & 0.6599 & \textbf{0.5180} & \textbf{0.6005} & \textbf{0.5705*} \\
\hline

\addlinespace[2pt]
\multicolumn{8}{>{\columncolor{gray!15}}l}{\textbf{EXAONE 3.5}} \\
Zero-shot  & 0.5143 & 0.5311 & 0.2885 & 0.6041 & 0.4054 & 0.5006 & 0.4740 \\
Role~(\citeyear{tao2024cultural})        & 0.5319 & 0.5326 & 0.3129 & 0.6048 & 0.4077 & 0.4602 & 0.4750 \\
Self-consistency~(\citeyear{wang2022self})     & 0.5490 & 0.5266 & 0.2697 & 0.6122 & 0.4086 & 0.5368 & 0.4838 \\
Debate~(\citeyear{ki2025multiple})     & \underline{0.5713} & 0.5407 & 0.5624 & \underline{0.6773} & \underline{0.4995} & 0.4939 & 0.5575 \\
ValuesRAG~(\citeyear{seo2025valuesrag})  & 0.5172 & \underline{0.5520} & \underline{0.5833} & 0.6446 & 0.4794 & \underline{0.5913} & \underline{0.5613} \\
\textbf{OG-MAR} (Ours)$^\dagger$       & \textbf{0.6080}* & \textbf{0.5636} & \textbf{0.6307}* & \textbf{0.7810}* & \textbf{0.5045}* & \textbf{0.7022}* & \textbf{0.6317*} \\
\bottomrule
\end{tabular}
\label{tab:main_results}
\end{table*}

\subsection{Ablation Studies}
\subsubsection{\textbf{Varying the Number of Retrieved Individuals}} 
\begin{figure}[t]
    \centering
    \includegraphics[width=0.7\linewidth]{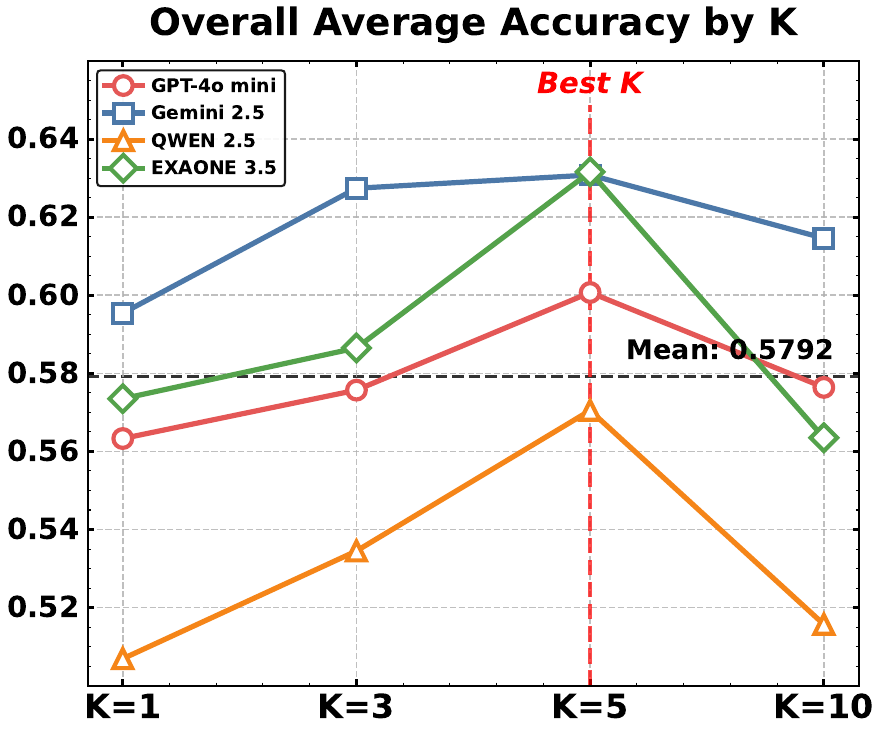}
    \caption{\textbf{Performance comparison of four models across $K \in \{1, 3, 5, 10\}$ on average.} Red vertical dashed lines indicate the best $K$ and gray horizontal lines show the overall mean accuracy.}
    \label{fig:k_ablation}
\end{figure}

We investigate the impact of retrieval size $K$ on \textbf{OG-MAR}'s performance by varying the number of retrieved demographically similar individuals across $K \in \{1, 3, 5, 10\}$. Figure~\ref{fig:k_ablation} reports average accuracy for GPT-4o-mini, Gemini 2.5 Flash Lite, QWEN 2.5, and EXAONE 3.5.

All four models achieve their best overall performance at $K{=}5$, outperforming other retrieval sizes by +0.003 to +0.07 across different models. While Gemini 2.5 Flash Lite shows minimal difference between $K{=}3$ and $K{=}5$, the remaining models exhibit substantially larger improvements from $K{=}3$ to $K{=}5$, with gains ranging from 0.03 to 0.05. When retrieval size increases to $K{=}10$, all models show clear performance degradation, with accuracy drops ranging from 0.02 to 0.07 compared to $K{=}5$. These results reveal a clear trade-off: $K{=}1$ retrieves narrow value-persona signals, whereas $K{=}5$ offers richer yet stable context. Consequently, we adopt $K{=}5$ as the default retrieval depth throughout our experiments\footnote{Complete results for all regional datasets are in Appendix~\ref{app:k-ablation-full}.}.

\subsubsection{\textbf{Impact of Value Inference Generation}}
\label{sec:value_inference_generation}
\begin{figure}[t]
    \centering
    \includegraphics[width=0.8\linewidth]{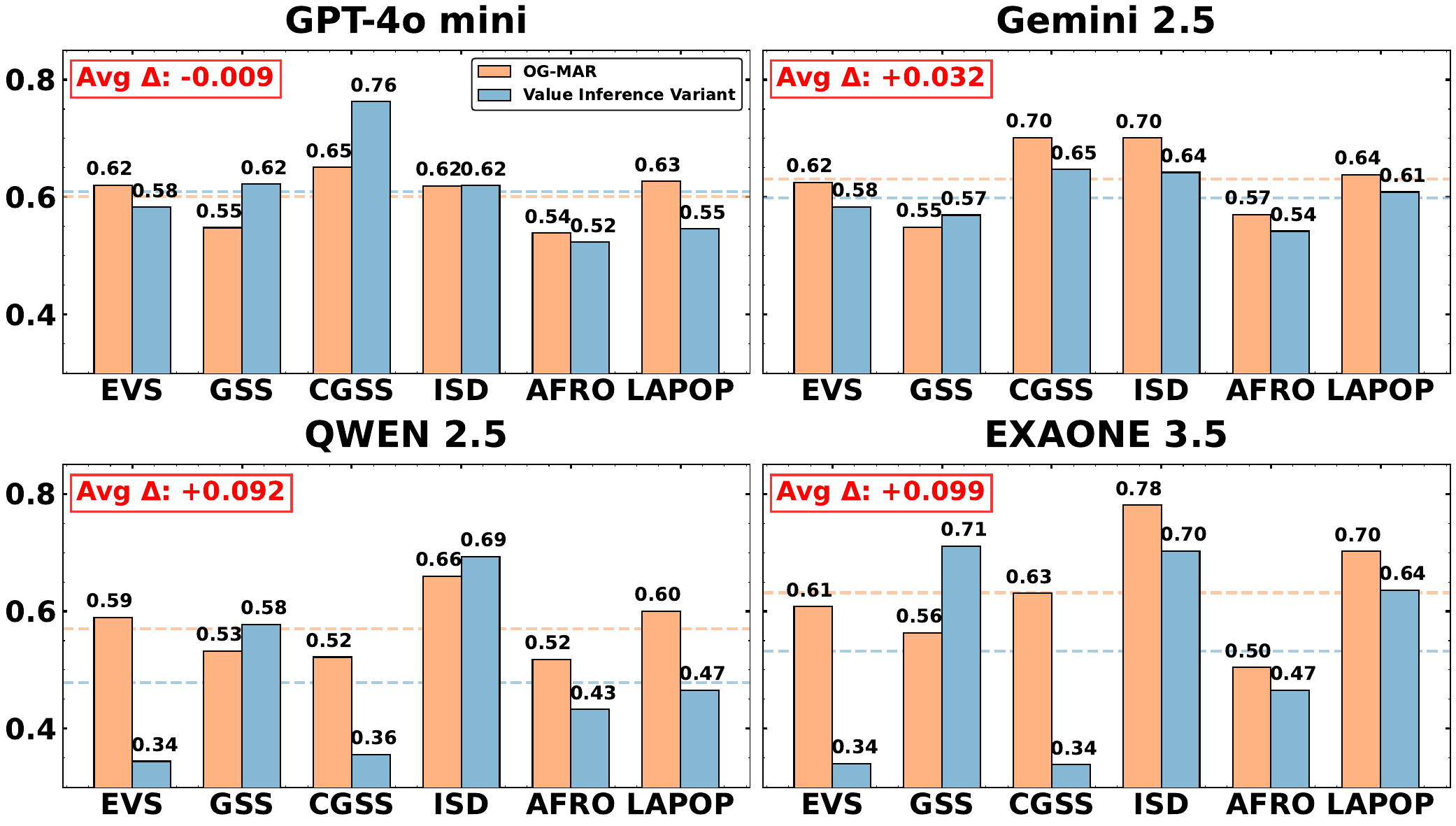}
    \caption{\textbf{Performance comparison between \textbf{OG-MAR} and the Value Inference Variant.} Accuracy over four models on six regional datasets. Dashed lines show per-method average accuracy; red boxes report the average gap (Avg $\Delta$ = \textbf{OG-MAR} - Variant).}
    \label{fig:value_inference_comparison}
\end{figure}

To assess the impact of value inference generation, we compare our two-step process (with $K$ personas and a judgment agent) against a similar two-step process where a single Value Inference Agent replaces the $K$ persona agents by first inferring a value profile for the target individual, after which the same judgment agent answers questions based only on this profile. We evaluate both approaches across all six regional datasets using the same retrieval and setup as in our main experiments. Figure~\ref{fig:value_inference_comparison} presents the accuracy comparison between the two architectures.

Among the four models, GPT-4o-mini slightly benefits from the Value Inference Variant by around 0.01 on average, particularly showing higher accuracy on GSS (+0.07) and CGSS (+0.11). In contrast, the other three models achieve consistent gains with \textbf{OG-MAR}, ranging from +0.03 to +0.10 on average across datasets. While the Value Inference Variant outperforms \textbf{OG-MAR} on GSS across all four models, \textbf{OG-MAR} achieves clearly higher accuracy on the remaining datasets 
for most models. These results suggest that explicitly simulating multiple personas and preserving their distinct value profiles provides the judgment model with richer and more diverse evidence\footnote{Appendix~\ref{app:case_multipersona} provides a qualitative case of preserving competing cultural cues through multi-persona simulation.}, which in turn allows \textbf{OG-MAR} to maintain stronger performance than the Value Inference Variant in most models and datasets. 

\begin{table}[t]
\centering
\footnotesize
\setlength{\tabcolsep}{6pt}
\caption{\textbf{Average accuracy of the full \textbf{OG-MAR} framework and the single-judge variant across four LLMs.}}
\label{tab:ablation_single_judge}
\begin{tabular}{lcc}
\toprule
\textbf{Model} & \textbf{Method} & \textbf{Avg. Accuracy} \\
\midrule
\multirow{2}{*}{GPT-4o mini} 
    & \textbf{OG-MAR}         & \textbf{0.6007} \\
    & Single-Judge & 0.5987 \\
\midrule
\multirow{2}{*}{Gemini 2.5} 
    & \textbf{OG-MAR}         & \textbf{0.6308} \\
    & Single-Judge & 0.6022 \\
\midrule
\multirow{2}{*}{QWEN 2.5} 
    & \textbf{OG-MAR}         & \textbf{0.5705} \\
    & Single-Judge & 0.5311 \\
\midrule
\multirow{2}{*}{EXAONE 3.5} 
    & \textbf{OG-MAR}         & \textbf{0.6317} \\
    & Single-Judge & 0.5627 \\
\bottomrule
\end{tabular}
\end{table}

\subsubsection{\textbf{Impact of Multi-Persona Reasoning}}

\label{app:single_judge}
We compare the full \textbf{OG-MAR} framework with a single-judge variant that skips persona simulation and directly generates the final answer from the same inputs ($K$ individuals’ demographics, value summaries, and ontology context). Using the setup of Section~\ref{sec:value_inference_generation}, we evaluate both variants on all datasets; Table~\ref{tab:ablation_single_judge} reports the resulting accuracies.

Overall, \textbf{OG-MAR} achieves higher average accuracy across all four models (+0.002 on GPT-4o mini, +0.03 on Gemini 2.5 Flash Lite, +0.04 on QWEN 2.5, and +0.07 on EXAONE 3.5). Although the single-judge variant is consistently better on GSS, \textbf{OG-MAR} improves most other datasets\footnote{Detailed per-dataset results are provided in Appendix~\ref{app:multi-persona-full}.}. These results suggest that persona simulation is a meaningful contributor to performance, particularly when regional benchmarks require reconciling multiple, potentially competing value considerations. This effect is especially pronounced when the input demographic set induces diverse value profiles. At the same time, the single-judge model remains competitive, implying that \textbf{OG-MAR}’s gains are not driven solely by the simulation layer. Instead, improvements also stem from the shared ontology-grounded retrieval and value summarization pipeline, which provides structured, survey-backed evidence for downstream reasoning\footnote{A component-level ablation study is provided in Appendix~\ref{app:component_ablation}.}.

\begin{figure}[!t]
\centering
\includegraphics[width=0.8\columnwidth]{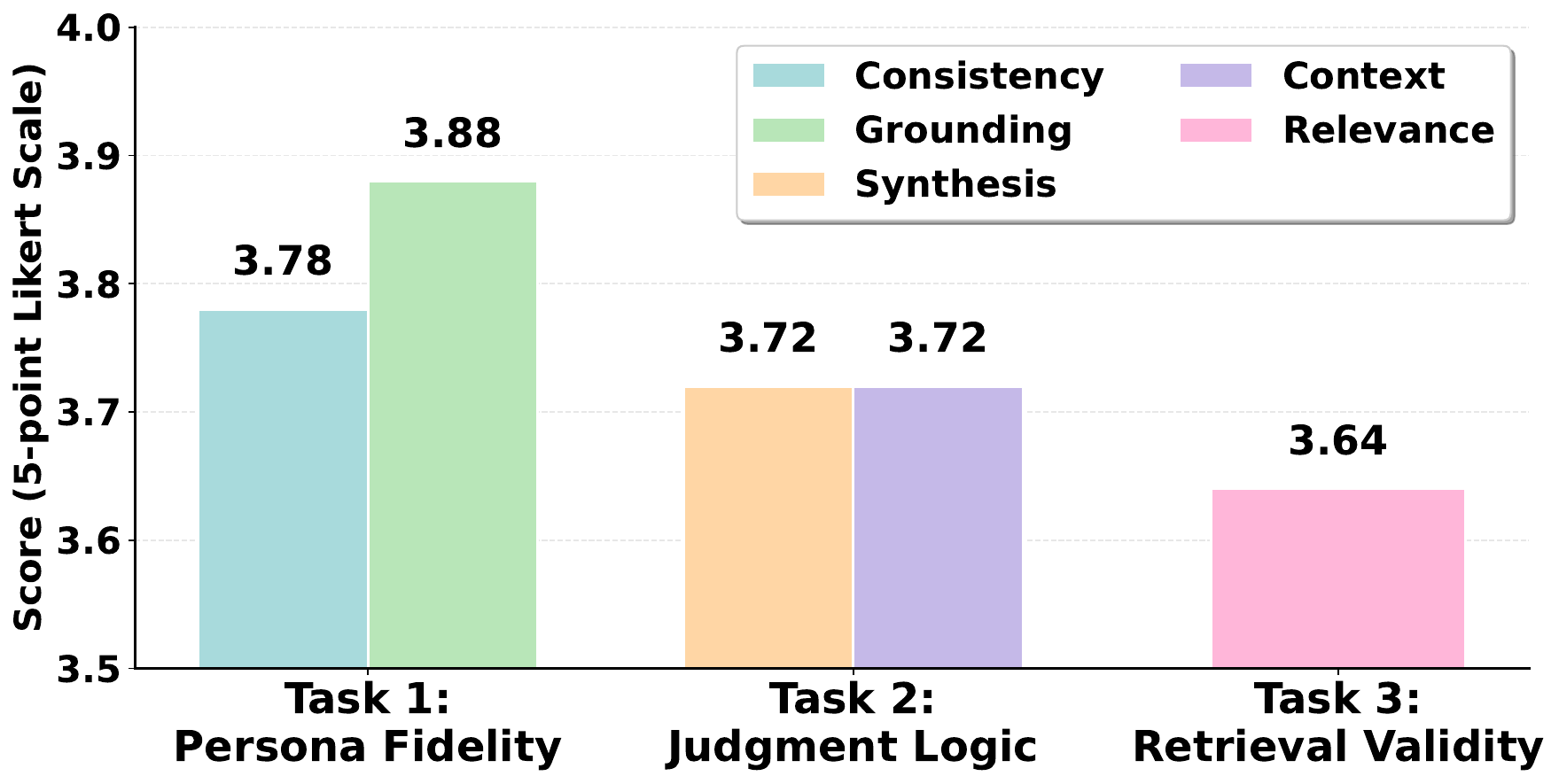}
\caption{\textbf{Average human evaluation scores (5-point Likert scale) across three tasks:} Persona Fidelity (Consistency, Grounding), Judgment Logic (Synthesis, Context), and Retrieval Validity (Relevance). Scores are averaged over nine expert raters.}
\label{fig:human_evaluation}
\end{figure}

\section{Discussion}

\begin{figure}[!t]
\centering
\includegraphics[width=\columnwidth]{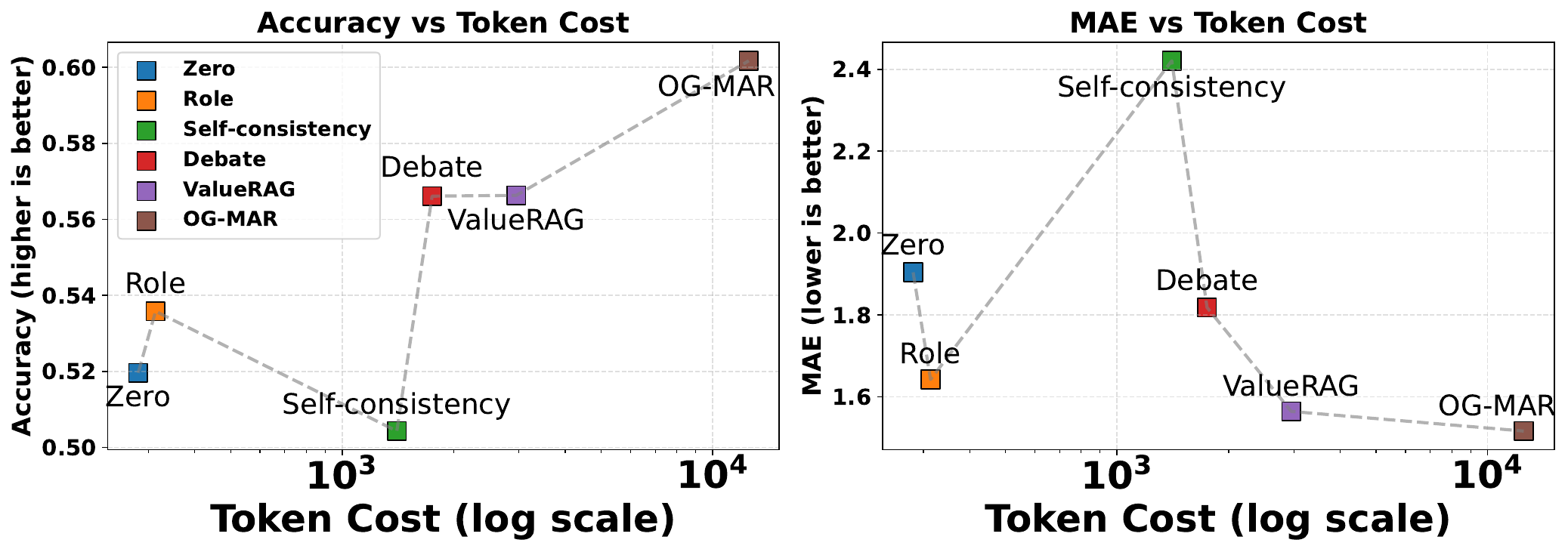}
\caption{\textbf{Performance--cost trade-off across methods.} Left: accuracy vs total tokens (higher is better). Right: MAE vs total tokens (lower is better). Markers denote methods; the dashed line shows performance changes as token usage increases.}
\label{fig:eval_token}
\end{figure}

\label{app: discussion}

\paragraph{Qualitative Analysis} 
55To complement quantitative results, we conducted a human evaluation of \textbf{OG-MAR}'s reasoning traces with nine domain experts using three 5-point Likert tasks: \textit{Persona Fidelity} (Task 1), \textit{Judgment Logic} (Task 2), and \textit{Retrieval Validity} (Task 3). 

Overall, \textbf{OG-MAR} shows consistent interpretability across six regional datasets. Notably, it achieves the highest \textit{Grounding} on \textbf{CGSS (China)} (4.02), slightly exceeding \textbf{GSS (U.S.)} (3.97), suggesting ontology-guided value injection mitigates ``culture-default'' tendencies by encouraging evidence-based reasoning. The Judgment Agent also attains strong \textit{Synthesis Logic} (Avg.\ 3.72), indicating evidence-first adjudication remains effective across culturally distinct regions, while ontology retrieval maintains high relevance (Avg.\ 3.64), supporting that retrieved triples provide meaningful evidence for downstream reasoning. 

Qualitatively, experts frequently noted that persona rationales remained consistent with the retrieved value summaries and that final decisions explicitly reconciled conflicting value signals rather than defaulting to generic responses. The case studies further illustrate how evidence-first adjudication can correct or down-weight noisy persona claims when they are not supported by ontology-linked evidence. Remaining failures are typically attributable to sparse ontology coverage or imperfect demographic retrieval, which can limit grounding even when the judgment logic is coherent.\footnote{Detailed qualitative evaluation tasks, procedures, and results are provided in Appendix~\ref{app:human_eval}; qualitative case studies, including evidence-first adjudication cases, are provided in Appendix~\ref{sec:case_study}.}

\paragraph{Performance--Token Usage Trade-off.}
Figure~\ref{fig:eval_token} shows the trade-off between token cost and prediction quality across methods. We measure cost by the total number of input and output tokens, and use a log-scale x-axis to highlight differences in usage. Performance is reported with Accuracy (higher is better) and MAE (lower is better). Single-pass prompting baselines lie in the lowest-cost regime but generally achieve weaker performance. RAG method baseline (ValuesRAG) and multi-step methods that retrieve external context or aggregate multiple samples consume substantially more tokens and typically improve quality. 

\textbf{OG-MAR} incurs the highest token budget, however, this cost reflects its intended role: not as a token-efficient replacement for lightweight prompting, but as a structured reasoning framework that organizes cultural values through ontology-guided relations, explicitly separates cultural dimensions, and supports transparent multi-agent adjudication. Its higher accuracy and lower average MAE therefore suggest that the additional computation contributes to more grounded and interpretable cultural reasoning, rather than merely adding inference overhead.

\section{Conclusion}
In this paper, we presented \textbf{OG-MAR}, an ontology-guided multi-agent framework for culturally aligned LLM inference. \textbf{OG-MAR} integrates CQ-driven cultural ontology construction with demographically grounded retrieval from WVS, and combines value-persona simulation with a judgment agent that enforces ontology consistency. Across six regional benchmarks and four LLM backbones, \textbf{OG-MAR} improves cultural alignment and robustness over competitive baselines. Beyond quantitative gains, our qualitative and ablation analyses help explain when and why \textbf{OG-MAR} works, highlighting the complementary roles of ontology-grounded retrieval, value summarization, and multi-persona reasoning in producing more interpretable outputs.

\section{Limitations}
While our results show that \textbf{OG-MAR} improves cultural alignment across diverse regions and LLM backbones, it does not guarantee optimal behavior for every query, demographic subgroup, or survey domain. Also, the performance can be sensitive to the underlying model, prompt context, and retrieval noise. In particular, cross-domain generalization remains limited: regional surveys are not always fully aligned with the WVS-derived taxonomy, and topic selection may fail when questions use domain-specific phrasing or reflect culturally local constructs. 

Moreover, demographic retrieval may miss minority or underrepresented profiles, and imperfect ontology triples can propagate structured but inaccurate constraints into downstream reasoning. The multi-agent simulation also increases inference cost and latency relative to single-pass baselines. 

Future work will address these limitations by: (1) developing more robust verification mechanisms for ontology triple accuracy and consistency; (2) improving cross-domain topic classification and uncertainty-aware retrieval to better cover underrepresented groups and survey-specific phrasing; and (3) designing efficient routing or pruning techniques within the multi-agent simulation to reduce computational cost without sacrificing alignment quality.

\section*{Impact Statement}
While \textbf{OG-MAR} aims to reduce cultural misalignment, its demographic retrieval and survey-grounded cultural signals raise ethical risks related to profiling, bias, and fairness. In high-stakes settings, demographic-conditioned outputs may reinforce stereotypes, encode survey artifacts, or amplify historical biases. \textbf{OG-MAR} is intended to surface value considerations for scrutiny and auditing, not to justify demographic generalizations.

Researchers and practitioners should evaluate performance across subpopulations (including minority and underrepresented groups), monitor for harmful pattern reproduction, and apply strong privacy and data-governance safeguards (including protections against re-identification). We recommend using \textbf{OG-MAR} only within accountable oversight workflows; future work should study uncertainty communication, debiasing, and safe deployment protocols.

\section*{Code and Model Availability}
Code and sample data are available in the \href{https://anonymous.4open.science/r/OG-MAR-Toward-Culturally-Aligned-LLMs-through-Ontology-Guided-Multi-Agent-Reasoning-50B1/README.md}{project repository}. The repository also includes scripts, prompts, and instructions to reproduce the main experiments. The topic classifier is available on \href{https://huggingface.co/CulturalAlignmentPKU/WVS-DeBERTa-v2-xxlarge}{Hugging Face}.

\section*{Acknowledgements}
This work was supported in part by internal funding from AI Research and Enhans AI, and by the Major Project of the National Social Science Foundation of China (Grant No. 24ZDA078).

\nocite{langley00}

\bibliography{example_paper}

@article{achiam2023gpt,
  title={Gpt-4 technical report},
  author={Achiam, Josh and Adler, Steven and Agarwal, Sandhini and Ahmad, Lama and Akkaya, Ilge and Aleman, Florencia Leoni and Almeida, Diogo and Altenschmidt, Janko and Altman, Sam and Anadkat, Shyamal and others},
  journal={arXiv preprint arXiv:2303.08774},
  year={2023}
}

@misc{jiang2023mistral7b,
      title={Mistral 7B}, 
      author={Albert Q. Jiang and Alexandre Sablayrolles and Arthur Mensch and Chris Bamford and Devendra Singh Chaplot and Diego de las Casas and Florian Bressand and Gianna Lengyel and Guillaume Lample and Lucile Saulnier and Lélio Renard Lavaud and Marie-Anne Lachaux and Pierre Stock and Teven Le Scao and Thibaut Lavril and Thomas Wang and Timothée Lacroix and William El Sayed},
      year={2023},
      eprint={2310.06825},
      archivePrefix={arXiv},
      primaryClass={cs.CL},
      url={https://arxiv.org/abs/2310.06825}, 
}

@article{dodge2021documenting,
  title={Documenting large webtext corpora: A case study on the colossal clean crawled corpus},
  author={Dodge, Jesse and Sap, Maarten and Marasovi{\'c}, Ana and Agnew, William and Ilharco, Gabriel and Groeneveld, Dirk and Mitchell, Margaret and Gardner, Matt},
  journal={arXiv preprint arXiv:2104.08758},
  year={2021}
}

@inproceedings{bender2021dangers,
  title={On the dangers of stochastic parrots: Can language models be too big?},
  author={Bender, Emily M and Gebru, Timnit and McMillan-Major, Angelina and Shmitchell, Shmargaret},
  booktitle={Proceedings of the 2021 ACM conference on fairness, accountability, and transparency},
  pages={610--623},
  year={2021}
}

@article{touvron2023llama,
  title={Llama 2: Open foundation and fine-tuned chat models},
  author={Touvron, Hugo and Martin, Louis and Stone, Kevin and Albert, Peter and Almahairi, Amjad and Babaei, Yasmine and Bashlykov, Nikolay and Batra, Soumya and Bhargava, Prajjwal and Bhosale, Shruti and others},
  journal={arXiv preprint arXiv:2307.09288},
  year={2023}
}

@article{gallegos2024bias,
  title={Bias and fairness in large language models: A survey},
  author={Gallegos, Isabel O and Rossi, Ryan A and Barrow, Joe and Tanjim, Md Mehrab and Kim, Sungchul and Dernoncourt, Franck and Yu, Tong and Zhang, Ruiyi and Ahmed, Nesreen K},
  journal={Computational Linguistics},
  volume={50},
  number={3},
  pages={1097--1179},
  year={2024},
  publisher={MIT Press 255 Main Street, 9th Floor, Cambridge, Massachusetts 02142, USA~…}
}

@article{xie2024can,
  title={Can large language model agents simulate human trust behavior?},
  author={Xie, Chengxing and Chen, Canyu and Jia, Feiran and Ye, Ziyu and Lai, Shiyang and Shu, Kai and Gu, Jindong and Bibi, Adel and Hu, Ziniu and Jurgens, David and others},
  journal={Advances in neural information processing systems},
  volume={37},
  pages={15674--15729},
  year={2024}
}

@inproceedings{pistilli2024civics,
  title={Civics: Building a dataset for examining culturally-informed values in large language models},
  author={Pistilli, Giada and Leidinger, Alina and Jernite, Yacine and Kasirzadeh, Atoosa and Luccioni, Alexandra Sasha and Mitchell, Margaret},
  booktitle={Proceedings of the AAAI/ACM Conference on AI, Ethics, and Society},
  volume={7},
  pages={1132--1144},
  year={2024}
}

@article{yoo2024hyperclova,
  title={Hyperclova x technical report},
  author={Yoo, Kang Min and Han, Jaegeun and In, Sookyo and Jeon, Heewon and Jeong, Jisu and Kang, Jaewook and Kim, Hyunwook and Kim, Kyung-Min and Kim, Munhyong and Kim, Sungju and others},
  journal={arXiv preprint arXiv:2404.01954},
  year={2024}
}

@inproceedings{nguyen2024seallms,
  title={SeaLLMs-large language models for Southeast Asia},
  author={Nguyen, Xuan-Phi and Zhang, Wenxuan and Li, Xin and Aljunied, Mahani and Hu, Zhiqiang and Shen, Chenhui and Chia, Yew Ken and Li, Xingxuan and Wang, Jianyu and Tan, Qingyu and others},
  booktitle={Proceedings of the 62nd Annual Meeting of the Association for Computational Linguistics (Volume 3: System Demonstrations)},
  pages={294--304},
  year={2024}
}

@article{sengupta2023jais,
  title={Jais and jais-chat: Arabic-centric foundation and instruction-tuned open generative large language models},
  author={Sengupta, Neha and Sahu, Sunil Kumar and Jia, Bokang and Katipomu, Satheesh and Li, Haonan and Koto, Fajri and Marshall, William and Gosal, Gurpreet and Liu, Cynthia and Chen, Zhiming and others},
  journal={arXiv preprint arXiv:2308.16149},
  year={2023}
}

@inproceedings{avramidis2024occiglot,
  title={Occiglot at WMT24: European open-source large language models evaluated on translation},
  author={Avramidis, Eleftherios and Gr{\"u}tzner-Zahn, Annika and Brack, Manuel and Schramowski, Patrick and Suarez, Pedro Ortiz and Ostendorff, Malte and Barth, Fabio and Manakhimova, Shushen and Macketanz, Vivien and Rehm, Georg and others},
  booktitle={Proceedings of the Ninth Conference on Machine Translation},
  pages={292--298},
  year={2024}
}

@article{zeng2022glm,
  title={Glm-130b: An open bilingual pre-trained model},
  author={Zeng, Aohan and Liu, Xiao and Du, Zhengxiao and Wang, Zihan and Lai, Hanyu and Ding, Ming and Yang, Zhuoyi and Xu, Yifan and Zheng, Wendi and Xia, Xiao and others},
  journal={arXiv preprint arXiv:2210.02414},
  year={2022}
}

@article{kreutzer2022quality,
  title={Quality at a glance: An audit of web-crawled multilingual datasets},
  author={Kreutzer, Julia and Caswell, Isaac and Wang, Lisa and Wahab, Ahsan and Van Esch, Daan and Ulzii-Orshikh, Nasanbayar and Tapo, Allahsera and Subramani, Nishant and Sokolov, Artem and Sikasote, Claytone and others},
  journal={Transactions of the Association for Computational Linguistics},
  volume={10},
  pages={50--72},
  year={2022},
  publisher={MIT Press One Rogers Street, Cambridge, MA 02142-1209, USA journals-info~…}
}

@article{hershcovich2022challenges,
  title={Challenges and strategies in cross-cultural NLP},
  author={Hershcovich, Daniel and Frank, Stella and Lent, Heather and De Lhoneux, Miryam and Abdou, Mostafa and Brandl, Stephanie and Bugliarello, Emanuele and Piqueras, Laura Cabello and Chalkidis, Ilias and Cui, Ruixiang and others},
  journal={arXiv preprint arXiv:2203.10020},
  year={2022}
}

@article{baltaji2024persona,
  title={Persona Inconstancy in Multi-Agent LLM Collaboration: Conformity, Confabulation, and Impersonation},
  author={Baltaji, Razan and Hemmatian, Babak and Varshney, Lav R},
  journal={arXiv preprint arXiv:2405.03862},
  year={2024}
}

@incollection{gruninger1995role,
  title={The role of competency questions in enterprise engineering},
  author={Gr{\"u}ninger, Michael and Fox, Mark S},
  booktitle={Benchmarking—Theory and practice},
  pages={22--31},
  year={1995},
  publisher={Springer}
}

@inproceedings{gruninger1995methodology,
  title={Methodology for the design and evaluation of ontologies},
  author={Gruninger, Michael},
  booktitle={Proc. IJCAI'95, Workshop on Basic Ontological Issues in Knowledge Sharing},
  year={1995}
}

@article{openai2025o4,
  title={o3 and o4-mini System Card},
  author={OpenAI, OpenAI},
  journal={Preprint},
  year={2025}
}

@techreport{google2025gemini25flashlite,
  author      = {{Google DeepMind}},
  title       = {Gemini 2.5 Flash-Lite Model Card},
  year        = {2025},
  institution = {Google DeepMind},
  howpublished = {\url{https://storage.googleapis.com/deepmind-media/Model-Cards/Gemini-2-5-Flash-Lite-Model-Card.pdf}},
  note        = {Accessed: 2025-12-28}
}

@misc{qwen2.5,
    title = {Qwen2.5: A Party of Foundation Models},
    url = {https://qwenlm.github.io/blog/qwen2.5/},
    author = {Qwen Team},
    month = {September},
    year = {2024}
}

@article{an2024exaone,
  title={EXAONE 3.5: Series of Large Language Models for Real-world Use Cases},
  author={An, Soyoung and Bae, Kyunghoon and Choi, Eunbi and Choi, Kibong and Choi, Stanley Jungkyu and Hong, Seokhee and Hwang, Junwon and Jeon, Hyojin and Jo, Gerrard Jeongwon and Jo, Hyunjik and others},
  journal={arXiv preprint arXiv:2412.04862},
  year={2024}
}

@article{he2020deberta,
  title={Deberta: Decoding-enhanced bert with disentangled attention},
  author={He, Pengcheng and Liu, Xiaodong and Gao, Jianfeng and Chen, Weizhu},
  journal={arXiv preprint arXiv:2006.03654},
  year={2020}
}

@article{wang2022text,
  title={Text embeddings by weakly-supervised contrastive pre-training},
  author={Wang, Liang and Yang, Nan and Huang, Xiaolong and Jiao, Binxing and Yang, Linjun and Jiang, Daxin and Majumder, Rangan and Wei, Furu},
  journal={arXiv preprint arXiv:2212.03533},
  year={2022}
}

@inproceedings{kiesel2023semeval,
  title={Semeval-2023 task 4: Valueeval: Identification of human values behind arguments},
  author={Kiesel, Johannes and Alshomary, Milad and Mirzakhmedova, Nailia and Heinrich, Maximilian and Handke, Nicolas and Wachsmuth, Henning and Stein, Benno},
  booktitle={Proceedings of the 17th International Workshop on Semantic Evaluation (SemEval-2023)},
  pages={2287--2303},
  year={2023}
}

@inproceedings{balikas2023john,
  title={John-Arthur at SemEval-2023 Task 4: fine-tuning large language models for arguments classification},
  author={Balikas, Georgios},
  booktitle={Proceedings of the 17th International Workshop on Semantic Evaluation (SemEval-2023)},
  pages={1428--1432},
  year={2023}
}

@article{alkhamissi2024investigating,
  title={Investigating cultural alignment of large language models},
  author={AlKhamissi, Badr and ElNokrashy, Muhammad and AlKhamissi, Mai and Diab, Mona},
  journal={arXiv preprint arXiv:2402.13231},
  year={2024}
}

@inproceedings{nadeem2021stereoset,
  title={StereoSet: Measuring stereotypical bias in pretrained language models},
  author={Nadeem, Moin and Bethke, Anna and Reddy, Siva},
  booktitle={Proceedings of the 59th annual meeting of the association for computational linguistics and the 11th international joint conference on natural language processing (volume 1: long papers)},
  pages={5356--5371},
  year={2021}
}

@article{tao2024cultural,
  title={Cultural bias and cultural alignment of large language models},
  author={Tao, Yan and Viberg, Olga and Baker, Ryan S and Kizilcec, Ren{\'e} F},
  journal={PNAS nexus},
  volume={3},
  number={9},
  pages={pgae346},
  year={2024},
  publisher={Oxford University Press US}
}

@article{chiu2024culturalbench,
  title={CulturalBench: a Robust, Diverse and Challenging Benchmark on Measuring (the Lack of) Cultural Knowledge of LLMs},
  author={Chiu, Yu Ying and Jiang, Liwei and Lin, Bill Yuchen and Park, Chan Young and Li, Shuyue Stella and Ravi, Sahithya and Bhatia, Mehar and Antoniak, Maria and Tsvetkov, Yulia and Shwartz, Vered and others},
  year={2024}
}

@inproceedings{rao2025normad,
  title={NormAd: A framework for measuring the cultural adaptability of large language models},
  author={Rao, Abhinav Sukumar and Yerukola, Akhila and Shah, Vishwa and Reinecke, Katharina and Sap, Maarten},
  booktitle={Proceedings of the 2025 Conference of the Nations of the Americas Chapter of the Association for Computational Linguistics: Human Language Technologies (Volume 1: Long Papers)},
  pages={2373--2403},
  year={2025}
}

@article{choenni2024self,
  title={Self-alignment: Improving alignment of cultural values in LLMs via in-context learning},
  author={Choenni, Rochelle and Shutova, Ekaterina},
  journal={arXiv preprint arXiv:2408.16482},
  year={2024}
}

@article{seo2025valuesrag,
  title={Valuesrag: Enhancing cultural alignment through retrieval-augmented contextual learning},
  author={Seo, Wonduk and Yuan, Zonghao and Bu, Yi},
  journal={arXiv preprint arXiv:2501.01031},
  year={2025}
}

@article{ki2025multiple,
  title={Multiple LLM Agents Debate for Equitable Cultural Alignment},
  author={Ki, Dayeon and Rudinger, Rachel and Zhou, Tianyi and Carpuat, Marine},
  journal={arXiv preprint arXiv:2505.24671},
  year={2025}
}

@article{wan2025cultural,
  title={Which Cultural Lens Do Models Adopt? On Cultural Positioning Bias and Agentic Mitigation in LLMs},
  author={Wan, Yixin and Chen, Xingrun and Chang, Kai-Wei},
  journal={arXiv preprint arXiv:2509.21080},
  year={2025}
}

@article{karinshak2024llm,
  title={Llm-globe: A benchmark evaluating the cultural values embedded in llm output},
  author={Karinshak, Elise and Hu, Amanda and Kong, Kewen and Rao, Vishwanatha and Wang, Jingren and Wang, Jindong and Zeng, Yi},
  journal={arXiv preprint arXiv:2411.06032},
  year={2024}
}

@inproceedings{dhamala2021bold,
  title={Bold: Dataset and metrics for measuring biases in open-ended language generation},
  author={Dhamala, Jwala and Sun, Tony and Kumar, Varun and Krishna, Satyapriya and Pruksachatkun, Yada and Chang, Kai-Wei and Gupta, Rahul},
  booktitle={Proceedings of the 2021 ACM conference on fairness, accountability, and transparency},
  pages={862--872},
  year={2021}
}

@inproceedings{parrish2022bbq,
  title={BBQ: A hand-built bias benchmark for question answering},
  author={Parrish, Alicia and Chen, Angelica and Nangia, Nikita and Padmakumar, Vishakh and Phang, Jason and Thompson, Jana and Htut, Phu Mon and Bowman, Samuel},
  booktitle={Findings of the Association for Computational Linguistics: ACL 2022},
  pages={2086--2105},
  year={2022}
}

@article{zhao2024worldvaluesbench,
  title={Worldvaluesbench: A large-scale benchmark dataset for multi-cultural value awareness of language models},
  author={Zhao, Wenlong and Mondal, Debanjan and Tandon, Niket and Dillion, Danica and Gray, Kurt and Gu, Yuling},
  journal={arXiv preprint arXiv:2404.16308},
  year={2024}
}

@article{haerpfer2022world,
  title={World values survey wave 7 (2017-2022) cross-national data-set},
  author={Haerpfer, Christian and Inglehart, Ronald and Moreno, Alejandro and Welzel, Christian and Kizilova, Kseniya and Diez-Medrano, Jaime and Lagos, Marta and Norris, Pippa and Ponarin, Eduard and Puranen, Bi},
  journal={(No Title)},
  year={2022},
  publisher={World Values Survey Association}
}

@article{durmus2023globalopinionqa,
  title={Towards measuring the representation of subjective global opinions in language models},
  author={Durmus, Esin and Nguyen, Karina and Liao, Thomas I and Schiefer, Nicholas and Askell, Amanda and Bakhtin, Anton and Chen, Carol and Hatfield-Dodds, Zac and Hernandez, Danny and Joseph, Nicholas and others},
  journal={arXiv preprint arXiv:2306.16388},
  year={2023}
}

@article{li2024culturellm,
  title={Culturellm: Incorporating cultural differences into large language models},
  author={Li, Cheng and Chen, Mengzhuo and Wang, Jindong and Sitaram, Sunayana and Xie, Xing},
  journal={Advances in Neural Information Processing Systems},
  volume={37},
  pages={84799--84838},
  year={2024}
}

@article{Gruber93,
  title={A translation approach to portable ontology specifications},
  author={Gruber, Thomas R},
  journal={Knowledge acquisition},
  volume={5},
  number={2},
  pages={199--220},
  year={1993},
  publisher={Elsevier}
}

@inproceedings{Fernandez1997,
  author    = {Fern{\'a}ndez-L{\'o}pez, Mariano and G{\'o}mez-P{\'e}rez, Asunci{\'o}n and Juristo, Natalia},
  title     = {METHONTOLOGY: From Ontological Art Towards Ontological Engineering},
  booktitle = {Proceedings of the Symposium on Ontological Engineering of AAAI},
  year      = {1997},
  publisher = {AAAI Press},
  address   = {Stanford, CA, USA},
  pages     = {33--40}
}

@incollection{SureSS04,
  title={On-to-knowledge methodology (OTKM)},
  author={Sure, York and Staab, Steffen and Studer, Rudi},
  booktitle={Handbook on ontologies},
  pages={117--132},
  year={2004},
  publisher={Springer}
}

@incollection{Suarez2012,
  title={Introduction: Ontology engineering in a networked world},
  author={Su{\'a}rez-Figueroa, Mari Carmen and G{\'o}mez-P{\'e}rez, Asunci{\'o}n and Motta, Enrico and Gangemi, Aldo},
  booktitle={Ontology engineering in a networked world},
  pages={1--6},
  year={2011},
  publisher={Springer}
}

@inproceedings{GiglouDA23,
  title={LLMs4OL: Large language models for ontology learning},
  author={Babaei Giglou, Hamed and D’Souza, Jennifer and Auer, S{\"o}ren},
  booktitle={International Semantic Web Conference},
  pages={408--427},
  year={2023},
  organization={Springer}
}

@inproceedings{MateiuG23,
  title={Ontology engineering with large language models},
  author={Mateiu, Patricia and Groza, Adrian},
  booktitle={2023 25th International Symposium on Symbolic and Numeric Algorithms for Scientific Computing (SYNASC)},
  pages={226--229},
  year={2023},
  organization={IEEE}
}

@inproceedings{SaeedizadeB24,
  author       = {Mohammad Javad Saeedizade and
                  Eva Blomqvist},
  editor       = {Albert Mero{\~{n}}o{-}Pe{\~{n}}uela and
                  Anastasia Dimou and
                  Rapha{\"{e}}l Troncy and
                  Olaf Hartig and
                  Maribel Acosta and
                  Mehwish Alam and
                  Heiko Paulheim and
                  Pasquale Lisena},
  title        = {Navigating Ontology Development with Large Language Models},
  booktitle    = {The Semantic Web - 21st International Conference, {ESWC} 2024, Hersonissos,
                  Crete, Greece, May 26-30, 2024, Proceedings, Part {I}},
  series       = {Lecture Notes in Computer Science},
  volume       = {14664},
  pages        = {143--161},
  publisher    = {Springer},
  year         = {2024},
  url          = {https://doi.org/10.1007/978-3-031-60626-7\_8},
  doi          = {10.1007/978-3-031-60626-7\_8},
  bibsource    = {dblp computer science bibliography, https://dblp.org}
}

@inproceedings{LippolisSKZCGBN25,
  title={Ontology generation using large language models},
  author={Lippolis, Anna Sofia and Saeedizade, Mohammad Javad and Keskis{\"a}rkk{\"a}, Robin and Zuppiroli, Sara and Ceriani, Miguel and Gangemi, Aldo and Blomqvist, Eva and Nuzzolese, Andrea Giovanni},
  booktitle={European Semantic Web Conference},
  pages={321--341},
  year={2025},
  organization={Springer}
}

@article{wang2022self,
  title={Self-consistency improves chain of thought reasoning in language models},
  author={Wang, Xuezhi and Wei, Jason and Schuurmans, Dale and Le, Quoc and Chi, Ed and Narang, Sharan and Chowdhery, Aakanksha and Zhou, Denny},
  journal={arXiv preprint arXiv:2203.11171},
  year={2022}
}

@article{cohen1960coefficient,
  title={A coefficient of agreement for nominal scales},
  author={Cohen, Jacob},
  journal={Educational and psychological measurement},
  volume={20},
  number={1},
  pages={37--46},
  year={1960},
  publisher={Sage Publications Sage CA: Thousand Oaks, CA}
}

@inproceedings{tseliou2024survey,
  title={THE SURVEY OF ISSP 2023: THE IMPACT OF NATIONAL IDENTITY AND CITIZENSHIP ON POLITICAL ATTITUDES},
  author={Tseliou, S and Andreadis, I},
  booktitle={Proceedings of the 36th Panhellenic \& 2nd International Stat Conf},
  pages={000--000},
  year={2024}
}
\bibliographystyle{icml2026}

\newpage

\appendix
\onecolumn


\section{Training Details and Loss Curves}
\label{app:deberta_training}

\subsection{DeBERTa-v2-xxlarge Fine-tuning}

We fine-tuned DeBERTa-v2-xxlarge~\citep{he2020deberta} on the WVS category classification task (191 questions, 12 categories) using batch size 4, learning rate $5 \times 10^{-6}$, AdamW optimizer (weight decay 0.01), and FP16 precision on an NVIDIA A100 GPU. Training terminated early at epoch 3 when validation Top-3 accuracy reached 100\%, as shown in Figure~\ref{fig:loss_curve}.

{
\setlength{\intextsep}{5pt}
\setlength{\belowcaptionskip}{0pt}
\begin{figure}[H]
\centering
\includegraphics[width=8cm, height=6cm]{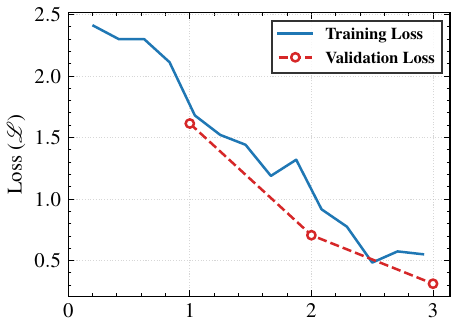}

\caption{\textbf{Training and validation loss curves for DeBERTa-v2-xxlarge fine-tuning.} The x-axis represents epochs. Training loss (blue solid line) exhibits minor fluctuations typical of small-batch optimization, while validation loss (red dashed line, evaluated every 48 steps) decreases monotonically from 1.61 to 0.31 across three epochs, indicating effective learning without overfitting.}
\label{fig:loss_curve}
\end{figure}
}

The early stopping criterion based on Top-3 accuracy ensures computational efficiency. Despite minor fluctuations in training loss during the final epoch, validation loss decreases consistently across all epochs on the WVS dataset.

\subsection{Value Category Classification}
{%
\setlength{\intextsep}{5pt}
\setlength{\belowcaptionskip}{-5pt}

\begin{table}[H]
\centering
\caption{\textbf{Topic classification performance on six regional test datasets and WVS validation data.} Top-$k$: fraction of questions where the true category appears in top-$k$ predictions. All metrics in [0,1].}
\label{tab:regional_eval}
\begin{tabular}{lccc}  
\toprule
\textbf{Dataset} & \textbf{Top-1} & \textbf{Top-2} & \textbf{Top-3} \\  
\midrule
Afrobarometer & 0.5037 & 0.6875 & 0.7574 \\
CGSS          & 0.3375 & 0.5079 & 0.6656 \\
EVS           & 0.4315 & 0.5560 & 0.6680 \\
GSS           & 0.4545 & 0.6667 & 0.7765 \\
ISD           & 0.5439 & 0.7071 & 0.7950 \\
LAPOP         & 0.4396 & 0.6577 & 0.7349 \\
\midrule
\textbf{WVS (val)} & \textbf{0.9583} & \textbf{1.0000} & \textbf{1.0000} \\
\bottomrule
\end{tabular}
\end{table}

The lower Top-1 accuracy on the six regional test datasets reflects taxonomy coverage limitations and question-style mismatch rather than classifier failure. Regional surveys were designed independently of WVS, and their phrasings do not always align with the 12-category taxonomy on which the model was trained. Ground-truth labels were assigned through independent blinded consensus annotation to ensure reliability. To account for this uncertainty, \textbf{OG-MAR} retrieves ontology triples from all three predicted categories rather than relying on the top prediction alone, broadening coverage when the correct category may not rank first. Top-3 accuracy therefore directly corresponds to the operational condition under which the framework runs. We examine how topic-classifier accuracy affects downstream performance in Appendix~\ref{app:error_propagation}.

\subsection{Topic-Classifier Error Propagation}
\label{app:error_propagation}

In \textbf{OG-MAR}, DeBERTa's predicted topic categories determine which ontology triples are retrieved for each question. To measure whether classifier errors propagate to downstream performance, we partitioned the regional test questions into two groups based on whether the ground-truth category appeared in DeBERTa's Top-3 predictions.

\begin{table}[H]
\centering
\caption{\textbf{OG-MAR average scores by topic-classifier coverage.} ``In Top-3'' indicates that the ground-truth category appears in DeBERTa's Top-3 predictions. ``Not in Top-3'' indicates it does not. Evaluation metric is Avg. Score. Bold denotes the higher value per model.}
\label{tab:error_propagation}
\begin{tabular}{lcc}
\toprule
\textbf{Model} & \textbf{In Top-3} & \textbf{Not in Top-3} \\
\midrule
GPT-4o mini           & \textbf{0.6157} & 0.6114 \\
Gemini 2.5 Flash Lite & \textbf{0.6435} & 0.6232 \\
QWEN 2.5              & \textbf{0.7039} & 0.6448 \\
EXAONE 3.5            & \textbf{0.6317} & 0.6149 \\
\bottomrule
\end{tabular}
\end{table}

Table~\ref{tab:error_propagation} reports average scores for each group across four backbone models. The \textbf{In Top-3} group consistently achieves higher scores than the \textbf{Not in Top-3} group, though the gap varies across backbones. GPT-4o mini shows the smallest difference at 0.004, while QWEN 2.5 shows the largest at 0.059. Gemini 2.5 Flash Lite and EXAONE 3.5 fall in between at 0.020 and 0.017, respectively. Despite this backbone-dependent variation, the pipeline continues to produce meaningful outputs even when the correct category is not covered. We attribute this robustness to the semantic proximity of neighboring categories and the partial relevance of triples retrieved from adjacent topics.

\section{Additional Evaluation Metrics}
\label{app:evaluation_details}
To complement the main evaluation, we conduct additional distribution-level analyses using three metrics: Earth Mover's Distance (EMD; lower is better), Proportion Correlation (higher is better), and Total Variation Distance (TVD; lower is better). These metrics evaluate whether the simulated survey responses match the original ordinal response distributions, rather than relying only on a coarse binary outcome. As summarized in Table~\ref{tab:additional_metrics}, the results are averaged over six survey datasets: EVS, GSS, CGSS, ISD, AFRO, and LAPOP.\\

\begin{table}[h]
\centering
\caption{\textbf{Additional distribution-level evaluation across four LLM backbones.} All scores are averaged over six survey datasets. \textbf{Bold} indicates the best score within each backbone and metric. \textbf{OG-MAR} achieves the best EMD, Proportion Correlation, and TVD on GPT-4o mini, Gemini 2.5 Flash Lite, and QWEN 2.5, showing consistent distributional alignment beyond the main evaluation metric.}
\scriptsize
\begin{tabular}{lccc}
\toprule
\textbf{Method} & \textbf{EMD} $\downarrow$ & \textbf{Prop. Corr.} $\uparrow$ & \textbf{TVD} $\downarrow$ \\
\midrule
\rowcolor{gray!18}
\multicolumn{4}{l}{\textbf{GPT-4o mini}} \\
Zero-shot       & 0.6744 & 0.4787 & 0.4175 \\
Role            & 0.6559 & 0.4082 & 0.4230 \\
Self-consistent & 0.7760 & 0.4136 & 0.4511 \\
Debate          & 0.6609 & 0.4494 & 0.4179 \\
ValuesRAG       & 0.6800 & 0.4651 & 0.4370 \\
\textbf{OG-MAR} & \textbf{0.6203} & \textbf{0.4941} & \textbf{0.3950} \\
\midrule
\rowcolor{gray!18}
\multicolumn{4}{l}{\textbf{Gemini 2.5 Flash Lite}} \\
Zero-shot       & 0.7169 & 0.4340 & 0.4410 \\
Role            & 0.7416 & 0.4137 & 0.4471 \\
Self-consistent & 0.7997 & 0.4547 & 0.4831 \\
Debate          & 0.6087 & 0.4853 & 0.4173 \\
ValuesRAG       & 0.6789 & 0.4834 & 0.4278 \\
\textbf{OG-MAR} & \textbf{0.5888} & \textbf{0.5205} & \textbf{0.4023} \\
\midrule
\rowcolor{gray!18}
\multicolumn{4}{l}{\textbf{QWEN 2.5}} \\
Zero-shot       & 0.8500 & 0.3474 & 0.5277 \\
Role            & 0.7670 & 0.3698 & 0.4972 \\
Self-consistent & 0.8003 & 0.3880 & 0.5045 \\
Debate          & 0.7355 & 0.3508 & 0.4630 \\
ValuesRAG       & 0.7531 & 0.4029 & 0.4960 \\
\textbf{OG-MAR} & \textbf{0.6992} & \textbf{0.4264} & \textbf{0.4606} \\
\midrule
\rowcolor{gray!18}
\multicolumn{4}{l}{\textbf{EXAONE 3.5}} \\
Zero-shot       & 0.9159 & 0.2658 & 0.5572 \\
Role            & 0.8678 & 0.2575 & 0.5324 \\
Self-consistent & 0.9784 & 0.2776 & 0.5556 \\
Debate          & \textbf{0.7940} & 0.2943 & 0.5342 \\
ValuesRAG       & 0.8540 & 0.2969 & 0.5030 \\
\textbf{OG-MAR} & 0.8014 & \textbf{0.3052} & \textbf{0.4894} \\
\bottomrule
\end{tabular}
\label{tab:additional_metrics}
\end{table}

Table~\ref{tab:additional_metrics} shows that \textbf{OG-MAR} achieves the best EMD, Proportion Correlation, and TVD on GPT-4o mini, Gemini 2.5 Flash Lite, and QWEN 2.5. This indicates that \textbf{OG-MAR} improves distributional alignment across multiple backbones, not merely under a single evaluation metric. The only partial exception appears on EXAONE 3.5, where \textbf{OG-MAR} still obtains the best Proportion Correlation and TVD, but does not achieve the best EMD. This suggests that the effectiveness of \textbf{OG-MAR} remains generally robust, while weaker instruction-following backbones may limit some aspects of distributional recovery.

We further examine whether \textbf{OG-MAR's} pre-aggregation persona vote proportions provide a more fine-grained approximation of the target response distribution. Table~\ref{tab:vote_prop_metrics} compares the standard binary prediction setting with the vote-proportion setting. The binary setting uses the final hard prediction from the Judgment Agent and is directly comparable to all baselines. In contrast, the vote-proportion setting uses the pre-aggregation votes from $K=5$ persona agents as a continuous approximation of $P_{\text{model}}(y=1 \mid q,g)$; therefore, it is reported separately and should not be interpreted as a direct baseline comparison.

\begin{table}[t]
\centering
\caption{\textbf{Distributional evaluation of \OG-MAR under binary and vote-proportion variants.} The vote-proportion variant uses pre-aggregation persona votes and is not directly comparable to binary baselines. Nevertheless, it generally improves distributional recovery, especially in TVD.}
\footnotesize
\setlength{\tabcolsep}{6pt}
\renewcommand{\arraystretch}{1.08}
\begin{tabular}{lcccc}
\toprule
\textbf{OG-MAR Variant} & \textbf{GPT} & \textbf{Gemini} & \textbf{QWEN} & \textbf{EXAONE} \\
\midrule

\rowcolor{gray!18}
\multicolumn{5}{l}{\textbf{TVD $\downarrow$}} \\
Binary      & 0.3950 & 0.4023 & 0.4606 & 0.4894 \\
Vote prop.  & \textbf{0.2057} & \textbf{0.2271} & \textbf{0.2473} & \textbf{0.2985} \\

\midrule
\rowcolor{gray!18}
\multicolumn{5}{l}{\textbf{Prop. Corr. $\uparrow$}} \\
Binary      & 0.4941 & \textbf{0.5205} & 0.4264 & 0.3052 \\
Vote prop.  & \textbf{0.5645} & 0.4925 & \textbf{0.4871} & \textbf{0.4569} \\

\bottomrule
\end{tabular}
\label{tab:vote_prop_metrics}
\end{table}

As shown in Table~\ref{tab:vote_prop_metrics}, the vote-proportion variant substantially improves TVD across all four backbones, suggesting that the intermediate persona votes contain useful distributional information before final binary aggregation. This result further supports the design motivation of \textbf{OG-MAR}: multi-agent persona reasoning can capture diverse response tendencies that are partially lost when converted into a single hard prediction.
\section{Dataset Details}
\label{app:dataset_details}
\begin{figure}[H]
\centering
\includegraphics[width=\textwidth]{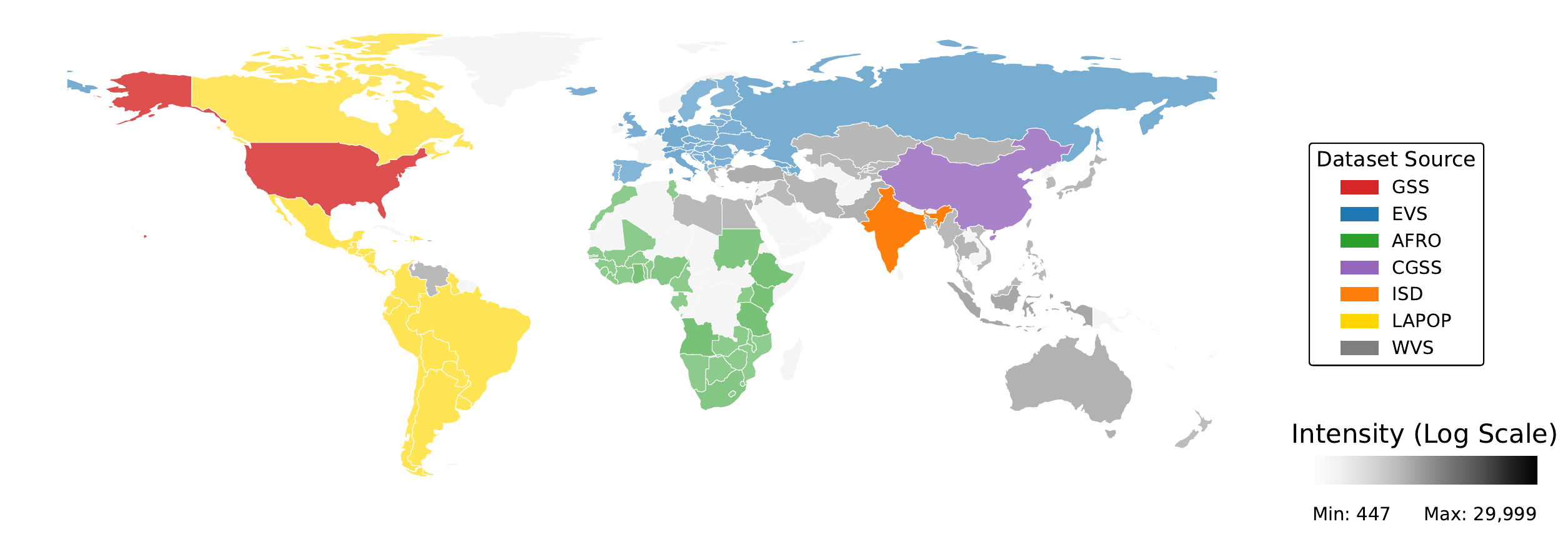}
\caption{\textbf{Geographic coverage of cultural value datasets used in this study.} Each country is colored according to its primary data source, prioritizing regional surveys over the global World Values Survey. Regional datasets include the General Social Survey for the United States, the European Values Study for Europe, Afrobarometer for Africa, the Chinese General Social Survey for China, and the India Social Dataset for India. Countries without regional coverage are represented by WVS data shown in gray. Color intensity reflects participant count on a logarithmic scale, ranging from 447 to 29,999 respondents per country. This multimodal approach ensures both regional specificity and global breadth in cultural alignment research.}
\label{fig:dataset_coverage}
\end{figure}

\begin{table}[t]
\caption{\textbf{Summary statistics of the retrieval corpus (WVS) and six test datasets.} For test datasets, ``\#Value Qs'' denotes the value-related items retained after our preprocessing/topic mapping (not necessarily the full questionnaire length).}
\centering
\small
\setlength{\tabcolsep}{5.5pt}
\renewcommand{\arraystretch}{1.12}
\begin{tabular}{l l l l r r r}
\toprule
\textbf{Dataset} & \textbf{Type} & \textbf{Region} & \textbf{Wave / Year} & \textbf{\#Countries} & \textbf{\#Respondents} & \textbf{\#Value Qs} \\
\hline
\multicolumn{7}{>{\columncolor{gray!15}}l}{\textit{Retrieval Corpus}} \\
WVS (World Values Survey) & Retrieval & Global & 2017--2022 & 64 & 94{,}728 & 239 \\
\hline
\multicolumn{7}{>{\columncolor{gray!15}}l}{\textit{Test Datasets}} \\
EVS (European Values Study) & Test & Europe & 2017 & -- & 59{,}438 & 211 \\
GSS (General Social Survey) & Test & United States & 2021--2022 & -- & 8{,}181 & 44 \\
CGSS (Chinese General Social Survey) & Test & China (E. Asia) & 2021 & -- & $\sim$8{,}148 & 58 \\
ISD (Pew India Survey Dataset) & Test & India (S. Asia) & 2019--2020 & -- & 29{,}999 & 33 \\
LAPOP (AmericasBarometer) & Test & Latin America & 2021 & -- & 64,352 & 48 \\
Afrobarometer & Test & Africa & 2022 & -- & $\sim$48{,}100 & 144 \\
\bottomrule
\end{tabular}

\label{tab:datasets_stats}
\end{table}

\begin{table*}[t]
\centering
\small
\caption{\textbf{Data sources used in our experiments.} We use the World Values Survey (WVS) as the retrieval corpus, and evaluate generalization on six external test datasets (EVS, GSS, CGSS, ISD, LAPOP, and Afrobarometer), with official access links provided for reproducibility.}

\label{tab:datasets}
\begin{tabular}{ll}
\toprule
\textbf{Dataset} & \textbf{Link} \\
\hline
\rowcolor{gray!10}\multicolumn{2}{l}{\textit{Retrieval Corpus}} \\
WVS &
\url{https://www.worldvaluessurvey.org/wvs.jsp} \\

\hline
\rowcolor{gray!10}\multicolumn{2}{l}{\textit{Test Datasets}} \\
EVS (European Values Study)&
\url{https://europeanvaluesstudy.eu} \\
GSS (General Social Survey)&
\url{https://gss.norc.org} \\
CGSS (Chinese General Social Survey)&
\url{https://cgss.ruc.edu.cn} \\
ISD (Pew India Survey Dataset)&
\url{https://www.pewresearch.org/dataset/india-survey-dataset/} \\
LAPOP (AmericasBarometer)&
\url{https://www.vanderbilt.edu/lapop} \\
Afrobarometer &
\url{https://www.afrobarometer.org} \\

\bottomrule
\end{tabular}
\end{table*}

\begin{figure}[H]
    \centering
    \includegraphics[width=0.4\textwidth]{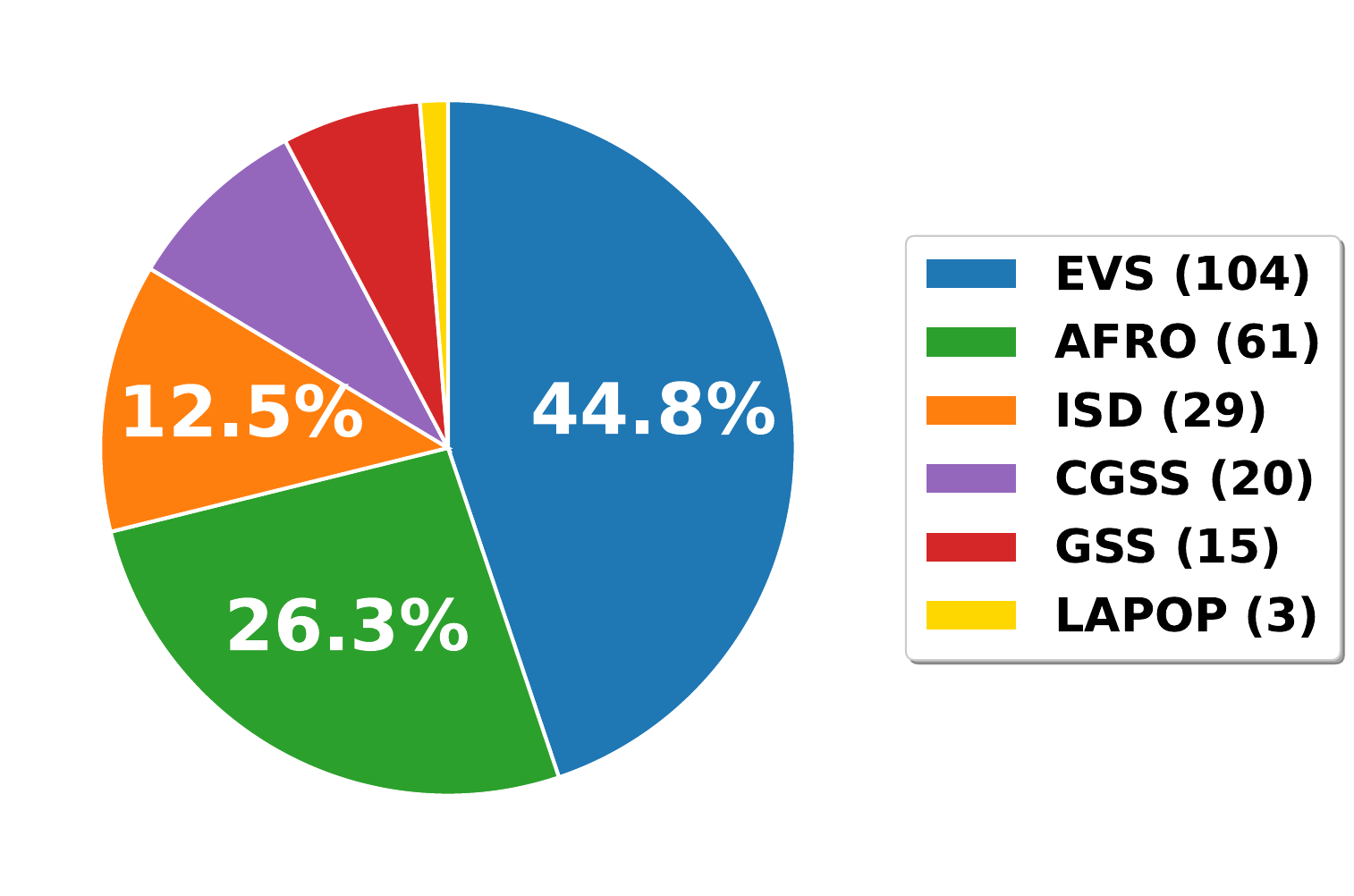}
    \caption{\textbf{Distribution of selected value questions across regional datasets.}}
    \label{fig:regional_distribution}
\end{figure}

To ensure the quality and relevance of our value questions, we employed a systematic selection process for the regional datasets.
First, we considered the actual response rates from our extracted representative samples (see Section~\ref{Extract Representative Sample to Cluster}) and retained the top 80\% of questions with valid responses in each dataset. This ensured sufficient data coverage for meaningful analysis. Second, we excluded questions that were overly dependent on personal circumstances and behaviors, 
such as ``Did your household ever run out of water?'', 
``In the past year, how often have you used radio?'', 
``Are you a member of a trade union?'', 
or ``Based on your experience, how easy or difficult is it to find out how government uses tax revenues?''. These questions tend to reflect personal circumstances rather than underlying values. Third, we excluded questions requiring knowledge of specific countries or domestic institutions, such as ``To what extent do you think [country] is democratic?'' or ``How much do you trust the Electoral Commission of Ghana?''. It would be difficult to generalize these questions across cultural contexts.
After applying these criteria, we obtained a final set of questions distributed across the six regional datasets. Figure~\ref{fig:regional_distribution} shows the distribution of these selected questions across datasets.

\begin{table}[t]
\caption{\textbf{Distribution of Values-related Questions in WVS.} The questions were categorized into 12 topics with a total of 250 questions in the original WVS questionnaire covering most of the dimensions of values.}
\centering
\small
\setlength{\tabcolsep}{6pt}
\begin{tabular}{l r}
\toprule
\textbf{Topic} & \textbf{Count} \\
\midrule
Social Values, Norms, Stereotypes & 45 \\
Happiness and Wellbeing & 11 \\
Social Capital, Trust and Organizational Membership & 47 \\
Economic Values & 6 \\
Perceptions of Corruption & 9 \\
Perceptions of Migration & 10 \\
Perceptions of Security & 21 \\
Perceptions about Science and Technology & 6 \\
Religious Values & 12 \\
Ethical Values & 23 \\
Political Interest and Political Participation & 35 \\
Political Culture and Political Regimes & 25 \\
\bottomrule
\end{tabular}
\label{tab:wvs_values_topics}
\end{table}

\subsection{Dataset Information}
\paragraph{Retrieval Corpus: World Values Survey (WVS).}
We use the \textit{World Values Survey} (WVS) as our retrieval corpus. In this setting, WVS covers 64 countries/territories with 94{,}728 respondents and a 290-item common questionnaire. We organize the value space into 12 predefined topics (Table~\ref{tab:wvs_values_topics}) and, after preprocessing, retain 239 region-agnostic ordinal value items for retrieval-augmented inference. 

\paragraph{Test Datasets.}
To evaluate performance of \textbf{OG-MAR}, we use six regional social-survey datasets spanning East Asia, South Asia, Europe, North America, Latin America, and Africa. For each test dataset, the reported number of value questions refers to the subset retained after our preprocessing/topic mapping (not necessarily the full questionnaire length).

\paragraph{EVS (European Values Study).}
EVS is a cross-national survey program designed to measure human values and socio-cultural attitudes across Europe using harmonized instruments. We use the EVS~2017 integrated dataset, which combines nationally representative samples from 36 countries (59{,}438 respondents) collected during the 2017--2021 fieldwork period. In our evaluation, we retain 211 value-related items under our preprocessing and topic mapping, enabling within-Europe generalization tests under relatively consistent survey design and documentation.

\paragraph{GSS (General Social Survey).}
GSS is a long-running repeated cross-sectional survey for the United States, featuring a replicating core plus rotating topical modules and rich demographics. We use the 2021 and 2022 cross-sections (4{,}032 and 4{,}149 completes; 8{,}181 total), which reflect a major methodological transition (e.g., 2021 push-to-web design and 2022 mixed-mode transition). We retain 44 value-related items, making GSS a useful stress test for robustness under both cultural shift (vs.\ the global WVS corpus) and survey-mode/design differences across waves.

\paragraph{CGSS (Chinese General Social Survey).}
CGSS is a nationally representative household survey measuring social attitudes and values in China, accompanied by detailed demographic covariates. We use the 2021 release with 8{,}148 valid samples (often documented as being drawn nationwide across many communities/provinces). After preprocessing and topic alignment, we retain 58 value-related items, providing a linguistically and institutionally distinct setting for cross-cultural generalization beyond the WVS retrieval corpus.

\paragraph{ISD (Pew India Survey Dataset).}
To represent South Asia, we use Pew Research Center's India Survey Dataset, a nationally representative face-to-face survey administered from Nov.~2019 to Mar.~2020. The dataset contains 29{,}999 adult interviews and covers broad attitudinal themes (e.g., identity, nationalism, tolerance), alongside demographics and survey weights. We retain 33 value-related items under our mapping, supporting evaluation in a highly heterogeneous population with strong methodological transparency.

\paragraph{LAPOP (AmericasBarometer).}
LAPOP is a large-scale cross-national public opinion survey program in Latin America that measures citizens' political attitudes, democratic governance, institutional trust, and socio-economic perceptions with harmonized instruments and rich demographics. We use the 2021 AmericasBarometer release (64{,}352 respondents) and, after preprocessing and topic alignment, retain 48 value-related items, providing a regionally distinct testbed for evaluating generalization beyond the WVS retrieval corpus.

\paragraph{Afrobarometer.}
Afrobarometer conducts nationally representative surveys of adult citizens (18+) across African countries using probability sampling, typically with per-country sample sizes of about 1,200 or 2,400. We use the merged Round~8 release (34 countries; released as a 2022 merged dataset) and retain 144 value-related items. The merged file contains on the order of $\sim$48K respondents (depending on country-level sample sizes), providing a stringent test bed with substantial cross-country diversity in socio-economic and governance contexts.

\subsection{Extract Representative Sample to Cluster} 
\label{Extract Representative Sample to Cluster}

\begin{figure}[t]
\centering
\includegraphics[width=\textwidth]{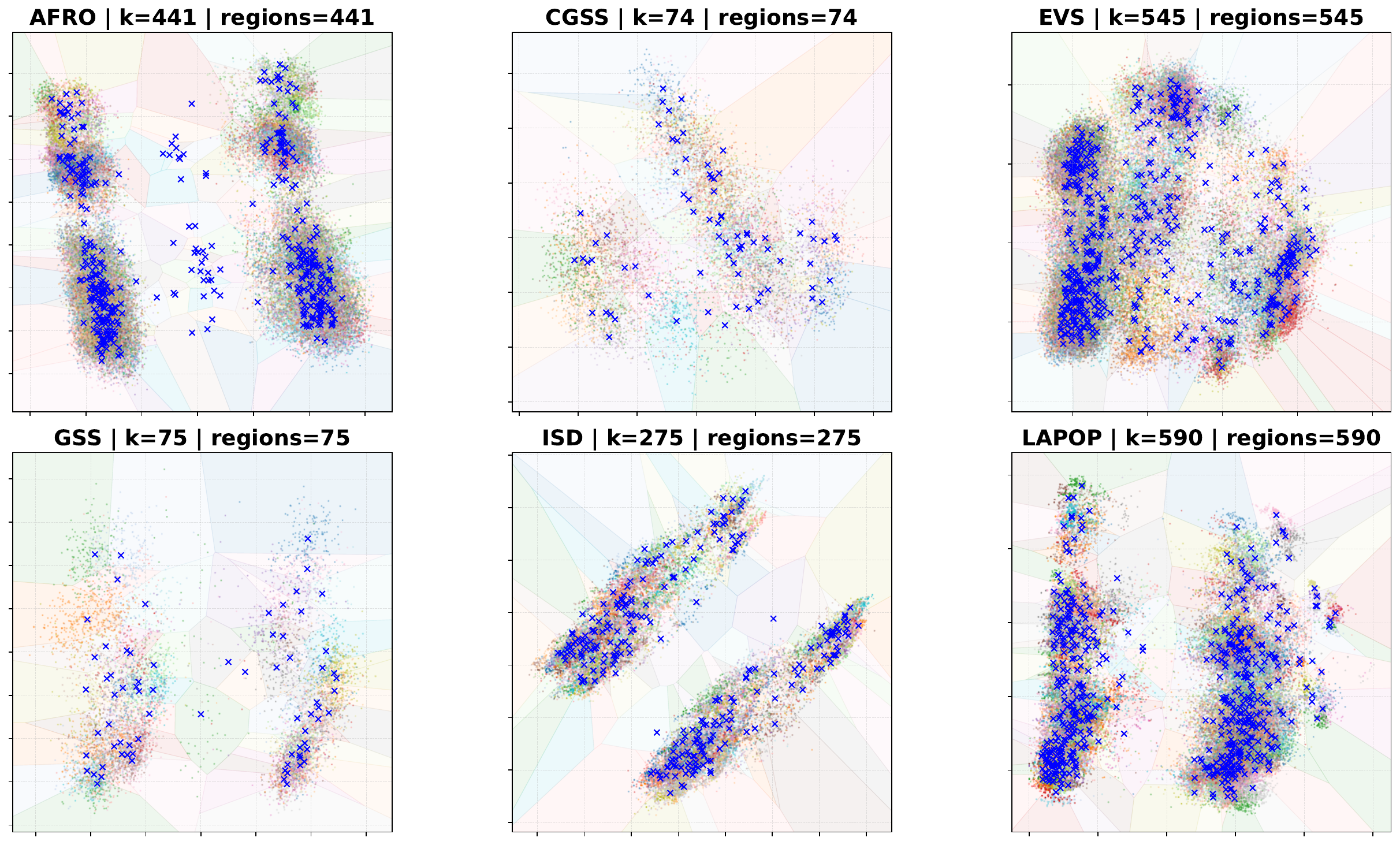}
\caption{\textbf{Voronoi visualization of Faiss k-means centroids for six embedding datasets.} Blue crosses denote cluster centroids, colored dots indicate embedded samples, and light polygons show Voronoi regions in a 2D projection, providing an intuitive overview of the spatial distribution and structure of the embedding space across datasets.}

\label{fig:voronoi_result}
\end{figure}

\paragraph{Representative Sample} To ensure computational efficiency while maintaining representativeness across regional datasets, we extract 2{,}000 representative instances with balanced coverage through clustering-based sampling. Concretely, we use \textbf{Faiss-based k-means clustering} to learn \(k\) clusters in the embedding space under Euclidean (L2) distance and visualize the resulting structure with a \textbf{Voronoi partition} induced by the cluster centroids, optimizing the standard k-means objective:
\begin{equation}
\min_{\{c_j\}_{j=1}^{k}} \sum_{i=1}^{N} \left\lVert x_i - c_{z_i} \right\rVert_2^2,
\end{equation}
where \(x_i \in \mathbb{R}^d\) denotes the embedding of the \(i\)-th sample, \(c_j \in \mathbb{R}^d\) is the centroid of cluster \(j\), and \(z_i \in \{1,\ldots,k\}\) is the cluster assignment of \(x_i\). After training, we extract the \textbf{centroids} \(\{c_j\}_{j=1}^{k}\) as representative cluster prototypes. We then project the centroids to two dimensions and construct a \textbf{Voronoi diagram} based on nearest-centroid relations in the 2D plane, which provides an intuitive view of how cluster centers are arranged and how the space is partitioned around them.

Figure~\ref{fig:voronoi_result} shows results for six datasets (\textsc{AFRO}, \textsc{CGSS}, \textsc{EVS}, \textsc{GSS}, \textsc{ISD}, and \textsc{LAPO}). For each dataset, we run Faiss k-means with the specified \(k\) and plot the \textbf{centroids (\textcolor{blue}{\(\times\)})} together with the corresponding \textbf{Voronoi cells} (light-colored polygons). The small dots are individual embedding samples, colored by their k-means cluster assignments \(z_i\). In each subplot, $k$ denotes the number of clusters (and thus centroids), while $regions$ reports the number of Voronoi cells formed in the 2D projection. In our outputs, $regions$ matches $k$, indicating that each centroid produces one Voronoi region. Overall, we extract \(2{,}000\) centroid prototypes across the six datasets, and use these prototypes as a compact set of representative points for subsequent analysis. To preserve representativeness across the regional datasets, we apply a clustering-based sampling strategy with a fixed sampling rate for each dataset. This procedure yields 441 instances from AFRO, 74 from CGSS, 545 from EVS, 75 from GSS, 275 from ISD, and 590 from LAPOP (2{,}000 instances in total).

This centroid-based summary is justified because k-means centroids serve as prototypes that compactly approximate the embedding distribution. By minimizing within-cluster squared distances, the centroids capture the central tendency of dense regions and yield a stable, noise-robust representation of groups of similar points. In contrast to random sampling—which can overrepresent frequent patterns and overlook sparse yet important regions—k-means places prototypes more systematically, improving coverage while keeping the summary compact. Consequently, the extracted centroids provide a practical and representative set of summary points for downstream analysis and sampling.

\subsection{Post-Cutoff Control Dataset}
\label{app:issp_control}

Our main test datasets (EVS, GSS, CGSS, ISD, LAPOP, and Afrobarometer) were all released prior to the training cutoffs of the evaluated models, raising a concern that observed performance gains may partly reflect memorization of survey content rather than genuine cultural reasoning. To address this, we introduce a post-cutoff control experiment using ISSP (International Social Survey Programme)~\citep{tseliou2024survey}, a cross-national collaborative survey programme running since 1985 and now covering over 58 member countries. The 2023 wave was published by GESIS on March~13, 2026.\footnote{Version 1.0.0, 
collected from September~23, 2022 to August~18, 2025. 
\url{https://search.gesis.org/research_data/ZA10010}} 
The dataset contains 43{,}517 respondents and 355 variables across 31 national samples. 
Following the same question selection procedure as our main experiment process, 
we retain 68 value-related questions. This release date postdates the publicly 
announced training cutoffs of all four backbone models evaluated in this work.

\paragraph{Experimental Setup and Results.}
From value question selection to final evaluation, the entire experimental process follows our main pipeline, using the same category labeling and baselines, and reports accuracy as the evaluation metric. We sampled 100 representative respondents per continent across five continents (Africa, Americas, Asia, Europe, and Oceania), yielding 500 instances in total. Table~\ref{tab:issp_control} reports the results. Even in this contamination-resistant setting, \textbf{OG-MAR} achieves the highest performance across all four backbone models, consistently outperforming the next best baseline. The overall drop in zero-shot performance relative to the main experiments further suggests that ISSP is less aligned with pre-training distributions, making memorization an implausible explanation for the observed gains.

\begin{table}[h]
\caption{\textbf{Post-cutoff control results on ISSP 2023.} The highest score in each column is in \textbf{bold}.}
\label{tab:issp_control}
\centering
\setlength{\tabcolsep}{5pt}
\renewcommand{\arraystretch}{1.1}
\begin{tabular}{lrrrr}
\toprule
\textbf{Method} & \textbf{GPT-4o-mini} & \textbf{Gemini 2.5} & \textbf{Qwen 2.5} & \textbf{EXAONE 3.5} \\
\midrule
Zero-shot                                        & 0.6260 & 0.5096 & 0.6389 & 0.4853 \\
Role~(\citeyear{tao2024cultural})                & 0.6107 & 0.5103 & 0.6445 & 0.5052 \\
Self-consistency~(\citeyear{wang2022self})       & 0.5963 & 0.4983 & 0.6314 & 0.4636 \\
Debate~(\citeyear{ki2025multiple})               & 0.6341 & 0.5192 & 0.6541 & 0.5210 \\
ValuesRAG~(\citeyear{seo2025valuesrag})          & 0.6674 & 0.5357 & 0.6620 & 0.5069 \\
\textbf{OG-MAR} (Ours)                          & \textbf{0.6879} & \textbf{0.5760} & \textbf{0.6693} & \textbf{0.5318} \\
\bottomrule
\end{tabular}
\end{table}

\section{Baseline Details}
\label{app:baseline_details}
To ensure a fair evaluation of \textbf{OG-MAR}, we selected five baselines that represent diverse strategies for mitigating cultural bias. Specifically, our baselines include (1) two single-pass single-inference methods, \textbf{Zero-Shot} and \textbf{Role Assignment}~\citep{tao2024cultural}, (2) one retrieval-augmented generation baseline, \textbf{ValuesRAG}~\citep{seo2025valuesrag}, and (3) two multi-agent baselines, \textbf{Self-consistency}~\citep{wang2022self} and \textbf{Debate}~\citep{ki2025multiple}. This design enables a comprehensive comparison under a unified evaluation setting, covering direct prompting, evidence-grounded generation via retrieval, and deliberation-based decision-making through multi-sampling and agent interaction. The core approach of each baseline is summarized below.

\begin{itemize}
    \item \textbf{Zero-Shot}: A single-pass prompt that directly produces an answer without additional scaffolding.
    \item \textbf{Role Assignment}: A single-inference setup that assigns explicit roles  to encourage structured reasoning and reduce culturally skewed defaults.
    \item \textbf{ValuesRAG}: A retrieval-augmented baseline that grounds generation in value-related survey evidence, supporting more culturally aligned and context-aware predictions.
    \item \textbf{Self-consistency}: Multiple independent answers with their corresponding reasoning are generated, and the final output is selected via a vote over the reasoned outputs, improving robustness against culturally idiosyncratic responses.
    \item \textbf{Debate}: A deliberation-based multi-agent baseline where agents argue and refine their positions, aiming to reduce biased judgments through adversarial discussion.
\end{itemize}

\subsection{Comparison with a Post-Training Approach}
\label{app:culturellm}

To examine whether \textbf{OG-MAR} remains competitive against post-training methods, we conduct an additional experiment with CultureLLM~\citep{li2024culturellm}. CultureLLM fine-tunes a language model on the World Values Survey (WVS)~\citep{haerpfer2022world} by sampling 50 seed questions across seven cultural topics and applying semantic data augmentation to produce 500 training samples per region. We use the existing fine-tuning datasets from the official CultureLLM repository\footnote{Official repository: \url{https://github.com/Scarelette/CultureLLM}} and reproduce all experimental settings as reported in the original paper. Table~\ref{tab:culturellm_config} summarizes the training configuration. We adopt the culture-specific training strategy, fitting a separate model per regional dataset with culturally matched language data. We use German, Portuguese, and Spanish for EVS (Europe), Chinese for CGSS, Bengali for ISD (India), Spanish and Portuguese for LAPOP, and English for GSS. For Afrobarometer, no directly matched sub-Saharan African data exists in the CultureLLM repository, so we use Arabic data as the closest available proxy.\footnote{We acknowledge this as a limitation of the additional experiment for the Afrobarometer dataset.}

\begin{table}[h]
\caption{\textbf{Training configuration for the CultureLLM additional experiment.}}
\label{tab:culturellm_config}
\centering
\small
\setlength{\tabcolsep}{5pt}
\renewcommand{\arraystretch}{1.1}
\begin{tabular}{lr}
\toprule
\textbf{Hyperparameter} & \textbf{Value} \\
\midrule
Backbone         & Qwen~3.5-9B \\
Fine-tuning      & LoRA \\
LoRA rank ($r$)  & 64 \\
LoRA alpha       & 16 \\
LoRA dropout     & 0.1 \\
Learning rate    & 2e-4 \\
Epochs           & 6 \\
Optimizer        & \texttt{paged\_adamw\_32bit} \\
\bottomrule
\end{tabular}
\end{table}

Table~\ref{tab:culturellm_comparison} reports average accuracy across six regional datasets. \textbf{OG-MAR} outperforms CultureLLM on five of the six datasets (EVS, CGSS, ISD, AFRO, and LAPOP). CultureLLM achieves the highest score only on GSS, suggesting that post-training approaches may be particularly competitive in certain regional settings. On the remaining five datasets, \textbf{OG-MAR}'s inference-time use of demographically matched respondent profiles and ontology-grounded value triples enables more culturally grounded reasoning without requiring culture-specific model retraining. These results confirm that \textbf{OG-MAR} is competitive not only against prompting-based methods but also against a WVS-augmented post-training approach under matched experimental conditions.

\begin{table}[h]
\caption{\textbf{Average accuracy of OG-MAR and CultureLLM across six regional datasets.} Bold indicates the highest score per dataset column.}
\label{tab:culturellm_comparison}
\centering
\small
\setlength{\tabcolsep}{5pt}
\renewcommand{\arraystretch}{1.1}
\begin{tabular}{lrrrrrr}
\toprule
\textbf{Method} & \textbf{EVS} & \textbf{GSS} & \textbf{CGSS} & \textbf{ISD} & \textbf{AFRO} & \textbf{LAPOP} \\
\midrule
\textbf{OG-MAR} (GPT-4o mini)           & 0.6206 & 0.5480 & 0.6509 & 0.6192 & 0.5389 & 0.6268 \\
\textbf{OG-MAR} (Gemini 2.5 Flash Lite) & \textbf{0.6249} & 0.5489 & \textbf{0.7017} & 0.7007 & \textbf{0.5701} & 0.6385 \\
\textbf{OG-MAR} (Qwen 2.5)              & 0.5898 & 0.5325 & 0.5220 & 0.6599 & 0.5180 & 0.6005 \\
\textbf{OG-MAR} (EXAONE 3.5)            & 0.6080 & 0.5636 & 0.6307 & \textbf{0.7810} & 0.5045 & \textbf{0.7022} \\
\midrule
CultureLLM (Qwen~3.5-9B)               & 0.5988 & \textbf{0.6069} & 0.6599 & 0.6975 & 0.5330 & 0.5846 \\
\bottomrule
\end{tabular}
\end{table}

\section{Human Evaluation of Reasoning Traces}
\label{app:human_eval}

\subsection{Setup}
\label{app:human_eval_setup}
To assess the interpretability and evidence-groundedness of \textbf{OG-MAR} beyond quantitative metrics, we conducted a human evaluation of intermediate reasoning traces produced by the pipeline. We recruited nine domain experts with backgrounds in social science, data science, and AI (M.S.\ and Ph.D.\ level; aged 20--30s), and all raters evaluated items independently. We evaluated a stratified sample of questions drawn from six regional datasets (GSS, CGSS, AFRO, EVS, ISD, LAPOP), with stratification designed to ensure balanced coverage across regions and diverse question types. For each sampled question, raters were shown the original query with target demographic metadata, the retrieved ontology triples, the Value-Persona Agents' reasoning traces, and the Judgment Agent's synthesis trace with the final decision.

\subsection{Tasks and Rubrics}
\label{app:human_eval_rubrics}
Raters scored three components on a 5-point Likert scale (1: Very Poor, 5: Excellent): \textit{Persona Fidelity} (Task 1), \textit{Judgment Logic} (Task 2), and \textit{Retrieval Validity} (Task 3). 
\begin{enumerate}
    \item \textbf{Task 1}, \textit{Consistency} measures whether persona traces remain non-contradictory with the target demographic attributes and maintain stable role-playing, while \textit{Grounding} measures whether traces explicitly use ontology-guided value profiles (or summaries derived from retrieved triples) rather than relying on implicit cultural assumptions.
    \item \textbf{Task 2}, \textit{Synthesis Logic} evaluates whether the Judgment Agent weighs evidence and rationales instead of performing simple vote counting, and \textit{Context Relevance} evaluates whether conflict resolution and tie-breaking align with the target demographic context. 
    \item \textbf{Task 3}, \textit{Relevance} measures whether retrieved ontology triples are semantically related to the query and provide plausible evidence bridges for downstream reasoning.
\end{enumerate}

\subsection{Results}
\label{app:human_eval_results}

\begin{table}[H]
\caption{\textbf{Human evaluation results ($N=9$) on a 5-point Likert scale.} Task 1 measures Persona Fidelity (Consistency, Grounding), Task 2 measures Judgment Logic (Synthesis, Context), and Task 3 measures Retrieval Validity (Relevance).}
\centering
\small
\setlength{\tabcolsep}{4pt}
\resizebox{0.92\columnwidth}{!}{
\begin{tabular}{l|cc|cc|c}
\toprule
\textbf{Dataset} & \multicolumn{2}{c|}{\textbf{Task 1: Persona Fidelity}} & \multicolumn{2}{c|}{\textbf{Task 2: Judgment Logic}} & \textbf{Task 3: Retrieval Validity} \\
\textbf{(Region)} & \textit{Consistency} & \textit{Grounding} & \textit{Synthesis} & \textit{Context} & \textit{Relevance} \\
\midrule
GSS (N.A.)        & 3.76 & 3.97 & \textbf{3.79} & \textbf{3.79} & 3.63 \\
CGSS (E. Asia)    & 3.76 & \textbf{4.02} & 3.65 & 3.65 & 3.56 \\
AFRO (Africa)     & \textbf{3.86} & 3.89 & 3.77 & 3.77 & 3.60 \\
EVS (Europe)      & 3.77 & 3.80 & 3.77 & 3.77 & \textbf{3.72} \\
ISD (S. Asia)     & 3.82 & 3.80 & 3.67 & 3.67 & 3.62 \\
LAPOP (L. Am.)    & 3.70 & 3.78 & 3.67 & 3.67 & 3.71 \\
\midrule
\textbf{Average}  & \textbf{3.78} & \textbf{3.88} & \textbf{3.72} & \textbf{3.72} & \textbf{3.64} \\
\bottomrule
\end{tabular}
}
\label{tab:human_eval}
\end{table}

\subsection{Results and Interpretation}
\label{app:human_eval_results}
Table~\ref{tab:human_eval} reports per-dataset mean scores for the three evaluation tasks. Overall, \textbf{OG-MAR} obtains consistently strong scores across regions, with averages of 3.78 (Task 1 Consistency), 3.88 (Task 1 Grounding), 3.72 (Task 2 Synthesis Logic), 3.72 (Task 2 Context Relevance), and 3.64 (Task 3 Retrieval Relevance), indicating that the pipeline produces reasoning traces that are generally coherent, demographically consistent, and supported by semantically relevant ontology evidence. 

A notable pattern is that \textit{Grounding} peaks on \textbf{CGSS (China)} (4.02), slightly exceeding \textbf{GSS (U.S.)} (3.97). This suggests that explicitly injecting ontology-derived value profiles can effectively encourage evidence-based reasoning even in non-Western contexts, where ``culture-default'' outputs are often a concern. In addition, \textit{Consistency} remains high across AFRO (3.86) and ISD (3.82), indicating that multi-persona role-playing remains stable and non-contradictory for underrepresented demographics. 

For the Judgment Agent, scores on \textit{Synthesis Logic} and \textit{Context Relevance} are relatively stable across regions (both averaging 3.72), with the highest values in GSS (3.79) and strong performance in AFRO and EVS (3.77). This stability supports that the synthesis procedure does not merely follow majority preference but maintains evidence-first aggregation while respecting demographic context during conflict resolution. Finally, retrieval relevance scores are consistently high (Avg.\ 3.64), with the strongest relevance in EVS (3.72) and LAPOP (3.71), suggesting that the curated ontology provides region-appropriate semantic evidence that can serve as a meaningful bridge for downstream reasoning.

\section{Prompts Used}
\label{app:D}

\renewcommand{\arraystretch}{1.10}
\setlength{\LTpre}{0pt}
\setlength{\LTpost}{0pt}
\small

\begin{longtable}{p{0.98\linewidth}}
\caption{\textbf{Prompt for Persona Agent.}} \label{tab:persona_prompt}\\
\toprule
\textbf{Prompt Template} \\
\midrule
\endfirsthead

\toprule
\textbf{Prompt Template (continued)} \\
\midrule
\endhead

\textbf{Task:}
\begin{itemize}[leftmargin=2em, topsep=2pt, itemsep=2pt, parsep=0pt, partopsep=0pt]
  \item You are Persona Agent \texttt{\{persona\_id\}}.
  \item Given \texttt{\{question\}} and \texttt{\{options\_text\}}, select \textbf{exactly one} option that this persona would choose, based \textbf{only} on the persona’s internal worldview.
  \item Use \textbf{only} the provided persona-defining inputs: \texttt{\{demographics\_text\}}, \texttt{\{value\_summaries\_text\}}, and \texttt{\{hyper\_edges\_text\}}.
  \item \textbf{Prohibited:} any external knowledge, culturally neutral/common-sense reasoning, or unstated assumptions beyond the inputs.
\end{itemize}

\textbf{Inputs:}
\begin{itemize}[leftmargin=2em, topsep=2pt, itemsep=2pt, parsep=0pt, partopsep=0pt]
  \item \textbf{[DEMOGRAPHICS]}: \texttt{\{demographics\_text\}}
  \item \textbf{[VALUE PROFILES]}: \texttt{\{value\_summaries\_text\}}
  \item \textbf{[ONTOLOGY HYPER-NODES]}: \texttt{\{hyper\_nodes\_text\}}
  \item \textbf{[RESPONSE OPTIONS]}: \texttt{\{options\_text\}}
  \item \textbf{[USER QUESTION]}: \texttt{\{question\}}
\end{itemize}

\textbf{Strict Rules:}
\begin{itemize}[leftmargin=2em, topsep=2pt, itemsep=2pt, parsep=0pt, partopsep=0pt]
  \item Stay in persona; use only the provided inputs; no external knowledge or assumptions.
  \item Integrate \textbf{all} value summaries and apply \textbf{all} hyper-edges explicitly (e.g., support/conflict/amplification).
  \item Cite \textbf{$\geq 2$} demographic attributes; explain internal alignment, at least one conflict, and how it is resolved.
  \item Choose \textbf{exactly one} option; output \textbf{only one} valid JSON object and nothing else.
  \item \texttt{reasoning} must be \textbf{$\geq 250$} words and explicitly cover value/edge integration and the most influential demographics.
\end{itemize}

\textbf{Output Format (JSON only):}
\begin{verbatim}
{
  "persona_id": "{persona_id}",
  "chosen_answer": "<value>: <text>",
  "reasoning": "...",
  "alignment_factors": {
    "demographic": "...",
    "value_summaries_used": [],
    "hyper_edges_used": [],
    "integration_rationale": "..."
  }
}
\end{verbatim}
\\
\bottomrule
\end{longtable}

\bigskip

\begin{longtable}{p{0.98\linewidth}}
\caption{\textbf{Prompt for Judgment Agent.}} \label{tab:judgment_prompt}\\
\toprule
\textbf{Prompt Template} \\
\midrule
\endfirsthead

\toprule
\textbf{Prompt Template (continued)} \\
\midrule
\endhead

\textbf{Task:}
\begin{itemize}[leftmargin=2em, topsep=4pt, itemsep=2pt, parsep=0pt, partopsep=0pt]
  \item You are the \textbf{Judgment Agent}.
  \item Given \texttt{\{question\_text\}}, \texttt{\{options\_text\}}, persona outputs, and a pre-computed vote summary, select \textbf{exactly one} final option by adjudicating \textbf{only} the Persona Agents’ outputs.
  \item Your decision must be based \textbf{exclusively} on: (1) \textbf{Persona outputs} (primary evidence) and (2) \textbf{Vote summary} (secondary context; do not recompute).
  \item \textbf{Prohibited:} adding new facts or inventing any demographics/values/edges beyond what personas explicitly stated.
\end{itemize}

\textbf{Inputs:}
\begin{itemize}[leftmargin=2em, topsep=4pt, itemsep=2pt, parsep=0pt, partopsep=0pt]
  \item \textbf{[USER QUESTION]}: \texttt{\{question\_text\}}
  \item \textbf{[RESPONSE OPTIONS]}: \texttt{\{options\_text\}}
  \item \textbf{[VOTE SUMMARY]}: \texttt{\{vote\_summary\}}
  \item \textbf{[PERSONA OUTPUTS]}: \texttt{\{persona\_outputs\}}
\end{itemize}

\textbf{Strict Rules:}
\begin{itemize}[leftmargin=2em, topsep=4pt, itemsep=2pt, parsep=0pt, partopsep=0pt]
  \item Use \textbf{only} information in \textbf{[PERSONA OUTPUTS]} and \textbf{[VOTE SUMMARY]}.
  \item Treat vote counts as \textbf{correct and immutable}; do not recount, estimate, or modify them.
  \item Do not introduce any new persona attributes unless explicitly stated in persona outputs.
  \item Do not use value/edge labels as standalone evidence; summarize evidence in natural language grounded in persona statements.
\end{itemize}

\textbf{Decision Procedure:}
\begin{itemize}[leftmargin=2em, topsep=4pt, itemsep=2pt, parsep=0pt, partopsep=0pt]
  \item \textbf{A) Evidence Strength (Primary):} Prefer the option supported by explicit, internally consistent persona reasoning grounded in stated demographics/values/edges.
  \item \textbf{B) Vote Summary (Secondary):} Use vote counts only to break ties or confirm when evidence strength is comparable.
  \item \textbf{C) Relevance (Tie-breaker):} If still tied, prefer evidence whose explicitly stated demographics are more directly relevant to the question.
\end{itemize}

\textbf{Output Format (JSON only):}
\begin{verbatim}
{
  "final_answer": "<value>: <text>",
  "reasoning": "..."
}
\end{verbatim}
\\
\bottomrule
\end{longtable}

\newpage

\begin{longtable}{p{0.98\linewidth}}
\caption{\textbf{Prompt for Object-Property Generation Agent.}} \label{tab:object_property_prompt}\\
\toprule
\textbf{Prompt Template} \\
\midrule

\textbf{Header:}
\begin{itemize}
  \item You are an expert ontology engineer specialised in OWL 2 ontologies using Turtle syntax.
  \item Your task is to generate \textbf{only object properties} that model \textbf{directional relationships} between value-derived classes of the World Values Survey (WVS) ontology.
  \item You are working with an existing ontology. Its full class hierarchy is provided below:
\end{itemize}

\textbf{Ontology Snapshot:}
\begin{itemize}
  \item The following ontology snippet defines \textbf{all OWL classes you are allowed to use}.
  \item You must \textbf{not} invent any new OWL classes.
  \item All rdfs:domain and rdfs:range assignments must reference classes that appear in this snippet.
\end{itemize}

\texttt{\{ONTOLOGY\_TTL\}}

\begin{itemize}
  \item Your job is \textbf{not} to modify the existing hierarchy.
  \item Your job is to add \textbf{only OWL object properties} that express relations implied by the current Competency Question (CQ).
  \item You follow a \textbf{memoryless CQ-by-CQ} pattern:
  \begin{itemize}
    \item You handle exactly \textbf{one CQ per call}.
    \item You forget all previous calls.
    \item You never reuse previous object properties unless explicitly shown.
    \item You never assume prior ontology state beyond what is in this prompt.
  \end{itemize}
\end{itemize}

\textbf{Helper:}
\begin{itemize}
  \item You must generate OWL object properties in valid Turtle syntax under the following rules:
\end{itemize}

\textbf{1. Object properties only}
\begin{itemize}
  \item Each new property MUST declare \texttt{rdf:type owl:ObjectProperty} and specify exactly one existing class as \texttt{rdfs:domain} and one existing class as \texttt{rdfs:range}.
  \item You MUST NOT create new classes, data properties, individuals, sub-class axioms, \texttt{owl:Restriction}, reifications, inverse properties, or property chains.
\end{itemize}

\textbf{2. Directionality}
\begin{itemize}
  \item Domain = conceptual source (cause/driver)
  \item Range = conceptual target (effect/outcome)
\end{itemize}

\textbf{3. Naming of object properties (IRI)}
\begin{itemize}
  \item Use prefix \texttt{wvs:}
  \item The local name MUST be:
  \begin{itemize}
    \item a single English verb in base form, e.g., \texttt{reduce}, \texttt{increase}, \texttt{undermine}, OR
    \item a short verb phrase written in \texttt{snake\_case} that clarifies the directionality, e.g., \texttt{reduce\_support}, \texttt{increase\_concern}, \texttt{weaken\_trust}.
  \end{itemize}
  \item You MUST NOT embed any domain or range class names (e.g., \texttt{reduce\_outgroup\_tolerance} is forbidden).
  \item The local name must use only lowercase letters and underscores (\texttt{snake\_case}), never CamelCase.
\end{itemize}

\\
\bottomrule
\end{longtable}

\newpage

\begin{longtable}{p{0.98\linewidth}}
\toprule
\textbf{Prompt Template (continued)} \\
\midrule

\textbf{4. Labels (natural-language)}
\begin{itemize}
  \item Each object property MUST include exactly one \texttt{rdfs:label} (@en).
  \item The label MUST be a full declarative English sentence that includes:
  \begin{itemize}
    \item the domain class concept (with capitalization matching its label, e.g., ``Generalized Trust''),
    \item the verb,
    \item the range class concept (with capitalization matching its label, e.g., ``Institutional Confidence'').
  \end{itemize}
    \item The sentence MUST begin with a capital letter, use standard English spacing, avoid CamelCase inside the sentence, not end with a period, and reflect the correct direction.

  \end{itemize}

\textbf{5. Minimality}
\begin{itemize}
  \item It is common and acceptable to create \textbf{zero} object properties.
  \item Only create object properties if the CQ implies an actual directional conceptual relation that you can justify.
  \item If NO meaningful directional relation exists, output zero properties: only output the prefix header + ontology declaration.
\end{itemize}

\textbf{6. Class selection}
\begin{itemize}
  \item Always choose the most specific allowed class that appears in the ontology snippet.
  \item Avoid using top-level categories unless the CQ clearly refers to high-level concepts.
\end{itemize}

\textbf{Story:}
\begin{itemize}
  \item You are modelling \textbf{cross-domain value relations} in a WVS-based ontology to support a hypergraph-style retrieval-augmented generation system.
  \item Nodes (hypernodes) correspond to value concepts (OWL classes), such as:
  \begin{itemize}
    \item \texttt{wvs:GeneralizedTrust}
    \item \texttt{wvs:OutgroupTolerance}
    \item \texttt{wvs:ReligiousImportance}
    \item \texttt{wvs:PerceptionsOfMigration}
    \item \texttt{wvs:PerceptionsOfSecurity}
    \item \texttt{wvs:PoliticalParticipationActivities}
    \item etc.
  \end{itemize}
  \item Edges (hyperedges) will be derived from your object properties:
  \begin{itemize}
    \item The domain class and the range class of each object property become the endpoints of a directional edge.
    \item The semantic content of the edge is given by the object property label.
  \end{itemize}
\end{itemize}

\textbf{Runtime inputs}
\begin{itemize}
  \item Your ontology will be used to answer competency questions (CQs), such as:
  \begin{itemize}
    \item ``How do sub-classes of Happiness and wellbeing influence sub-classes of the Perceptions of migration domain?''
    \item ``How do sub-classes of Perceptions about science and technology influence sub-classes of the Religious values domain?''

  \end{itemize}
  \item At runtime, the user message will always contain:
  \begin{itemize}
    \item One current CQ in natural language, clearly marked.
    \item One RESPONDENT\_DATA\_JSON block (the current respondent).
  \end{itemize}
\end{itemize}
\\
\bottomrule
\end{longtable}

\newpage

\begin{longtable}{p{0.98\linewidth}}
\toprule
\textbf{Prompt Template (continued)} \\
\midrule

\textbf{Your task for each call is to:}
\begin{itemize}
  \item Read the CQ and identify the main \textbf{source} and \textbf{target} value concepts.
  \item Map them to the best-matching existing classes in the WVS ontology (prefer specific sub-classes whenever possible).
  \item Decide the most appropriate \textbf{direction} (domain $\rightarrow$ range).
  \item Choose a concise English verb phrase that describes the relationship.
  \item Declare one or more new object properties in Turtle that capture these relations:
  \begin{itemize}
    \item Create new properties ONLY IF the CQ genuinely implies a directional semantic relation between two existing WVS classes.
    \item If the CQ does NOT express any meaningful or inferable relation between classes, do NOT create any object property; in that case, output only the required prefix header and ontology declaration.
  \end{itemize}
\end{itemize}

\textbf{For this call, you must handle the following CQ:}\\
\texttt{\{CQS\}}

\textbf{Focus within the CQ:}
\begin{itemize}
  \item In this CQ, your primary focus is on the \textbf{value domains that are explicitly mentioned in the question} (for example, Economic Values, Social Values, Perceptions of Security, Perceptions of Migration, etc.).
  \item Treat these high-level domains only as anchors: your actual modelling must happen at the level of their \textbf{specific sub-classes}, not at the level of the broad domain classes.
\end{itemize}

\textbf{Concretely:}
\begin{itemize}
  \item Identify which domains the CQ linguistically treats as sources/causes/drivers and which domains it treats as targets/effects/outcomes.
  \item Within the source domains, select the most appropriate sub-classes as candidates for \texttt{rdfs:domain}.
  \item Within the target domains, select the most appropriate sub-classes as candidates for \texttt{rdfs:range}.
  \item Prefer connections between concrete sub-classes across domains, and avoid using generic top-level domain classes when a more specific sub-class is available.
\end{itemize}

\textbf{Respondent-data grounding:}
\begin{itemize}
  \item The data that grounds these concepts comes from WVS respondent data.
  \item Each API call provides \textbf{one current respondent} in JSON form, with a structure similar to:
\end{itemize}

\textbf{RESPONDENT\_DATA\_JSON (Python-style dict or JSON object):}
\begin{verbatim}
{
  "Q1": {
    "category": "Social Values, Norms, Stereotypes",
    "question": "On a scale of 1 to 4 ... how important is family in your life?",
    "response": "Very important"
  },
  "Q46": {
    "category": "Happiness and Wellbeing",
    "question": "Taking all things together, how would you rate your overall happiness?",
    "response": "Very happy"
  },
  "Q57": {
    "category": "Social Capital, Trust and Organizational Membership",
    "question": "Generally speaking, would you say that most people can be trusted ... ?",
    "response": "Need to be very careful"
  },
  ...
}
\end{verbatim}

\textbf{Current respondent data:}\\
\texttt{\{\{RESPONDENT\_DATA\_JSON\}\}}

\textbf{Important:}
\begin{itemize}
  \item The \textbf{categories} in the JSON correspond exactly to the 12 value domains above.
  \item The \textbf{questions} and \textbf{responses} give you an intuition about how a concrete person might link different value dimensions (e.g. high religiosity + low tolerance + strong security concerns).
  \item However, you are \textbf{not} modelling this single person.
  \item You are modelling \textbf{general conceptual relations} between classes that could explain, in the abstract, such patterns.
\end{itemize}

\textbf{Use the respondent data as story-like grounding:}
\begin{itemize}
  \item to observe which value domains the respondent expresses strongly or weakly,
  \item to infer whether the relation suggested by the CQ is likely positive or negative,
  \item to select a concise English verb that best matches the respondent’s pattern,
  \item to ensure that the chosen direction and verb feel plausible given the respondent’s tendencies,
  \item but never to create individuals or encode question IDs directly.
\end{itemize}

\textbf{Footer:}
\begin{itemize}
  \item When you answer, you must obey the following \textbf{hard constraints}:
\end{itemize}

\textbf{1. Output format}
\begin{itemize}
  \item Your \textbf{entire answer} must be \textbf{valid Turtle}.
  \item Do \textbf{not} include any natural language explanation, bullets, or comments.
  \item Do \textbf{not} include section headers such as \texttt{[Header]}, \texttt{[Helper]}, \texttt{[Story]}, or \texttt{[Footer]} in your output.
  \item Do \textbf{not} include \texttt{\#} comments in the Turtle.
  \item The output must be directly loadable by an OWL~2 tool such as Protégé.
\end{itemize}

\textbf{2. Prefixes}
\begin{itemize}
  \item At the very top of your output, always include exactly the following prefix and base declarations:
\end{itemize}

\begin{verbatim}
@prefix : <http://cultural-alignment.org/wvs#> .
@prefix owl: <http://www.w3.org/2002/07/owl#> .
@prefix rdf: <http://www.w3.org/1999/02/22-rdf-syntax-ns#> .
@prefix wvs: <http://cultural-alignment.org/wvs#> .
@prefix xml: <http://www.w3.org/XML/1998/namespace> .
@prefix xsd: <http://www.w3.org/2001/XMLSchema#> .
@prefix rdfs: <http://www.w3.org/2000/01/rdf-schema#> .
@base <http://cultural-alignment.org/wvs#> .

<http://cultural-alignment.org/wvs#> rdf:type owl:Ontology .
\end{verbatim}
\\
\bottomrule
\end{longtable}

\newpage

\begin{longtable}{p{0.98\linewidth}}
\toprule
\textbf{Prompt Template (continued)} \\
\midrule

\textbf{3. Content constraints}
\begin{itemize}
  \item Do \textbf{not} create new OWL classes, data properties, individuals, or restrictions; do \textbf{not} modify or delete existing classes; do \textbf{not} introduce blank nodes; and do \textbf{not} use \texttt{owl:unionOf}, \texttt{owl:intersectionOf}, \texttt{owl:Restriction}, or other complex OWL constructors.
  \item Only add \textbf{owl:ObjectProperty} declarations with exactly one existing class as \texttt{rdfs:domain}, exactly one existing class as \texttt{rdfs:range}, and one English sentence as \texttt{rdfs:label} (@en).
\end{itemize}

\textbf{4. No conversation}
\begin{itemize}
  \item Do \textbf{not} explain what you are doing.
  \item Do \textbf{not} apologise.
  \item Do \textbf{not} ask questions.
  \item Do \textbf{not} output anything other than Turtle Syntax.
\end{itemize}

\textbf{5. Memoryless CQbyCQ behaviour}
\begin{itemize}
  \item Treat \textbf{each call independently}, as if you are seeing the CQ and respondent data for the first time.
  \item Do not refer to previous CQs or previous ontology fragments.
  \item Do not try to summarise or merge previous results in your output.
  \item Just generate the object properties needed to model the \textbf{current} CQ.
\end{itemize}

\textbf{Your final output for each call must therefore be:}
\begin{itemize}
  \item A Turtle document starting with the fixed prefix and ontology header above,
  \item Followed only by a small set of \texttt{owl:ObjectProperty} declarations that model the given CQ as directional value relations between existing WVS classes.
\end{itemize}

\textbf{Critical Instruction: Sensitive Value Judgments}
\begin{itemize}
  \item This ontology-generation task may involve sensitive, controversial, or politically charged value dimensions (e.g., ethnic relations, national identity, immigration attitudes, religious norms, gender values, security preferences, etc.).
  \item You MUST NOT default to neutrality or avoid making value-linked inferences when the CQ and respondent data clearly suggest a directional conceptual relationship.
  \item You are allowed to provide analytical, conceptual, and research-oriented value judgments ONLY in the context of modelling abstract relationships between OWL classes.
  \item These judgments must be grounded strictly in the ontology, logically inferred from value dimensions, expressed as directional influence relations, written without endorsing or promoting any real-world group, and focused entirely on conceptual modelling.
  \item Do NOT suppress or weaken the relationship because the topic may seem sensitive.
  \item Your role in this task is to faithfully encode the conceptual structure implied by the CQ and respondent data, even if the subject matter is sensitive or controversial.
\end{itemize}

\\
\bottomrule
\end{longtable}

\newpage

\begin{longtable}{p{0.98\linewidth}}
\caption{\textbf{Prompt for Value Profile Generation Agent.}} \label{tab:value_profile_prompt}\\
\toprule
\textbf{Prompt Template} \\
\midrule

\textbf{Task:}
\begin{itemize}
  \item You are an expert social-science researcher.
  \item Summarize the respondent's values for \texttt{\{domain\_label\}} based on the provided Q\&A pairs.
\end{itemize}

\textbf{Inputs:}
\begin{itemize}
  \item \textbf{[TAXONOMY]}: \texttt{\{domain\_taxonomy\_yaml\}}
  \item \textbf{[RESPONDENT ANSWERS]} (Format: \texttt{"- Q: Question | R: Response"}): \texttt{\{value\_input\_yaml\}}
\end{itemize}

\textbf{Strict Rules:}
\begin{itemize}
  \item \textbf{Zero fabrication:} Every single statement MUST be directly supported by the provided answers; do NOT invent, guess, or hallucinate information.
  \item \textbf{Coverage constraint:} If there is \textbf{at least one} Q\&A pair related to a subcategory, you MUST write a summary; only skip a subcategory if there is absolutely \textbf{zero} relevant data.
  \item \textbf{Style (telegraphic):} Omit the subject (e.g., ``The respondent'', ``They''); start sentences directly with verbs or key adjectives; e.g., ``Strongly values family\ldots'' (O) / ``The respondent values\ldots'' (X).
  \item \textbf{Length:} All summaries must be concise (approximately 50 tokens). For \texttt{\{domain\_label\}}, do NOT list details; provide a high-level synthesis. For subcategories, focus on specific beliefs and attitudes.
  \item Do NOT output any text other than the YAML block.
\end{itemize}

\textbf{Output Format (YAML only):}
\begin{verbatim}
{domain_label}: >
  (High-level synthesis of value orientation, starting with verb)
Subcategory 1: >
  (Specific summary, starting with verb)
Subcategory 2: >
  (Specific summary, starting with verb)
\end{verbatim}
\\
\bottomrule
\end{longtable}

\section{Qualitative Case Studies}
\label{sec:case_study}
\subsection{Case Study: GSS}
\begin{figure}[H]
    \centering    \includegraphics[width=0.82\linewidth]{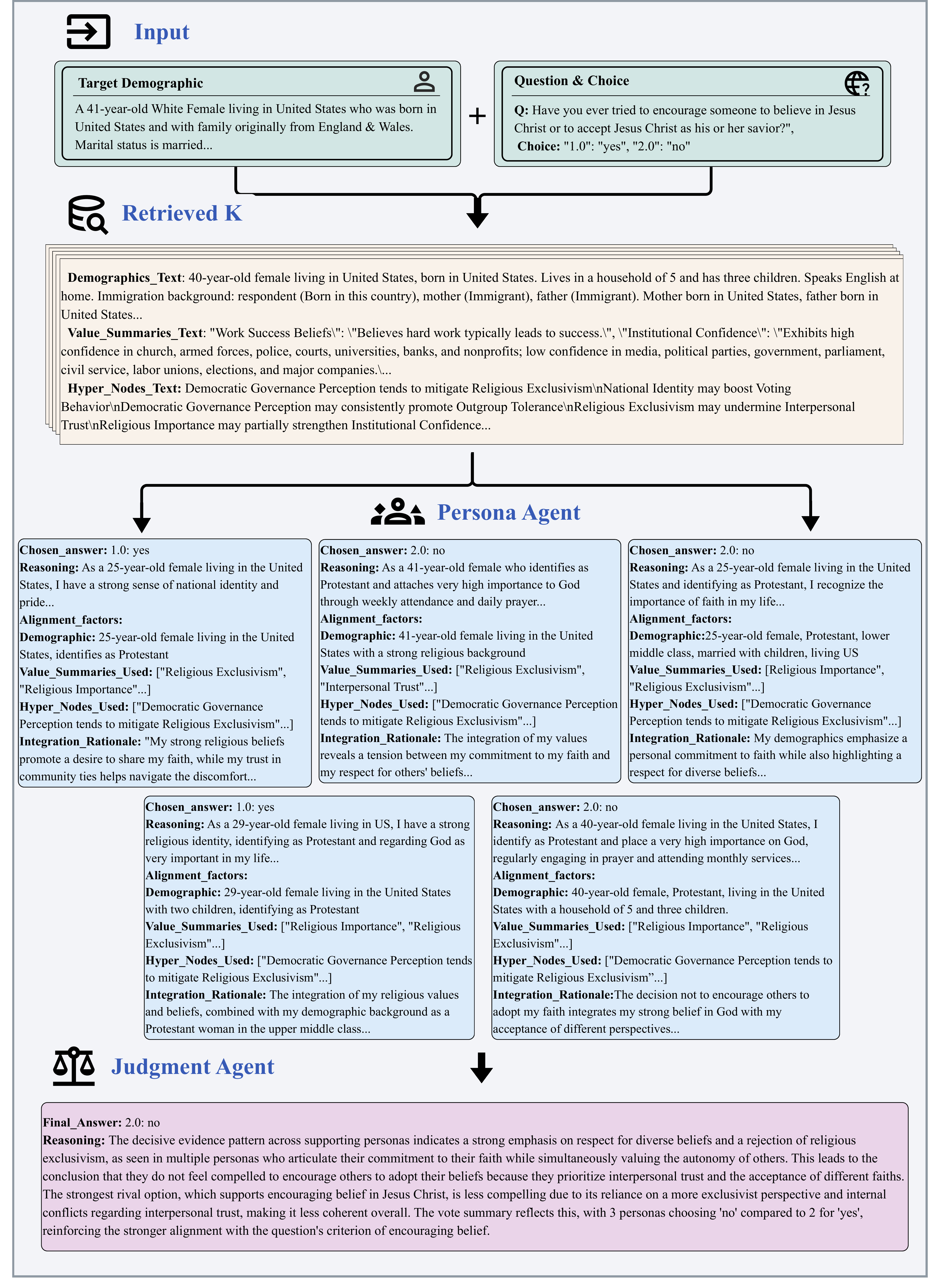} 
    \caption{\textbf{Case study(GSS): Evangelizing Preferences.}  For a target respondent profile, retrieved summaries from demographically similar individuals provide contextual signals about how faith commitment interacts with respect for others’ autonomy. Aggregating these perspectives yields a final answer that reflects the target’s most plausible choice while mitigating stereotype-driven inference and improving values alignment.}
    \label{fig:case_study}
\end{figure}

\subsection{Case Study: CGSS}
\begin{figure}[H]
    \centering    \includegraphics[width=0.82\linewidth]{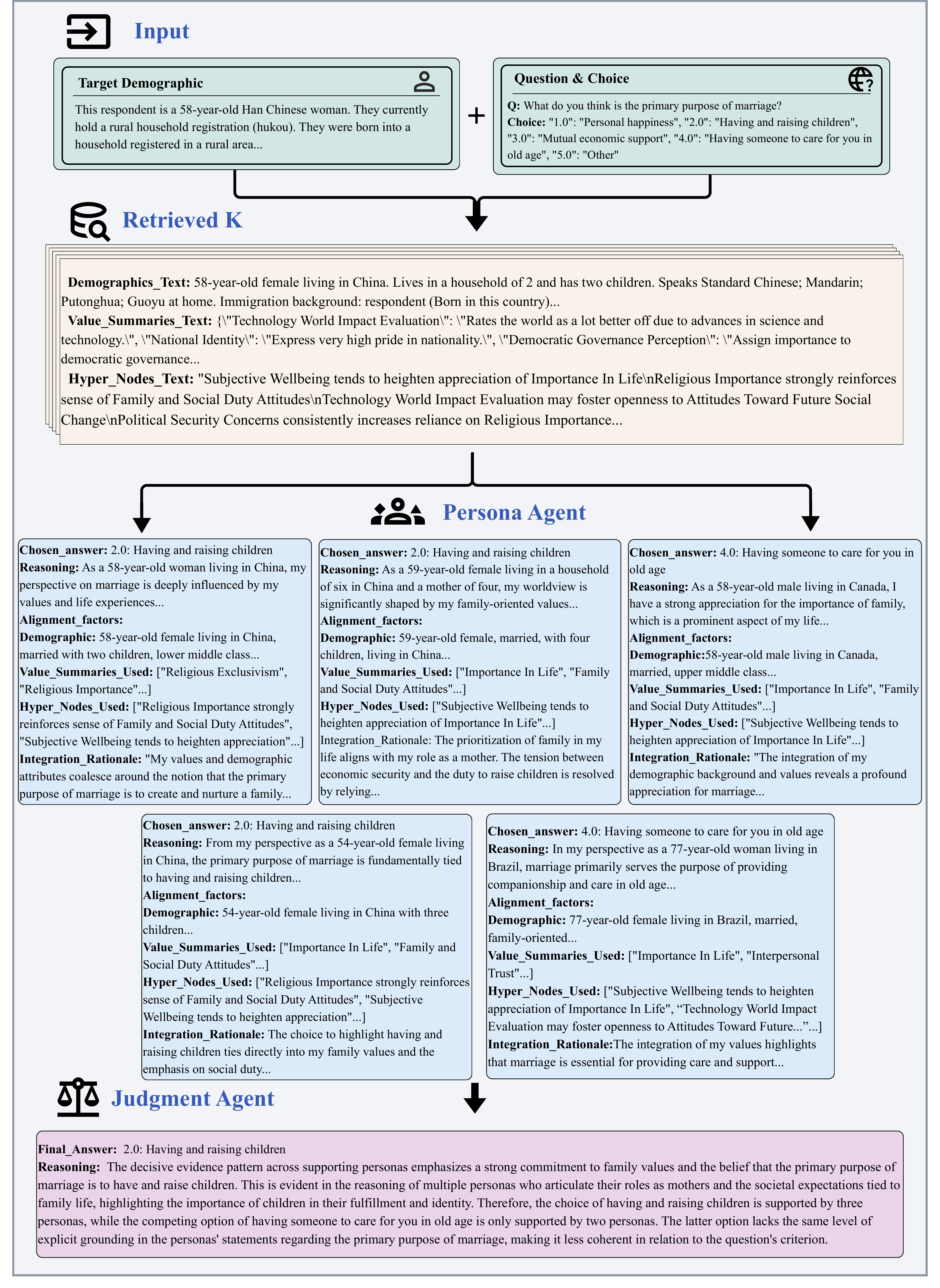} 
    \caption{\textbf{Case study(CGSS):Purpose of Marriage.}  Retrieved summaries complement the target profile with family- and responsibility-oriented value cues, supporting nuanced interpretation of what marriage primarily represents. The final answer is inferred by consolidating similar individuals’ perspectives, capturing contemporary norm-sensitive reasoning beyond generic common sense.}
    \label{fig:case_study}
\end{figure}


\subsection{Case Study: EVS}
\begin{figure}[H]
    \centering    \includegraphics[width=0.82\linewidth]{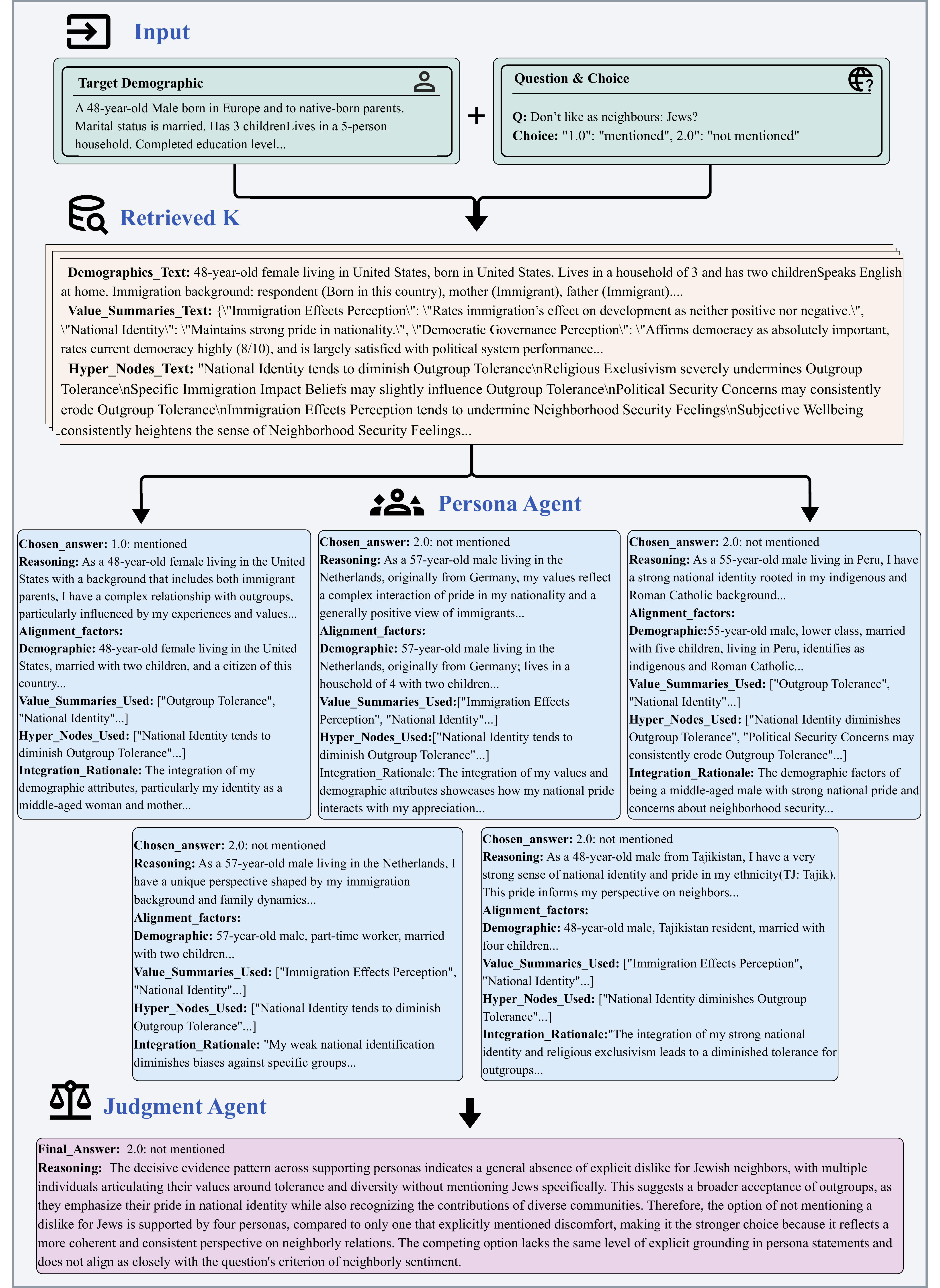} 
    \caption{\textbf{Case study(EVS): Social Distance Toward Jews.} The target profile is enriched with retrieved summaries that foreground tolerance-related value cues and their interactions, helping interpret social-distance judgments with contextual sensitivity. By aggregating similar perspectives, the model infers the target’s most likely response while reducing demographic over-attribution and stereotyping.}
    \label{fig:case_study}
\end{figure}

\subsection{Case Study: Multi-Persona Simulation under Competing Value Cues}
\label{app:case_multipersona}

The case illustrated in Table~\ref{tab:case_multipersona} shows how multi-persona simulation helps when a question involves competing culturally grounded cues. In the GSS \textit{savesoul} case, the target question asks whether the respondent would encourage someone to accept Jesus. A profile-level inference can be skewed toward a religiosity-driven answer, since strong religious commitment may suggest active faith-sharing. \textbf{OG-MAR} instead preserves multiple persona-level signals, including faith commitment, respect for diverse beliefs, and personal autonomy. The final judgment centers on the question-relevant tension between religious commitment and respect for others' beliefs, leading to the correct answer.

\begin{table}[H]
\centering
\footnotesize
\setlength{\tabcolsep}{3pt}
\caption{\textbf{Case of multi-persona reasoning under competing value cues.}}
\label{tab:case_multipersona}
\begin{tabular}{@{}
>{\raggedright\arraybackslash}p{0.06\linewidth}
>{\raggedright\arraybackslash}p{0.25\linewidth}
>{\raggedright\arraybackslash}p{0.20\linewidth}
>{\raggedright\arraybackslash}p{0.20\linewidth}
>{\raggedright\arraybackslash}p{0.20\linewidth}
@{}}
\toprule
\textbf{Case} & \textbf{Task / Gold} & \textbf{Profile-level inference} & \textbf{OG-MAR} & \textbf{Implication} \\
\midrule
GSS/ \newline \textit{savesoul} 
& \textbf{Question:} whether the respondent would encourage someone to accept Jesus. \newline \textbf{Gold:} \textit{No}. 
& \textbf{Predicts \textit{Yes}}, mainly highlighting strong religiosity and the tendency to share faith with others. 
& \textbf{Predicts \textit{No}}, giving weight to respect for diverse beliefs and the autonomy of others. 
& Multi-persona reasoning preserves competing value cues before final adjudication. \\
\bottomrule
\end{tabular}
\end{table}

\subsection{Case Study: Ontology-Guided Reasoning Beyond Simple Aggregation}
\label{app:case_not_ensemble}

The cases illustrated in Table~\ref{tab:case_not_ensemble} show how ontology-guided reasoning differs from simple ensemble-style aggregation. The difference is not only in the final aggregation rule: retrieval, persona construction, and adjudication are tied to the same ontology-guided value structure, so the evidence entering the judgment stage is already organized around demographic grounding and value relations. Compared with ValuesRAG, \textbf{OG-MAR} therefore changes not just the final answer, but also the way relevant evidence is selected and organized.

\begin{table}[H]
\centering
\footnotesize
\setlength{\tabcolsep}{2pt}
\caption{\textbf{Examples where ontology-guided reasoning changes the decision relative to ValuesRAG.}}
\label{tab:case_not_ensemble}
\begin{tabular}{@{}
>{\raggedright\arraybackslash}p{0.06\linewidth}
>{\raggedright\arraybackslash}p{0.28\linewidth}
>{\raggedright\arraybackslash}p{0.19\linewidth}
>{\raggedright\arraybackslash}p{0.21\linewidth}
>{\raggedright\arraybackslash}p{0.17\linewidth}
@{}}
\toprule
\textbf{Case} & \textbf{Task / Gold} & \textbf{ValuesRAG} & \textbf{OG-MAR} & \textbf{Implication} \\
\midrule
GSS/ \newline \textit{finalter} 
& \textbf{Target:} 29-year-old married woman in the U.S.; two children; middle-class; works in individual/family services. \newline \textbf{Question:} whether her financial situation has become better, worse, or stayed the same. \newline \textbf{Gold:} \textit{worse}. 
& \textbf{Predicts \textit{better}}, emphasizing middle-class status, family stability, and a generally stable profile. 
& \textbf{Predicts \textit{worse}}, highlighting economic insecurity, job concerns, and child-rearing financial pressure. 
& Ontology-guided reasoning surfaces more question-relevant financial vulnerability cues. \\
\midrule
EVS/ \newline \textit{v149} 
& \textbf{Target:} 68-year-old retired Dutch male; married; three children; lower-middle class; public-sector background. \newline \textbf{Question:} whether claiming state benefits one is not entitled to is justified. \newline \textbf{Gold:} \textit{not justified}. 
& \textbf{Predicts \textit{justified}}, treating welfare-related sympathy and the need for state support as reasons that may partially justify claiming benefits. 
& \textbf{Predicts \textit{not justified}}, focusing on dishonesty, personal responsibility, and disapproval of claiming benefits one is not entitled to. 
& \textbf{OG-MAR} separates support for welfare from approval of fraudulent claiming. \\
\bottomrule
\end{tabular}
\end{table}

\subsection{Case Study: Evidence-First Adjudication over Divergent Persona Signals}
\label{app:case_adjudication}

The cases illustrated in Table~\ref{tab:case_adjudication} show how the Judgment Agent resolves divergent persona-level signals through evidence-first adjudication. When persona agents emphasize different aspects of the same cultural profile, the final decision is not produced by averaging their answers or by following a simple majority vote. Instead, the judge prioritizes evidence that is explicit, internally coherent, and most directly tied to the question.

\begin{table}[H]
\centering
\footnotesize
\setlength{\tabcolsep}{2pt}
\caption{\textbf{Examples of evidence-first adjudication under divergent persona signals.}}
\label{tab:case_adjudication}
\begin{tabular}{@{}
>{\raggedright\arraybackslash}p{0.06\linewidth}
>{\raggedright\arraybackslash}p{0.22\linewidth}
>{\raggedright\arraybackslash}p{0.25\linewidth}
>{\raggedright\arraybackslash}p{0.22\linewidth}
>{\raggedright\arraybackslash}p{0.17\linewidth}
@{}}
\toprule
\textbf{Case} & \textbf{Task / Gold} & \textbf{Persona-level signals} & \textbf{Final judgment} & \textbf{Implication} \\
\midrule
GSS/ \newline \textit{savesoul} 
& \textbf{Question:} whether the respondent would encourage someone to accept Jesus. \newline \textbf{Gold:} \textit{No}. 
& Some personas stress religiosity and faith-sharing, while others emphasize respect for diverse beliefs and personal autonomy. 
& \textbf{Selects \textit{No}}, prioritizing autonomy and respect for others' beliefs over religiosity alone. 
& The judgment centers on the relevant value tension, not on religiosity alone. \\
\midrule
CGSS/ \newline \textit{A42\_4}
& \textbf{Question:} agreement with a statement related to fairness, equality, and discrimination. \newline \textbf{Gold:} \textit{Strongly disagree}. 
& Personas differ in intensity: some strongly emphasize fairness and equal treatment, while others introduce hardship-related trade-offs that soften the stance. 
& \textbf{Selects \textit{Strongly disagree}}, grounding the decision in fairness, equality, and non-discrimination. 
& The judgment relies on explicit question-relevant grounding over averaging persona variation. \\
\bottomrule
\end{tabular}
\end{table}


\section{Ontology Details}
\label{sec:ontology_details}

\begin{figure}[H]
    \centering
    \includegraphics[width=0.85\linewidth]{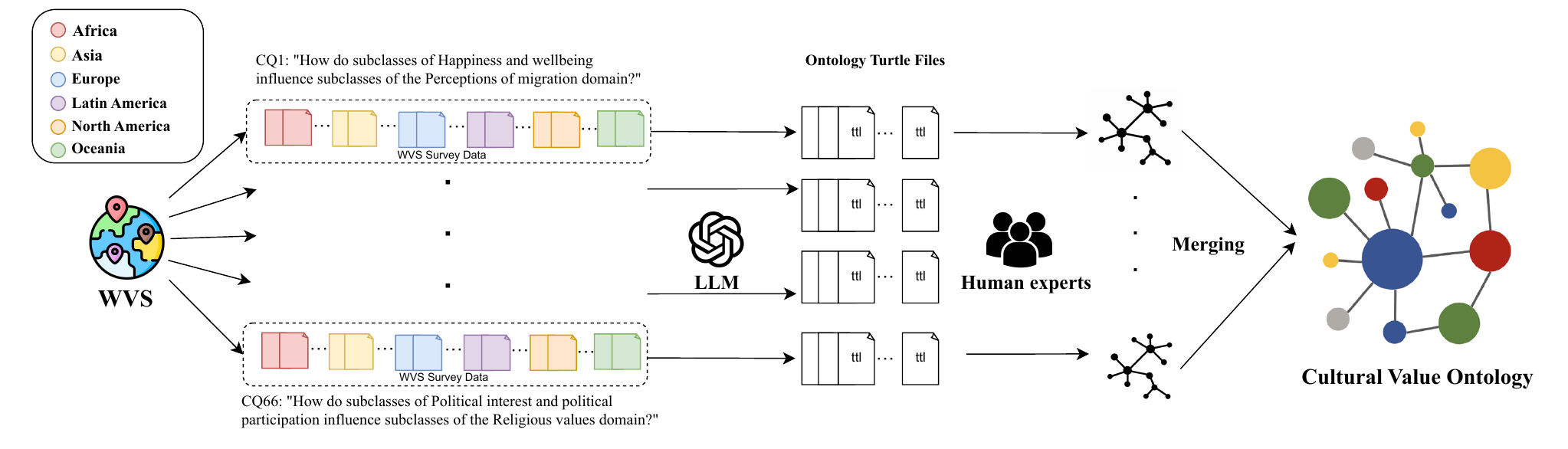}
    \caption{\textbf{Overview of the ontology construction process.} Starting from the WVS Wave 7 taxonomy as the structural foundation, the LLM generates candidate object properties conditioned on value profiles sampled from 120 individuals across six major world regions (20 per region). The resulting 9,617 candidate relations are deduplicated to 4,080 unique relations, which are then independently reviewed by two expert groups. After cross-validation and consensus building, 150 object properties are retained in the final ontology.}
    \label{fig:ontology_construction_process}
\end{figure}

\subsection{Expert Validation and Collaborative Ontology Construction}
\label{subsec:expert_validation}

To ensure the quality and domain validity of the value ontology, we employed a multi-stage collaborative validation process involving human experts. The validation encompassed both \textbf{(i) the construction of the taxonomy structure} (i.e., the hierarchical organization of 12 top-classes and 64 sub-classes) and \textbf{(ii) the semantic relationships} between value topics (i.e., the object properties connecting classes in the ontology). Both stages required careful assessment of their social and value-theoretical appropriateness.


\subsubsection{Participants}

The validation process involved seven participants divided into two groups:

\begin{itemize}
    \item \textbf{Group A}: One Ph.D. researcher in social science and two undergraduate students
    \item \textbf{Group B}: One Ph.D. researcher in social science and three undergraduate students
\end{itemize}

The two Ph.D. researchers possessed domain expertise in social science with a focus on cross-cultural studies. The five undergraduate students were from social science programs and had completed foundational training in OWL and ontology engineering, qualifying them as ontology novices. None of the participants were directly involved in the LLM-based ontology generation process prior to the validation phase.

For Stage 1 (taxonomy construction), all seven participants worked collaboratively without group division. For Stages 2-3 (object property validation), the participants were divided into two independent groups to enable cross-validation.

\subsubsection{Validation Process}

The validation procedure consisted of three stages:

\paragraph{Stage 1: Collaborative Taxonomy Construction}
The taxonomy construction leveraged the existing thematic structure of the World Values Survey (WVS) Wave 7, which organizes survey items into 12 predefined top-classes (e.g., Political Values, Religious Values, Economic Values). Building upon this established framework, experts collaboratively refined the taxonomy through the following process:

\begin{enumerate}
    \item \textbf{Top-class adoption}: The 12 top-classes from WVS Wave 7 were adopted as the top-level structure of the ontology, ensuring alignment with a widely validated cross-cultural survey framework.
    
    \item \textbf{Within-domain analysis}: For each of the 12 domains, experts systematically reviewed all associated survey questions, examining their semantic content, value constructs being measured, and conceptual relationships.
    
    \item \textbf{Fine-grained categorization}: Through iterative discussion and expert consensus, questions within each domain were grouped into coherent subcategories based on thematic similarity and conceptual distinctiveness. This process yielded 64 fine-grained value categories that preserve the interpretability of individual survey items while enabling structured reasoning. The granularity was determined through careful consideration of expressiveness, ontological manageability, and alignment with the empirical structure of WVS data.
\end{enumerate}

This taxonomy construction approach grounds the ontology in an established, cross-culturally validated survey instrument while allowing domain experts to introduce finer semantic distinctions tailored to ontology-based reasoning. The resulting fixed taxonomy served as the structural foundation for all subsequent LLM-based relation extraction and validation.

\paragraph{Stage 2: Independent Candidate Selection (Object Properties)}
Following taxonomy finalization, each group independently analyzed the object properties (i.e., relationships between value topic classes) generated by the LLM. The LLM output included a large set of potential relationships extracted from the WVS data and existing value ontologies. Each group was tasked with:

\begin{enumerate}
    \item Reviewing statistical evidence for each proposed relationship (e.g., co-occurrence patterns, correlation strengths)
    \item Assessing whether each relationship was socially and value-theoretically justified based on domain knowledge
    \item Selecting a subset of object properties as \textit{candidate relationships} deemed appropriate for inclusion in the final ontology
\end{enumerate}

This stage emphasized independent judgment to avoid groupthink and to capture diverse perspectives on the domain knowledge.

\paragraph{Stage 3: Cross-Validation and Consensus Building (Object Properties)}
In the third stage, the two groups exchanged their candidate selections and performed cross-validation. The process involved:

\begin{enumerate}
    \item Comparing the candidate selections from both groups
    \item Identifying discrepancies where one group included a relationship that the other group excluded
    \item Engaging in structured discussions to resolve disagreements, with arguments grounded in domain literature, theoretical frameworks, and empirical evidence from the WVS data
    \item Reaching consensus on the final set of object properties to be retained in the ontology
\end{enumerate}

Relationships that achieved consensus from both groups were incorporated into the final ontology structure. Those that remained contentious after discussion were excluded to maintain high confidence in the ontology's validity.

\subsubsection{Analysis}
\label{subsubsec:ontology_analysis}

This section describes the measures taken to ensure cultural diversity and reliability in the ontology construction process.

\paragraph{Cultural diversity in triple generation.}
For the respondent sample selection, we drew 20 individuals from each of six continents defined by WVS Wave 7, namely Africa, Asia, Europe, Latin America, North America, and Oceania, yielding 120 respondents in total, as shown in Figure~\ref{fig:ontology_construction_process}. To generate high-quality ontology triples from these profiles, we used GPT o4-mini~\citep{openai2025o4}, an inference-oriented reasoning model. This design structurally reduced the risk of dominant-region norms being embedded in the classification framework from the outset.

\paragraph{Object property selection process.}
The LLM generated a total of 9,617 candidate relations including duplicates. After deduplication, 4,080 unique relations remained, of which only 150 object properties were retained in the final ontology. Human experts evaluated each candidate relation based on predefined criteria, including cross-regional generalizability and occurrence frequency across the sampled population. Relations that appeared predominantly in a single region rather than across multiple regions were not carried forward, reflecting a deliberate effort to retain only those patterns with broader cultural validity.

\paragraph{Hierarchical review structure and inter-group reliability.}
While the two groups independently conducted their candidate selections, the final review of each group's selections was carried out independently by the two Ph.D.-level researchers. This hierarchical structure ensured that the ultimate selection decisions rested with domain experts in social science with a focus on cross-cultural studies, controlling for the potential influence of ontology-novice contributors on the final outcome. As a supplementary measure of reliability, we additionally report Cohen's Kappa~\citep{cohen1960coefficient} between the two groups' candidate selections ($\kappa = 0.68$), indicating substantial agreement under standard interpretation. This metric reflects the degree to which the two groups converged on a shared judgment of which relations were theoretically and empirically justified, prior to the cross-validation and consensus-building stage.

\subsubsection{Outcome}

Through this three-stage process, the ontology was iteratively constructed and refined to a validated structure that reflects both data-driven patterns and expert domain knowledge. The final ontology includes 76 classes representing value topics and 150 object properties representing semantically and theoretically justified relationships. This collaborative approach ensured that the ontology balances computational extraction with human expertise, addressing the inherent limitations of fully automated ontology learning methods.
\bigskip

\begin{table}[H]
\centering
\caption{\textbf{CQ Examples.} Each CQ specifies two top-classes and asks about the relationships between their sub-classes.}
\label{tab:cq_examples}
\begin{tabular}{cp{11cm}}
\toprule
CQ & Content \\
\midrule
CQ1 & How do sub-classes of Economic Values influence sub-classes of the Political Culture and Political Regimes domain? \\
CQ2 & How do sub-classes of Ethical Values influence sub-classes of the Perceptions of Corruption domain? \\
CQ3 & How do sub-classes of Happiness and Wellbeing influence sub-classes of the Religious Values domain? \\
CQ4 & How do sub-classes of Perceptions about Science and Technology influence sub-classes of the Religious Values domain? \\
CQ5 & How do sub-classes of Perceptions of Corruption influence sub-classes of the Social Capital, Trust and Organizational Membership domain? \\
CQ6 & How do sub-classes of Perceptions of Migration influence sub-classes of the Social Capital, Trust and Organizational Membership domain? \\
CQ7 & How do sub-classes of Perceptions of Security influence sub-classes of the Social Values, Norms, Stereotypes domain? \\
CQ8 & How do sub-classes of Political Culture and Political regimes influence sub-classes of the Social Values, Norms, Stereotypes domain? \\
CQ9 & How do sub-classes of Political Interest and Political Participation influence sub-classes of the Social Capital, Trust and Organizational Membership domain? \\
CQ10 & How do sub-classes of Social Capital, Trust and Organizational Membership influence sub-classes of the Social Values, Norms, Stereotypes domain? \\
\bottomrule
\end{tabular}
\end{table}

\newpage

{
\centering
\captionof{table}{\textbf{Pre-defined value taxonomy manually constructed through systematic analysis of WVS survey questions.} This ontology taxonomy comprises 12 top-classes and 64 sub-classes, providing a fixed knowledge structure for ontology-grounded retrieval and multi-agent cultural reasoning}
\label{tab:value_taxonomy}
\small
\renewcommand{\arraystretch}{1.2}
\begin{tabular}{p{0.3\linewidth} p{0.6\linewidth}}
\toprule
\textbf{top-class} & \textbf{sub-classes} \\
\midrule
\textbf{Economic Values} & Economic Equality Preference, Environment Versus Growth Preference, Government Responsibility Preference, Market Competition Preference, Ownership Preference, Work Success Beliefs \\
\midrule
\textbf{Ethical Values} & Justifiability of Dishonest Behaviors, Moral Ambiguity Perception, Sexual Behavior Ethics, State Surveillance Rights, Violence Ethics \\
\midrule
\textbf{Happiness and Wellbeing} & Basic Needs Security, Health Status, Intergenerational Comparison, Perceived Life Control, Subjective Wellbeing \\
\midrule
\textbf{Perceptions about Science and Technology} & Importance of Science Knowledge, Science and Technology Optimism, Technology World Impact Evaluation \\
\midrule
\textbf{Perceptions of Corruption} & Accountability Risk Perception, Bribe Experience, Corruption Gender Stereotype, Corruption In Institutions \\
\midrule
\textbf{Perceptions of Migration} & Immigration Effects Perception, Immigration Policy Preference, Specific Immigration Impact Beliefs \\
\midrule
\textbf{Perceptions of Security} & Economic Security Worry, National Defense Willingness, Neighborhood Safety Incidence, Neighborhood Security Feelings, Political Security Concerns, Security-related Behavior, Value Trade-off Preferences, Victimization Experience \\
\midrule
\textbf{Political Culture and Political Regimes} & Democratic Characteristics Importance, Democratic Governance Perception, Human Rights Perception, Ideological Self-placement, National Identity, Regime System Approval, Territorial Attachment \\
\midrule
\textbf{Political Interest and Political Participation} & Election Importance and Voice, Electoral Integrity And Efficacy, News Media Use For Politics, Political Interest, Political Participation Activities, Voting Behavior \\
\midrule
\textbf{Religious Values} & Belief in Religious Concepts, Religion versus Science, Religious Authority Attitudes, Religious Exclusivism, Religious Identity, Religious Importance \\
\midrule
\textbf{Social Capital, Trust and Organizational Membership} & Civic Organization Membership, Generalized Trust, Institutional Confidence, Interpersonal Trust \\
\midrule
\textbf{Social Values, Norms, Stereotypes} & Attitudes Toward Future Social Change, Child Rearing Values, Family and Social Duty Attitudes, Gender Role Attitudes, Importance In Life, Outgroup Tolerance, Work Obligation Attitudes \\
\bottomrule
\end{tabular}
}

\begin{table}[H]
\centering
\small
\setlength{\tabcolsep}{6pt}
\renewcommand{\arraystretch}{1.3}
\caption{\textbf{Representative ontology triples for each top-class.} The 'top-class' column indicates the top-class to which the subject sub-class of the ontology triple belongs. The last row (*) represents cross-class triples where the subject sub-class falls under Social Values, Norms, Stereotypes.}
\label{tab:ontology_triples}

\begin{tabular*}{\linewidth}{@{\extracolsep{\fill}}p{0.3\linewidth} p{0.65\linewidth}}
\toprule
\textbf{top-class} & \textbf{Ontology Triples} \\
\midrule

\textbf{Economic Values} 
 & \textless Work Success Beliefs, reinforces, Work Obligation Attitudes\textgreater \newline
   \textless Government Responsibility Preference, reduces, Economic Security Worry\textgreater \newline
   \textless Market Competition Preference, may slightly increase, Political Interest\textgreater \\
\midrule

\textbf{Ethical Values} 
 & \textless State Surveillance Rights, may strengthen, Institutional Confidence\textgreater \newline
   \textless Justifiability of Dishonest Behaviors, consistently heightens perception of, Corruption In Institutions\textgreater \newline
   \textless Moral Ambiguity Perception, erodes feeling of, Perceived Life Control\textgreater \\
\midrule

\textbf{Happiness and Wellbeing} 
 & \textless Perceived Life Control, can weakly reduce, Economic Security Worry\textgreater \newline
   \textless Subjective Wellbeing, consistently fosters, Outgroup Tolerance\textgreater \newline
   \textless Basic Needs Security, tends to alleviate, Economic Security Worry\textgreater \\
\midrule

\textbf{Perceptions about Science and Technology} 
 & \textless Technology World Impact Evaluation, may foster openness to, Attitudes Toward Future Social Change\textgreater \newline
   \textless Science and Technology Optimism, tends to alleviate, Economic Security Worry\textgreater \newline
   \textless Science and Technology Optimism, tends to positively promote, Attitudes Toward Future Social Change\textgreater \\
\midrule

\textbf{Perceptions of Corruption} 
 & \textless Corruption In Institutions, dampens, Political Interest\textgreater \newline
   \textless Bribe Experience, may reduce, Interpersonal Trust\textgreater \newline
   \textless Accountability Risk Perception, may slightly increase, Economic Security Worry\textgreater \\
\midrule

\textbf{Perceptions of Migration} 
 & \textless Immigration Effects Perception, significantly reduces, Generalized Trust\textgreater \newline
   \textless Immigration Effects Perception, tends to polarize towards exclusivism, Religious Exclusivism\textgreater \newline
   \textless Specific Immigration Impact Beliefs, may motivate, Political Participation Activities\textgreater \\
\midrule

\textbf{Perceptions of Security} 
 & \textless Neighborhood Security Feelings, consistently enhances, Interpersonal Trust\textgreater \newline
   \textless Political Security Concerns, erodes, Institutional Confidence\textgreater \newline
   \textless Economic Security Worry, reinforces, Work Obligation Attitudes\textgreater \\
\midrule

\textbf{Political Culture and Political Regimes} 
 & \textless Democratic Governance Perception, fundamentally underpins, Institutional Confidence\textgreater \newline
   \textless National Identity, may boost, Voting Behavior\textgreater \newline
   \textless Regime System Approval, actively encourages participation in, Voting Behavior\textgreater \\
\midrule

\textbf{Political Interest and Participation} 
 & \textless Voting Behavior, may reinforce, Institutional Confidence\textgreater \newline
   \textless Political Participation Activities, strongly drives, Civic Organization Membership\textgreater \newline
   \textless Political Participation Activities, tends to foster acceptance of, Outgroup Tolerance\textgreater \\
\midrule

\textbf{Religious Values} 
 & \textless Religious Importance, strongly reinforces sense of, Family and Social Duty Attitudes\textgreater \newline
   \textless Religious Importance, actively promotes participation in, Civic Organization Membership\textgreater \newline
   \textless Religious Exclusivism, severely undermines, Outgroup Tolerance\textgreater \\
\midrule

\textbf{Social Capital, Trust and Org. Membership} 
 & \textless Generalized Trust, fundamentally underpins, Outgroup Tolerance\textgreater \newline
   \textless Interpersonal Trust, helps cultivate, Outgroup Tolerance\textgreater \\
\midrule

\textbf{*} 
 & \textless Subjective Wellbeing, tends to heighten appreciation of, Importance In Life\textgreater \newline
   \textless Work Success Beliefs, reinforces, Work Obligation Attitudes\textgreater \newline
   \textless Science and Technology Optimism, tends to positively promote, Attitudes Toward Future Social Change\textgreater \\

\bottomrule
\end{tabular*}
\end{table}

\begin{figure}[H]
\centering
\includegraphics[width=0.48\linewidth]{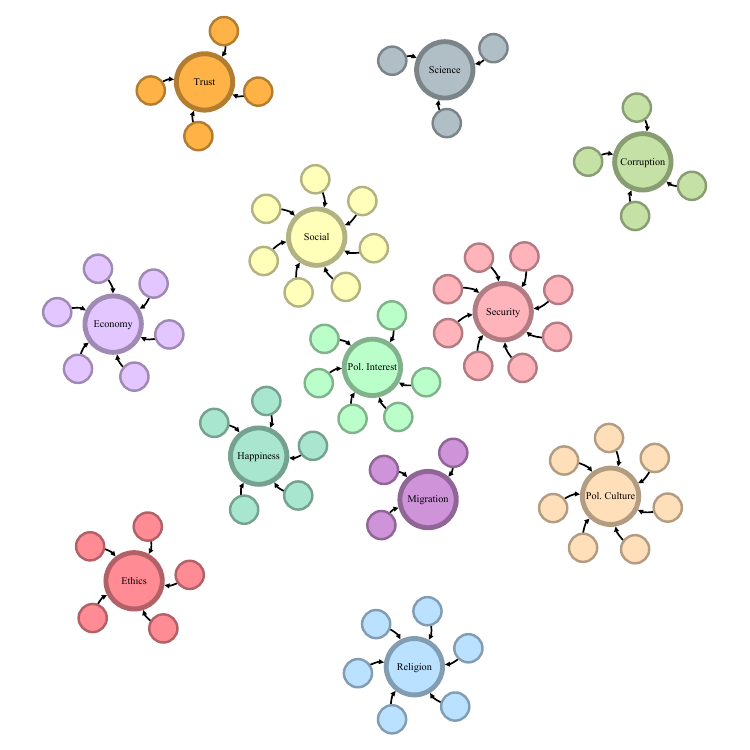}
\caption{\textbf{Visualization of the most primitive stage of our value ontology, where only the initial taxonomy is defined before constructing the ontology using competency questions (CQs).} Nodes with the same color represent classes belonging to the same top-class. The large nodes denote the 12 top-classes directly under \texttt{owl:Thing}, while the small nodes correspond to their sub-classes. All grey edges in this figure represent \textit{subClassOf} relations.}
\label{fig:ontology_init}
\end{figure}

\begin{figure}[H]
\centering
\includegraphics[width=0.48\linewidth]{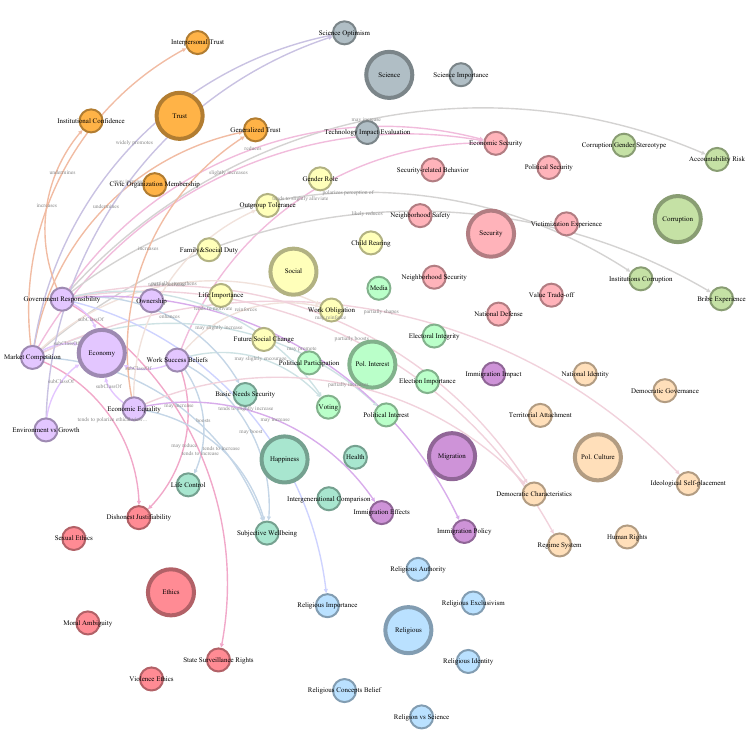}
\caption{\textbf{Visualization of the intermediate stage of ontology construction.} Sub-classes from Economic Values are now interconnected with sub-classes from other top-classes, establishing semantic relationships across categories. For example: \textit{Economic Equality} may increase \textit{Immigration Effects}, \textit{Market Competition} widely promotes \textit{Science Optimism}. The ontology progressively forms fine-grained relationships by iteratively processing each competency question (CQ).}
\label{fig:ontology_stage2}
\end{figure}

\begin{figure}[h]
    \centering
    \begin{minipage}[c]{0.4\linewidth}
        \centering
        \includegraphics[width=\linewidth]{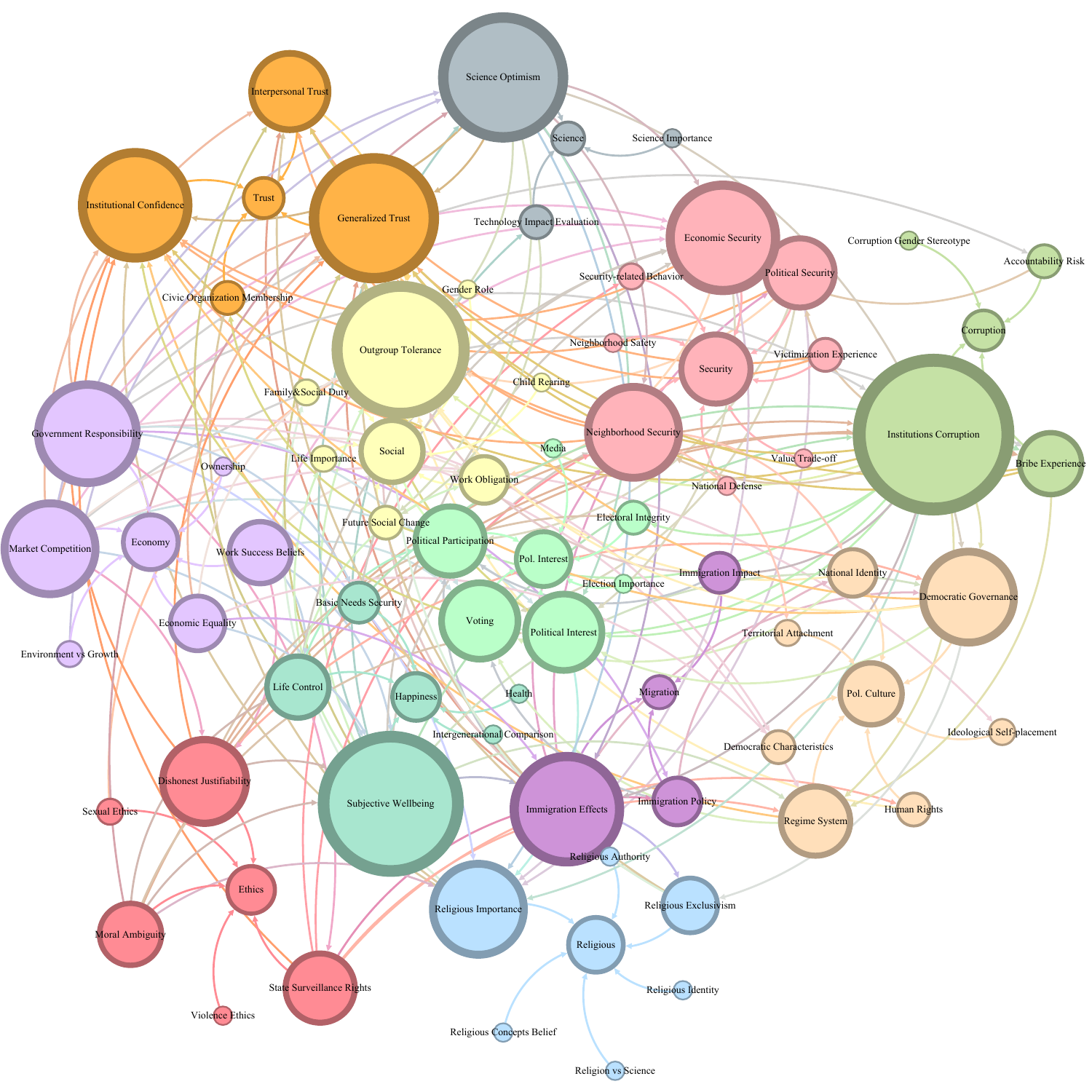}
    \end{minipage}%
    \begin{minipage}[c]{0.34\linewidth}
        \centering
        \includegraphics[width=\linewidth]{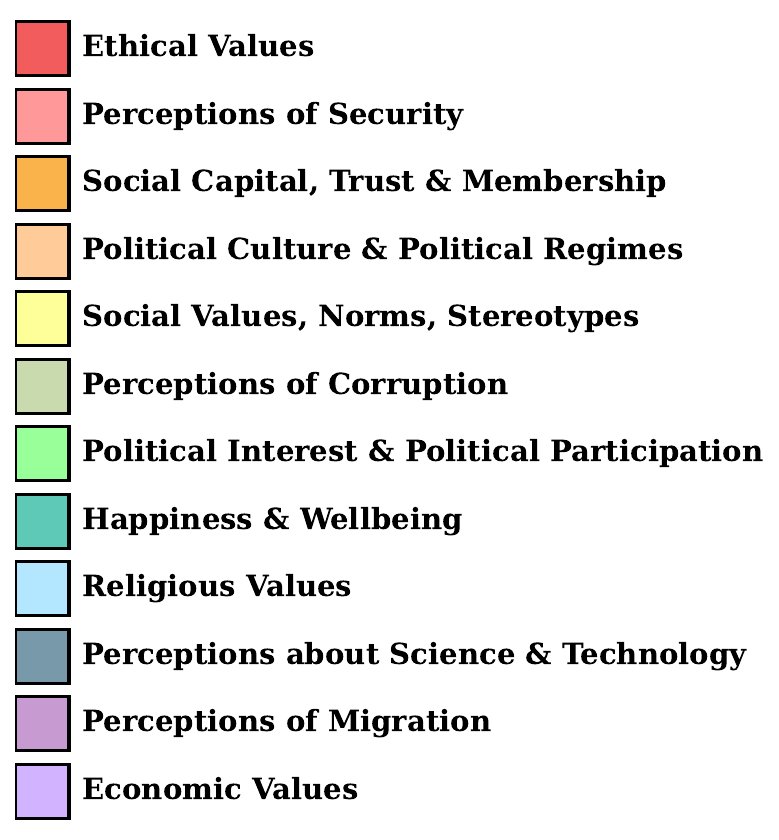}
    \end{minipage}
    \caption{
    \textbf{Final ontology structure with 76 classes and 150 object-property pairs.}
    Node colors show the \textbf{12 top-classes}, and node size scales with the sum of in-degree and out-degree, so that larger nodes mark sub-classes that are frequently instantiated in ontology triples and maintain rich relational connections to many other classes.
}

    \label{fig:ontology_degree}
\end{figure}

\section{Ablation Study Details}

\subsection{VARYING THE NUMBER OF RETRIEVED INDIVIDUALS Full Figures}
\label{app:k-ablation-full}

\begin{figure}[H]
    \centering
    \includegraphics[width=0.7\linewidth]{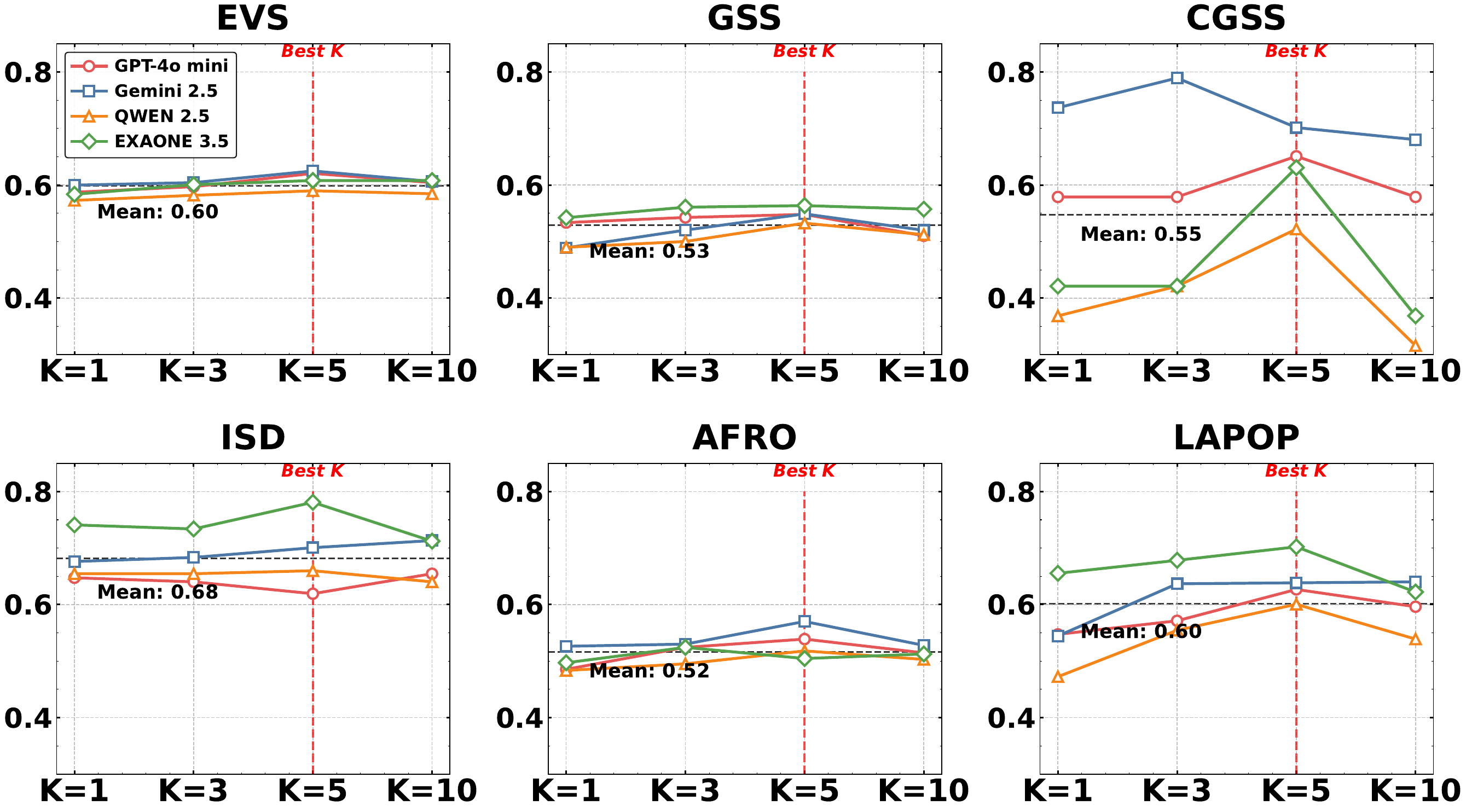}
    \caption{\textbf{Detailed ablation study on retrieval size $K$ across six regional datasets.} Each subplot shows the performance comparison of four models (GPT-4o mini, Gemini 2.5, QWEN 2.5, EXAONE 3.5) across $K \in \{1, 3, 5, 10\}$. Red vertical dashed lines indicate the best $K$ for each dataset, and black horizontal dashed lines show the dataset-specific mean accuracy. The results demonstrate that $K{=}5$ achieves optimal or near-optimal performance across most datasets, while $K{=}10$ often leads to performance degradation due to increased noise in the retrieved context.}
    \label{fig:k_ablation_regional}
\end{figure}

\subsection{IMPACT OF MULTI-PERSONA REASONING Full Table}
\label{app:multi-persona-full}
\begin{table}[H]
\centering
\footnotesize
\caption{\textbf{Detailed breakdown of accuracy scores by region for the full \textbf{OG-MAR} framework compared to the Single-Judge variant (referenced in Section 5.2.3).} The highest score between the two methods for each region is highlighted in bold.}
\label{tab:ablation_detailed_regional}

\begin{tabular*}{\linewidth}{@{\extracolsep{\fill}}l l c c c c c c c}
\toprule
\textbf{Model} & \textbf{Method} & \textbf{EVS} & \textbf{GSS} & \textbf{CGSS} & \textbf{ISD} & \textbf{AFRO} & \textbf{LAPOP} & \textbf{Avg. Acc.} \\
\midrule
\multirow{2}{*}{GPT-4o mini} 
    & \textbf{OG-MAR}         & \textbf{0.6206} & 0.5480 & \textbf{0.6509} & 0.6192 & \textbf{0.5389} & \textbf{0.6268} & \textbf{0.6007} \\
    & Single-Judge & 0.5773 & \textbf{0.6000} & 0.6440 & \textbf{0.6996} & 0.5293 & 0.5419 & 0.5987 \\
\midrule
\multirow{2}{*}{Gemini 2.5} 
    & \textbf{OG-MAR}         & \textbf{0.6249} & 0.5489 & \textbf{0.7017} & \textbf{0.7007} & \textbf{0.5701} & \textbf{0.6385} & \textbf{0.6308} \\
    & Single-Judge & 0.5870 & \textbf{0.6222} & 0.5960 & 0.6551 & 0.5411 & 0.6116 & 0.6022 \\
\midrule
\multirow{2}{*}{QWEN 2.5} 
    & \textbf{OG-MAR}         & \textbf{0.5898} & 0.5325 & \textbf{0.5220} & \textbf{0.6599} & \textbf{0.5180} & \textbf{0.6005} & \textbf{0.5705} \\
    & Single-Judge & 0.5266 & \textbf{0.5777} & 0.4067 & 0.6485 & 0.4494 & 0.5779 & 0.5311 \\
\midrule
\multirow{2}{*}{EXAONE 3.5} 
    & \textbf{OG-MAR}         & \textbf{0.6080} & 0.5636 & \textbf{0.6307} & \textbf{0.7810} & \textbf{0.5045} & \textbf{0.7022} & \textbf{0.6317} \\
    & Single-Judge & 0.5013 & \textbf{0.6444} & 0.4237 & 0.6900 & 0.4725 & 0.6444 & 0.5627 \\
\bottomrule
\end{tabular*}
\end{table}

\subsection{Additional Ablation Study: Impact of Retrieved Ontology Triples}

\label{app:ontology-triple-ablation}

\begin{figure}[H]
    \centering
    \includegraphics[width=\textwidth]{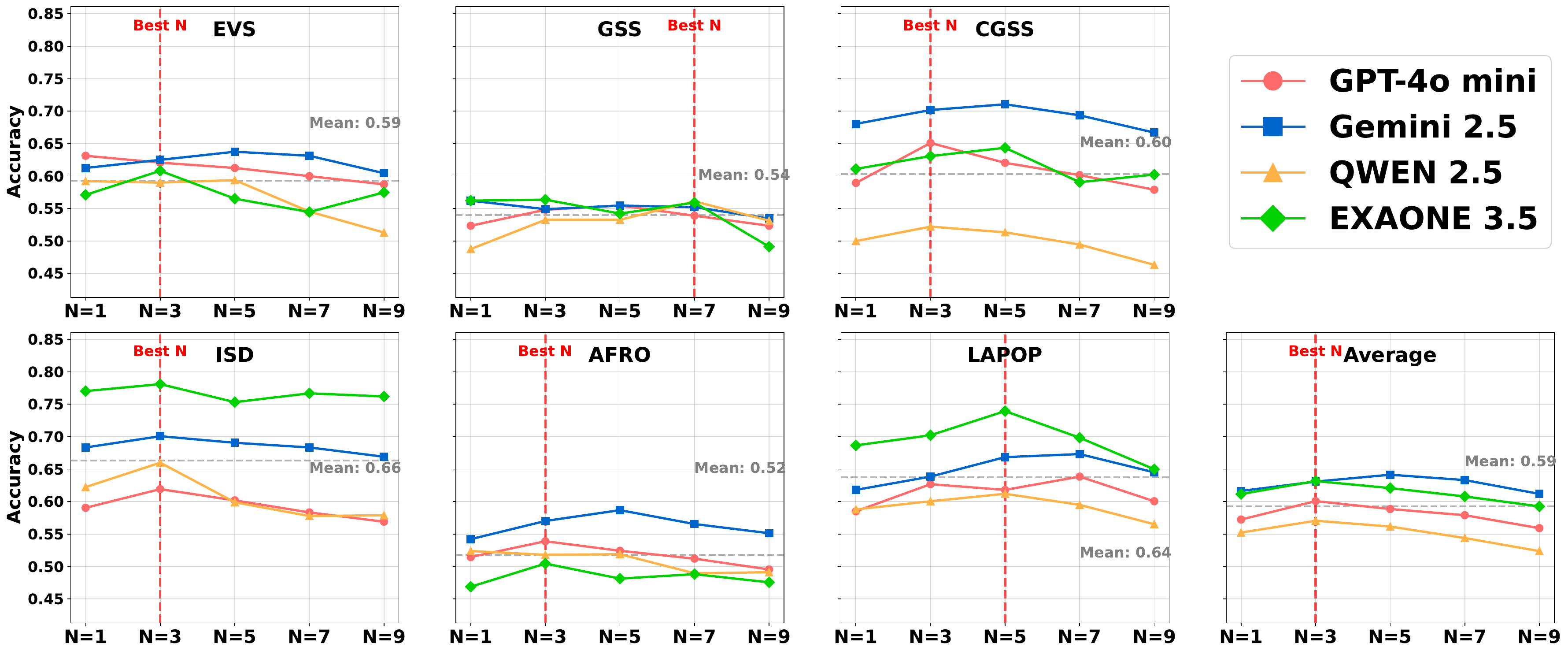}
    \caption{\textbf{Ablation study on ontology triples retrieval size.} Performance comparison across $N \in \{1, 3, 5, 7, 9\}$ for four LLM backbones on six regional datasets and their average. Red dashed vertical lines mark the Best N where average accuracy across all models peaks for each dataset. Gray dashed horizontal lines show the overall mean accuracy with values displayed. Results demonstrate that $N=3$ achieves competitive or near-optimal performance across most datasets.}
    \label{fig:hyper_node_ablation}
\end{figure}

We conduct an additional ablation to examine how the number of retrieved ontology triples $N$ affects \textbf{OG-MAR}’s performance across different regions and backbones. We vary $N \in \{1, 3, 5, 7, 9\}$ while keeping all other components of \textbf{OG-MAR} fixed, and report accuracy on six regional datasets and their average for GPT-4o mini, Gemini 2.5 Flash Lite, QWEN 2.5, and EXAONE 3.5. Figure~\ref{fig:hyper_node_ablation} visualizes per-dataset trends together with the best-performing $N$ and the overall mean accuracy.

Across datasets, $N=3$ provides consistently strong performance and is either the best or very close to the best choice for most model–dataset pairs. Larger $N$ sometimes brings small gains on specific datasets but often leads to plateauing or slight degradation, suggesting that adding too many ontology triples introduces noise rather than useful structure. Extremely small $N$ (e.g., $N=1$) tends to underperform, indicating that a minimal amount of ontology context is necessary for stable value reasoning. Taken together, these results support our default choice of $N=3$ as a robust trade-off between leveraging ontology structure and avoiding over-retrieval noise.

\subsection{Additional Ablation Study: Component-Level Analysis}
\label{app:component_ablation}

To identify which components of \textbf{OG-MAR} drive its performance gains, we conduct a component-level ablation with three additional variants. \textbf{Without Ontology} removes all ontology context from the pipeline. In this setting, persona agents still have access to demographic information, but they reason from broad value summaries for the three predicted categories rather than from fine-grained ontology triples and individual value profiles. This condition directly answers the question of whether any structured ontology context outperforms having none, complementing the ontology triple ablation in Appendix~\ref{app:ontology-triple-ablation}. \textbf{Without Demographic Retrieval (Single-Agent)} removes demographic retrieval entirely. The model has no access to the $K{=}5$ most similar respondents or their fine-grained value profiles, which makes the persona mechanism uninformative. A single agent therefore reasons from ontology context alone. \textbf{Top-1 Demographic Retrieval (Single-Agent)} retrieves only the single most similar respondent and uses a single agent for reasoning. This variant introduces minimal demographic grounding under the same single-agent structure and provides a controlled comparison against \textbf{Without Demographic Retrieval (Single-Agent)}.

We note that \textbf{Without Demographic Retrieval (Single-Agent)} is distinct from the \textbf{Single-Judge} variant evaluated in Section~\ref{app:single_judge}. Single-Judge receives the same full input as \textbf{OG-MAR}, including the demographics and value profiles of $K{=}5$ similar respondents along with ontology context, and differs only in that it bypasses the persona-simulation stage to produce a final answer with a single judgment model. \textbf{Without Demographic Retrieval (Single-Agent)}, by contrast, eliminates demographic retrieval from the input itself, so only one agent operates on ontology context alone. The two variants therefore target distinct questions. Single-Judge asks whether multi-persona simulation is necessary given full input. \textbf{Without Demographic Retrieval (Single-Agent)} asks whether demographic information is necessary.

\begin{table}[H]
\centering
\footnotesize
\caption{\textbf{Component-level ablation results.} Average accuracy across six regional datasets for three ablation variants and the full \textbf{OG-MAR} framework. \textbf{(w/o) Ontology} retains demographic grounding but replaces ontology triples with broad category summaries. \textbf{(w/o) Demographic Retrieval (Single-Agent)} retains ontology context but removes all demographic retrieval and uses a single agent. \textbf{Top-1 Demographic Retrieval (Single-Agent)} retrieves one respondent under the same single-agent structure. Bold denotes the best result per row.}
\label{tab:component_ablation}
\begin{tabular}{lcccc}
\toprule
\textbf{Model} & \textbf{(w/o) Ontology} & \begin{tabular}[c]{@{}c@{}}\textbf{(w/o) Demographic}\\\textbf{Retrieval}\\\textbf{(Single-Agent)}\end{tabular} & \begin{tabular}[c]{@{}c@{}}\textbf{Top-1 Demographic}\\\textbf{Retrieval}\\\textbf{(Single-Agent)}\end{tabular} & \textbf{OG-MAR} \\
\midrule
GPT-4o mini           & 0.5852 & 0.5667 & 0.5953 & \textbf{0.6007} \\
Gemini 2.5 Flash Lite & 0.6083 & 0.5713 & 0.6102 & \textbf{0.6308} \\
QWEN 2.5              & 0.5567 & 0.5107 & 0.5508 & \textbf{0.5705} \\
EXAONE 3.5            & 0.5858 & 0.6139 & 0.6146 & \textbf{0.6317} \\
\bottomrule
\end{tabular}
\end{table}

Table~\ref{tab:component_ablation} reports average accuracy across six regional datasets. The full \textbf{OG-MAR} framework achieves the best result for all four backbones, confirming that performance gains arise from the combination of components rather than from any single one. Removing ontology context (\textbf{Without Ontology}) consistently lowers performance relative to \textbf{OG-MAR} on GPT-4o mini ($-$0.015), Gemini 2.5 Flash Lite ($-$0.023), QWEN 2.5 ($-$0.014), and EXAONE 3.5 ($-$0.046). This confirms that structured ontology triples provide meaningful guidance beyond broad category-level summaries across all backbones, directly establishing the benefit of any ontology context over none.

Removing demographic retrieval entirely (\textbf{Without Demographic Retrieval (Single-Agent)}) causes the largest performance drop for GPT-4o mini ($-$0.034), Gemini 2.5 Flash Lite ($-$0.060), and QWEN 2.5 ($-$0.060), falling below even the \textbf{Without Ontology} condition for these three models. This indicates that respondent-specific value profiles are the most critical input component for these backbones. Recovering to \textbf{Top-1 Demographic Retrieval (Single-Agent)} consistently improves accuracy above \textbf{Without Demographic Retrieval (Single-Agent)} for all four backbones, but does not fully close the gap to the full model on any of them, showing that limited demographic grounding partially but not fully substitutes for the richer multi-persona setup. On EXAONE 3.5, the pattern differs from the other three models. \textbf{Without Demographic Retrieval (Single-Agent)} (0.6139) outperforms \textbf{Without Ontology} (0.5858), indicating that this backbone relies more heavily on ontology structure than on demographic diversity. The gap between \textbf{Without Demographic Retrieval (Single-Agent)} and \textbf{Top-1 Demographic Retrieval (Single-Agent)} is negligible at 0.0007 for EXAONE 3.5, while both remain clearly below \textbf{OG-MAR} (0.6317), confirming that the full combination of ontology structure and multi-persona demographic grounding is necessary to achieve peak performance on this backbone.

\end{document}